\documentclass[10pt,twocolumn,letterpaper]{article}
\usepackage{titling}
\usepackage[pagenumbers]{cvpr} %
\usepackage[accsupp]{axessibility}

\usepackage{amsmath}
\usepackage{amssymb}
\usepackage{booktabs}
\usepackage{bbding}
\usepackage{graphicx}
\usepackage{animate}
\usepackage{grffile}
\usepackage{siunitx} %
\usepackage{enumitem}
\usepackage{float}
\usepackage[table, usenames, dvipsnames]{xcolor}
\usepackage[accsupp]{axessibility}

\definecolor{mycyan}{rgb}{0,0.76,0.75}
\definecolor{LightGrey}{rgb}{0.92,0.92,0.92}

\usepackage[pagebackref,breaklinks,colorlinks]{hyperref}

\usepackage[capitalize]{cleveref}
\crefname{section}{Sec.}{Secs.}
\Crefname{section}{Section}{Sections}
\Crefname{table}{Table}{Tables}
\crefname{table}{Tab.}{Tabs.}

\usepackage{times}
\usepackage{epsfig}
\usepackage{amsmath}
\usepackage{amssymb}
\usepackage{booktabs}
\usepackage{makecell}
\usepackage{multirow}
\usepackage{tabularx}
\usepackage{bbm}
\usepackage{caption} 
\usepackage{subcaption}
\usepackage{animate}
\usepackage{pifont}     %

\newcommand{\bff}{\mathbf{f}}\newcommand{\bF}{\mathbf{F}} %

\newcommand{\bI}{\mathbf{I}}

\newcommand{\bp}{\mathbf{p}}\newcommand{\bP}{\mathbf{P}}

\newcommand{\bs}{\mathbf{s}}
\newcommand{\bt}{\mathbf{t}}\newcommand{\bT}{\mathbf{T}}
\newcommand{\bu}{\mathbf{u}}

\newcommand{\nR}{\mathbb{R}}

\newcommand{\cE}{\mathcal{E}}

\newcommand{\cL}{\mathcal{L}}

\newcommand{\figref}[1]{Fig.~\ref{#1}}
\newcommand{\secref}[1]{Sec.~\ref{#1}}

\newcommand{\tabref}[1]{Table~\ref{#1}}

\DeclareMathOperator*{\argmax}{argmax~}

\makeatletter
\DeclareRobustCommand\onedot{\futurelet\@let@token\@onedot}
\def\@onedot{\ifx\@let@token.\else.\null\fi\xspace}
\def\eg{e.g\onedot} 
\def\ie{i.e\onedot}

\def\wrt{wrt\onedot}

\def\etal{et~al\onedot}

\makeatother

\newcommand*\rot{\rotatebox{90}}

\newcommand{\boldparagraph}[1]{\vspace{0.5em}\noindent{\bf #1.}}
\newcommand{\boldparagraphnoperiod}[1]{\vspace{0.5em}\noindent{\bf #1}}
\renewcommand{\paragraph}[1]{\boldparagraph{#1}}

\definecolor{darkgreen}{rgb}{0,0.7,0}
\definecolor{newyellow}{rgb}{1,0.8,0.05}
\definecolor{newgreen}{rgb}{0.2,0.8,0.2}

\newcolumntype{P}[1]{>{\centering\arraybackslash}m{#1}}

\newcommand{\width}{0.19\textwidth}
\newcommand{\figbenchmark}{
\begin{figure*}[t]
        \centering
        \setlength{\tabcolsep}{0.1em}
        \renewcommand{\arraystretch}{0.5}
        \hfill{}
        \begin{tabular}{lccccc}
            \multicolumn{6}{c}{
            \includegraphics[width=0.8\textwidth]{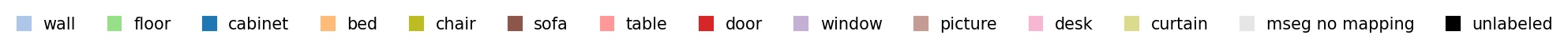}}\\
            \rot{\footnotesize\hspace{7pt}ScanNet~\cite{Dai2017CVPR}} & 
            \includegraphics[width=\width]{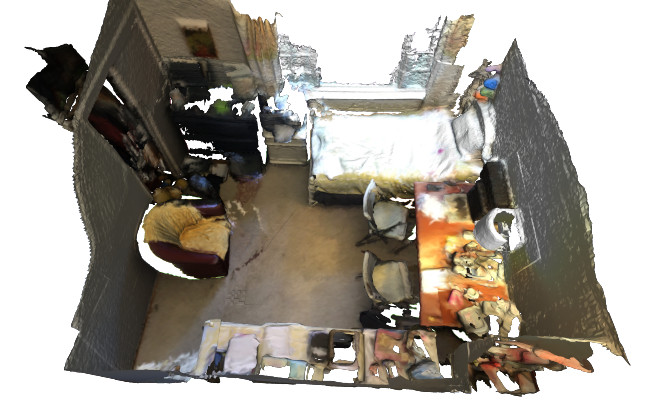}&
            \includegraphics[width=\width]{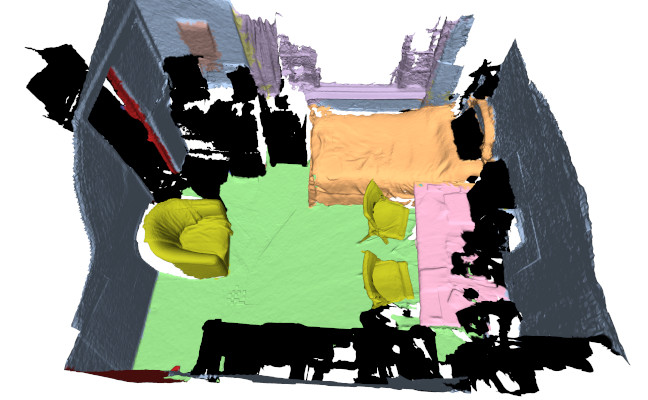}&
            \includegraphics[width=\width]{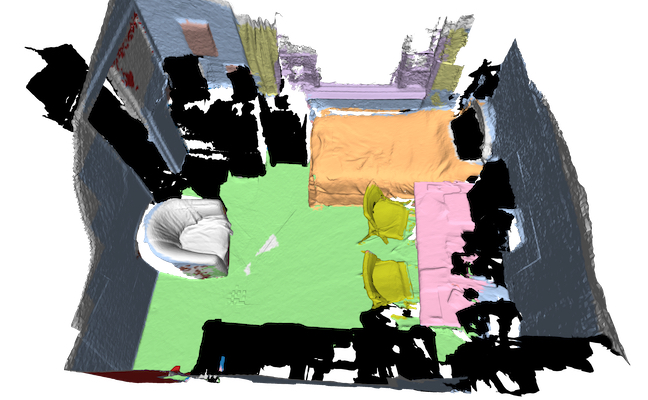}&
            \includegraphics[width=\width]{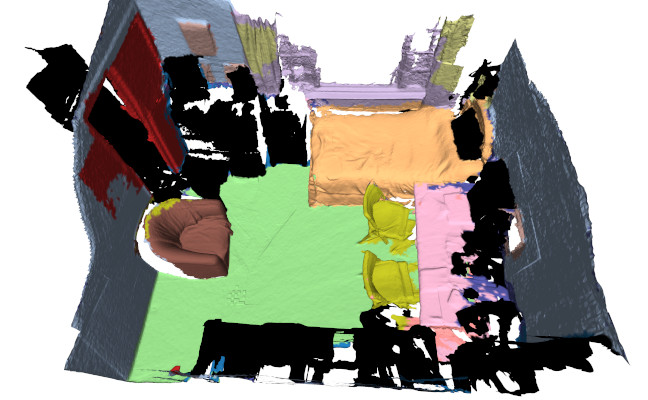}&
            \includegraphics[width=\width]{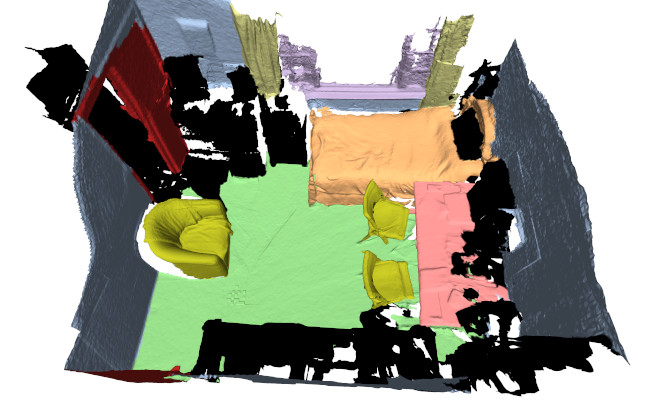}\\
            \multicolumn{6}{c}{
            \includegraphics[width=0.96\textwidth]{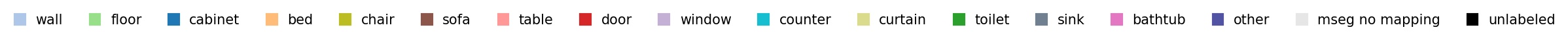}}\\
            \rot{\footnotesize\hspace{1pt} Matterport3D~\cite{Chang2017THREEDV}} & 
            \includegraphics[width=\width]{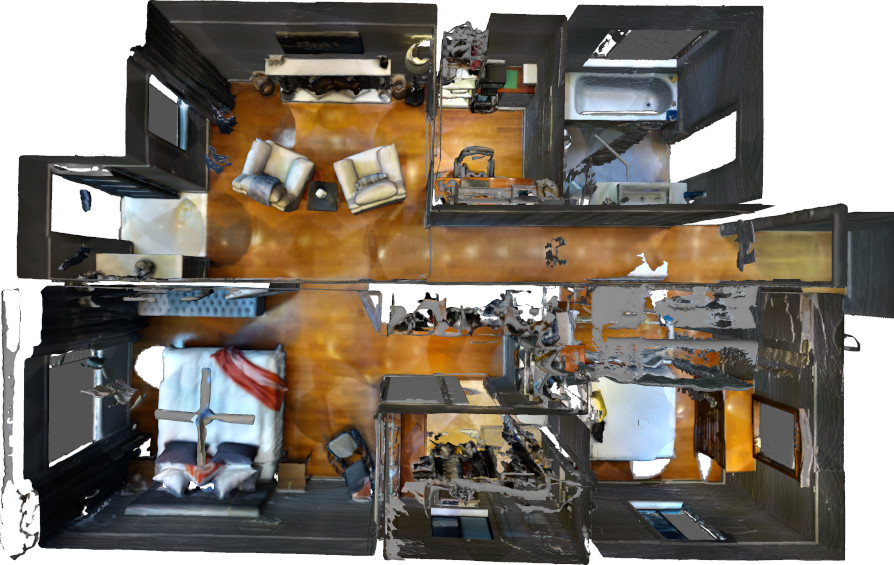}&
            \includegraphics[width=\width]{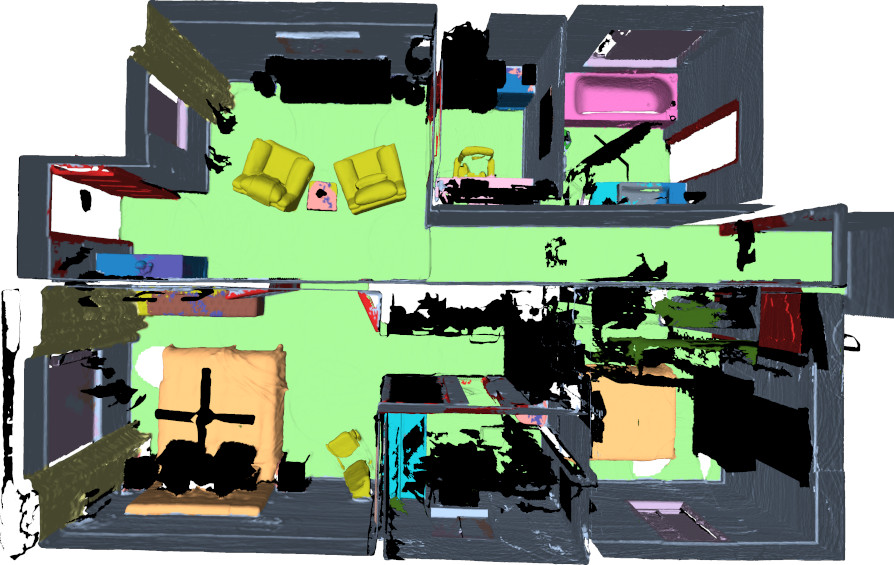}&
            \includegraphics[width=\width]{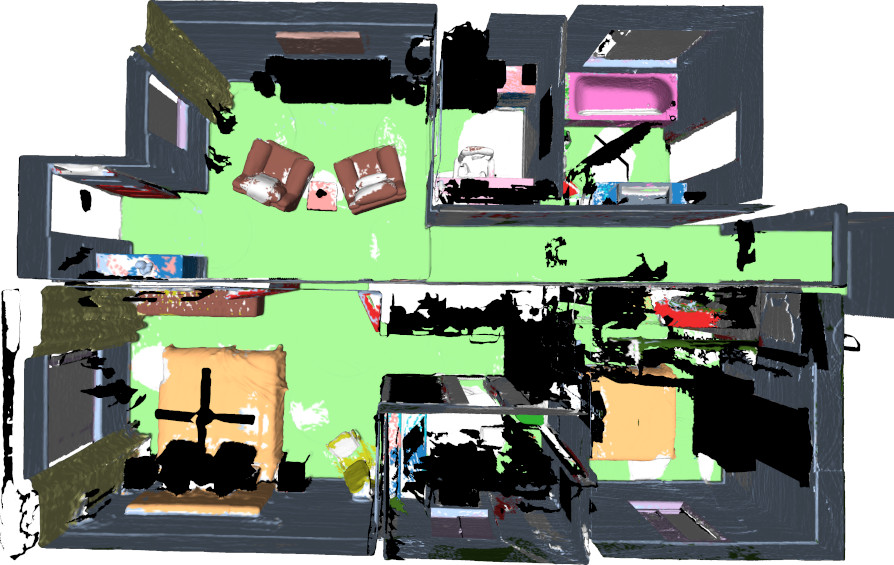}&
            \includegraphics[width=\width]{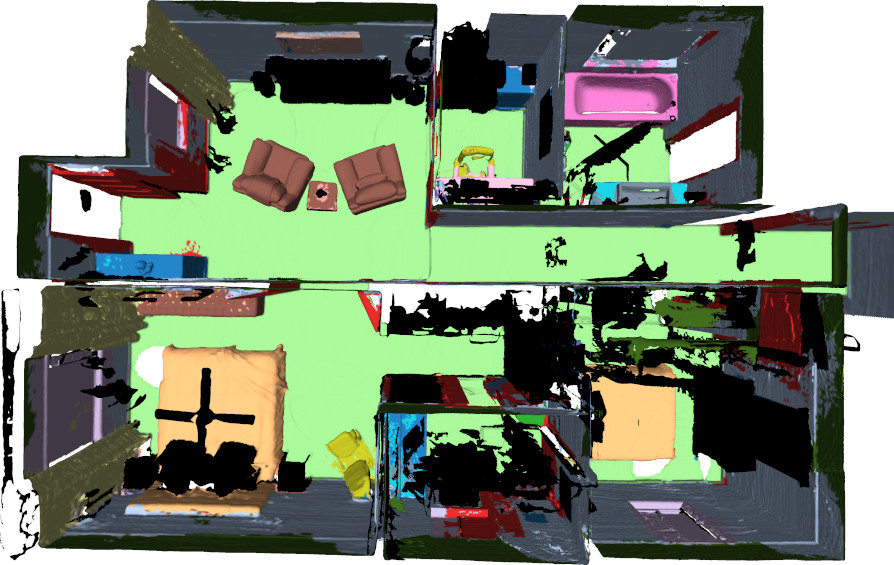}&
            \includegraphics[width=\width]{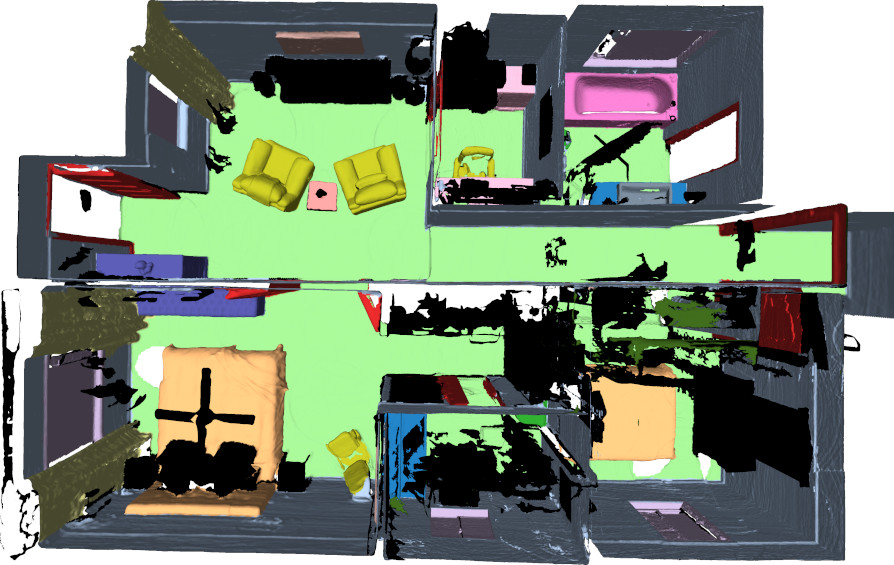}\\
            \multicolumn{6}{c}{
            \includegraphics[width=0.72\textwidth]{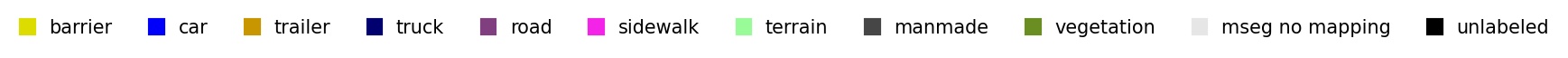}}\\
            \rot{\footnotesize\hspace{0pt}nuScenes~\cite{Caesar2020CVPR}} & 
            \includegraphics[width=\width]{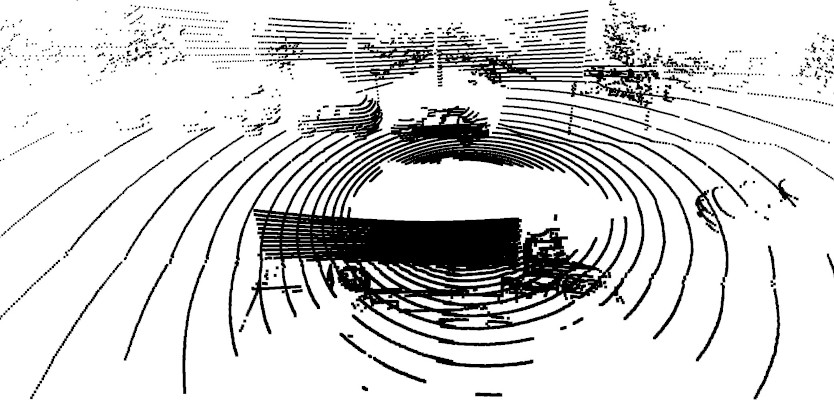}&
            \includegraphics[width=\width]{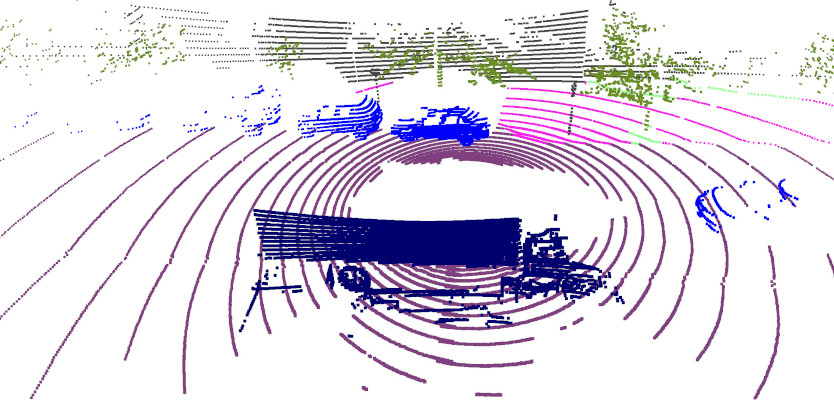}&
            \includegraphics[width=\width]{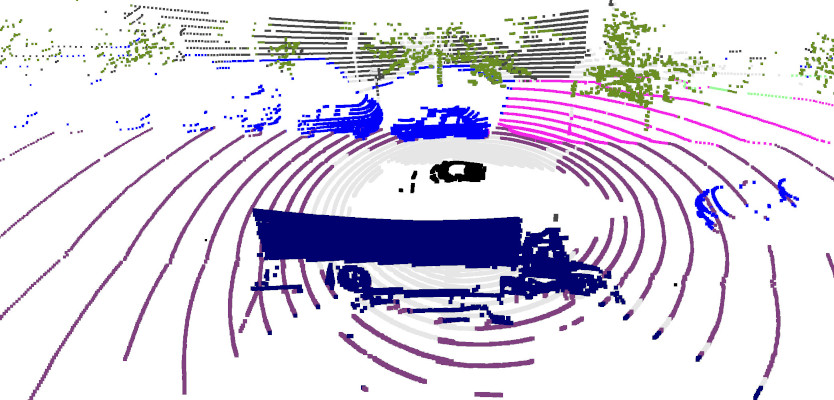}&
            \includegraphics[width=\width]{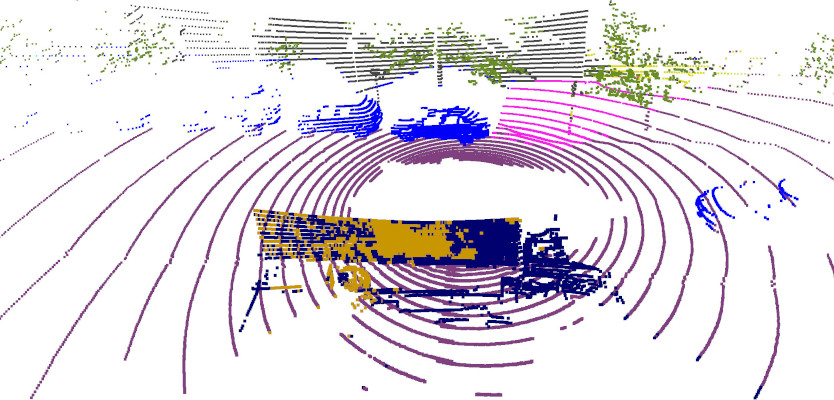}&
            \includegraphics[width=\width]{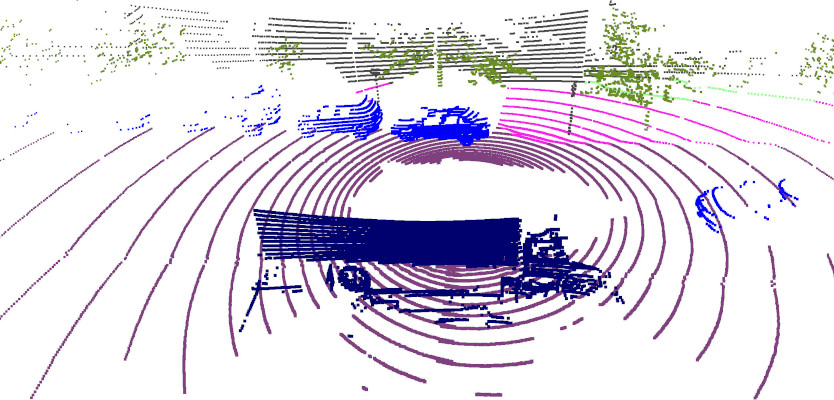}\\
            \\
              {} &{\small Input} &
              {\small Fully supervised~\cite{Choy2019CVPR}} &
              {\small MSeg Voting~\cite{Lambert2020CVPR}} &
              {\small \textbf{Ours}} &
              {\small GT Segmentation}\\
        \end{tabular}
        \caption{
        \textbf{Qualitative comparisons.}  Images of 3D semantic segmentation results on public indoor and outdoor benchmarks.
        }
        \label{fig:benchmark}
\end{figure*}
}

\newcommand{\figproperties}{
\begin{figure*}
  \centering
  \small
  \includegraphics[width=\textwidth]{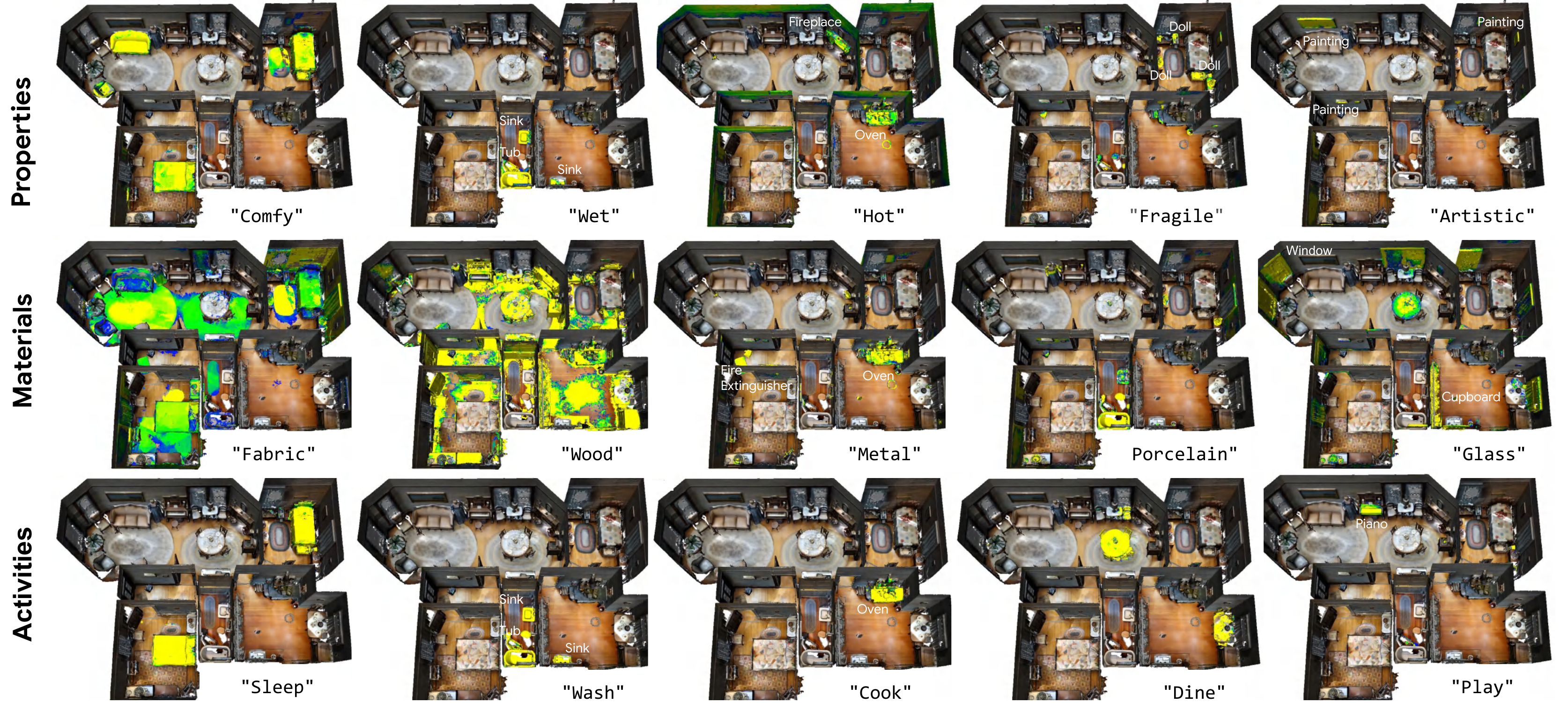}
  \caption{\textbf{Open-vocabulary 3D Scene Exploration}.  Examples of discovering properties, surface materials, and activity sites within a scene using open-vocabulary queries.   For each example, the query text is listed below (e.g., ``Comfy''), and the 3D points are colored based on their cosine similarity to the clip embedding for the query text -- yellow is {\color{newyellow}highest}, green is {\color{green}middle}, blue is {\color{blue}low}, and uncolored is lowest.}
  \label{fig:query_features}
\end{figure*}
}

\newcommand{\figretrieval}{
\begin{figure}[t]
  \centering
  \small
  \includegraphics[width=0.47\textwidth]{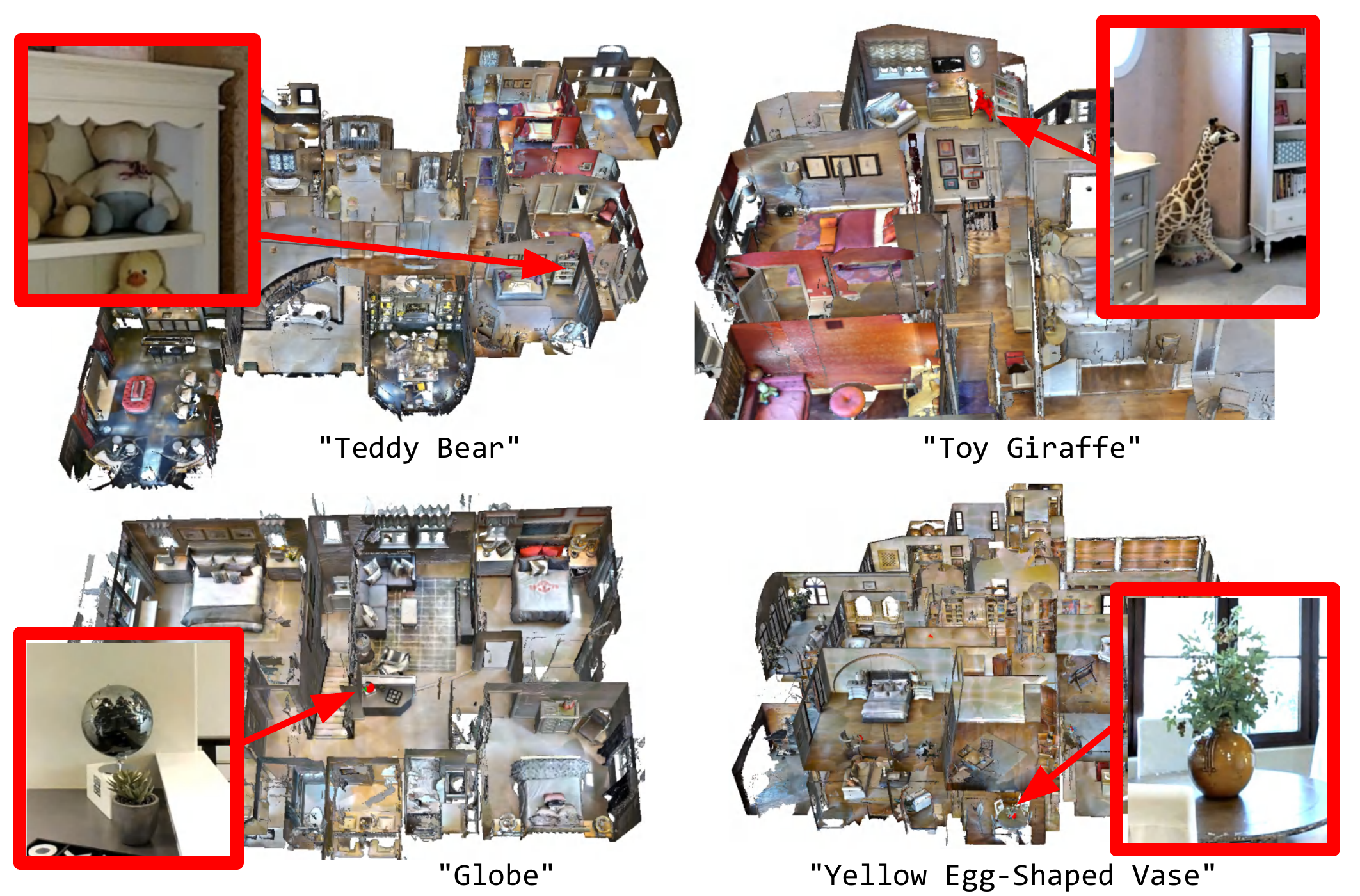}
  \caption{\textbf{Open-vocabulary 3D Search.}  These images show the 3D point within a database of 3D house models that best matches a text query.   The inset image shows a zoomed view of the match.}
  \label{fig:retrieval}
\end{figure}
}

\newcommand{\figcoembedding}{
\begin{figure}[t]
  \centering
  \small
  \includegraphics[width=0.48\textwidth]{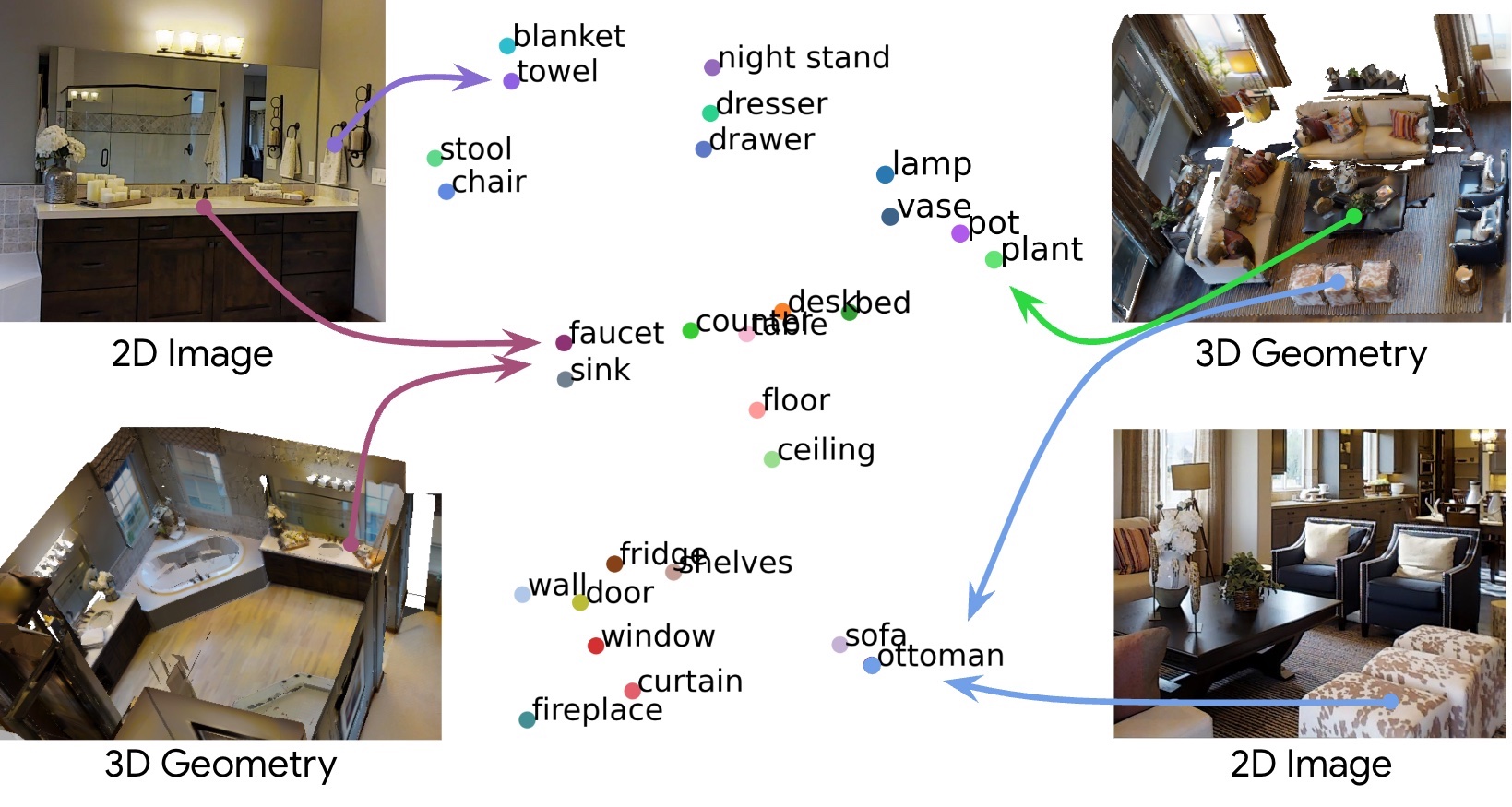}
  \caption{\textbf{Key idea.} We co-embed 3D points with text and image pixels 
  in the CLIP feature space (visualized with T-SNE~\cite{Van2008JMLR}) which has structure learned from large image and text repositories.}
  \label{fig:coembedding}
\end{figure}
}

\newcommand{\figimageretrieval}{
\begin{figure}[t]
  \centering
  \small
  \includegraphics[width=0.47\textwidth]{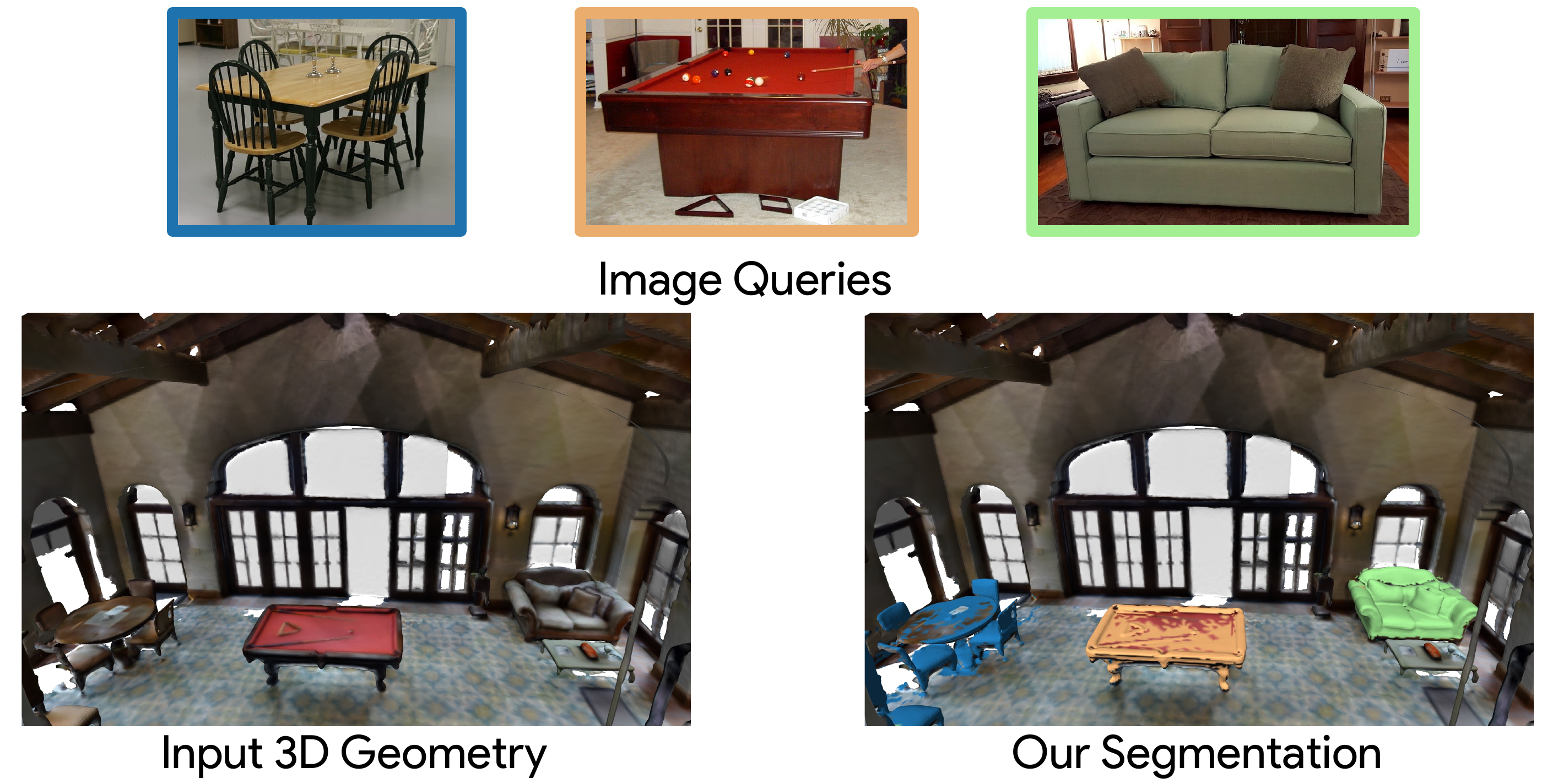}
  \caption{\textbf{Image-based 3D Object Detection.} 
  A 3D scene (bottom left) can be queried with images from Internet (top) to find matching 3D points (bottom right).   The colors of the image query outlines indicate the corresponding matches in the 3D point cloud.
  All 3 images are under Creative Commons licenses.
  }
  \label{fig:img_retrieval}
\end{figure}
}

\newcommand{\ewidth}{0.5\textwidth}
\newcommand{\figensembleratio}{
\begin{figure*}[t]
        \centering
        \vspace{3em}
        \begin{tabular}{cc}
            \includegraphics[width=\ewidth]{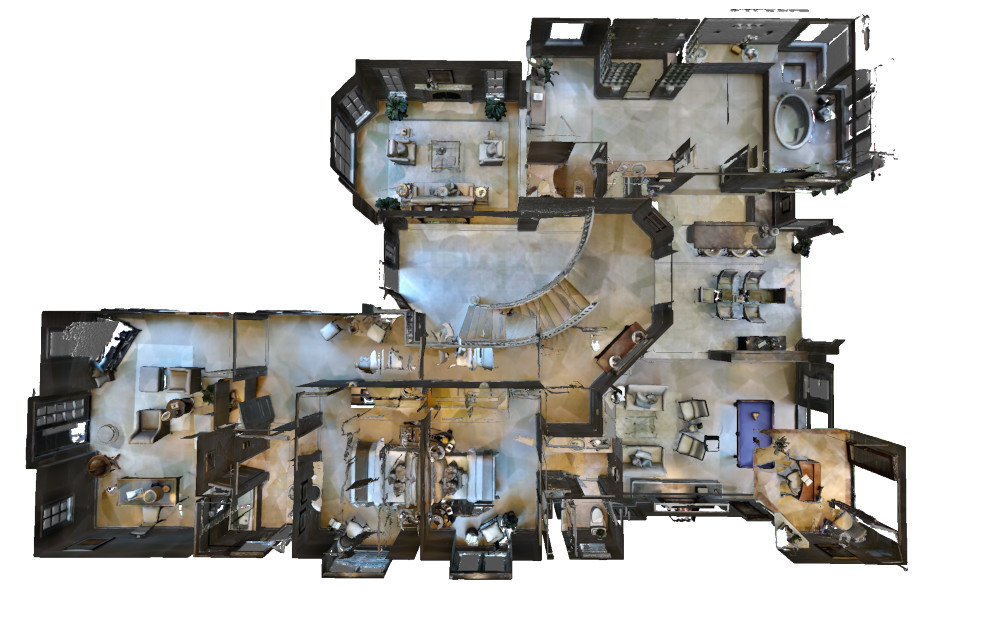}& 
            \includegraphics[width=0.43\textwidth]{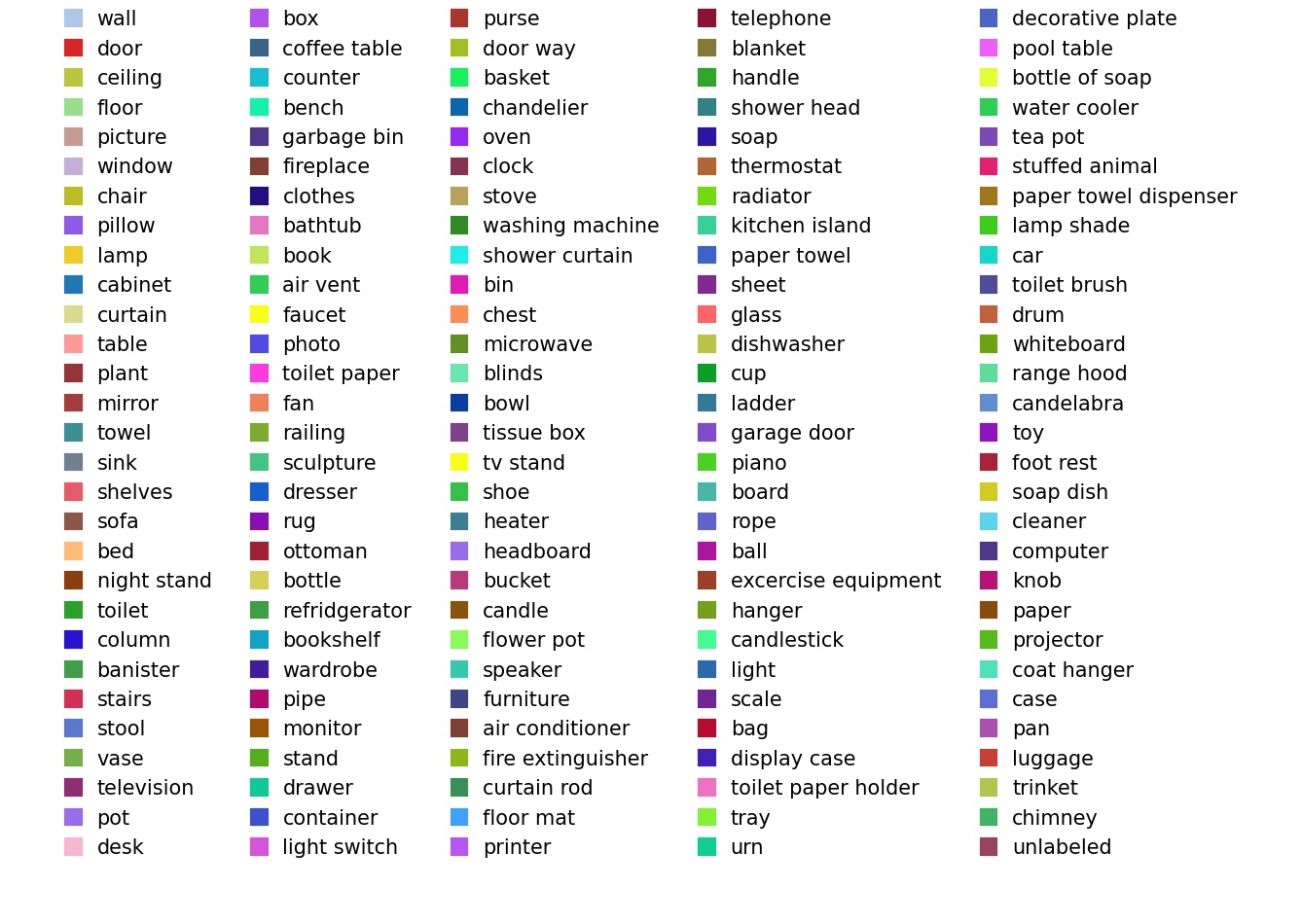}
            \\
            Input 3D Geometry & 160-class label sets \vspace{1em}\\
            \includegraphics[width=\ewidth]{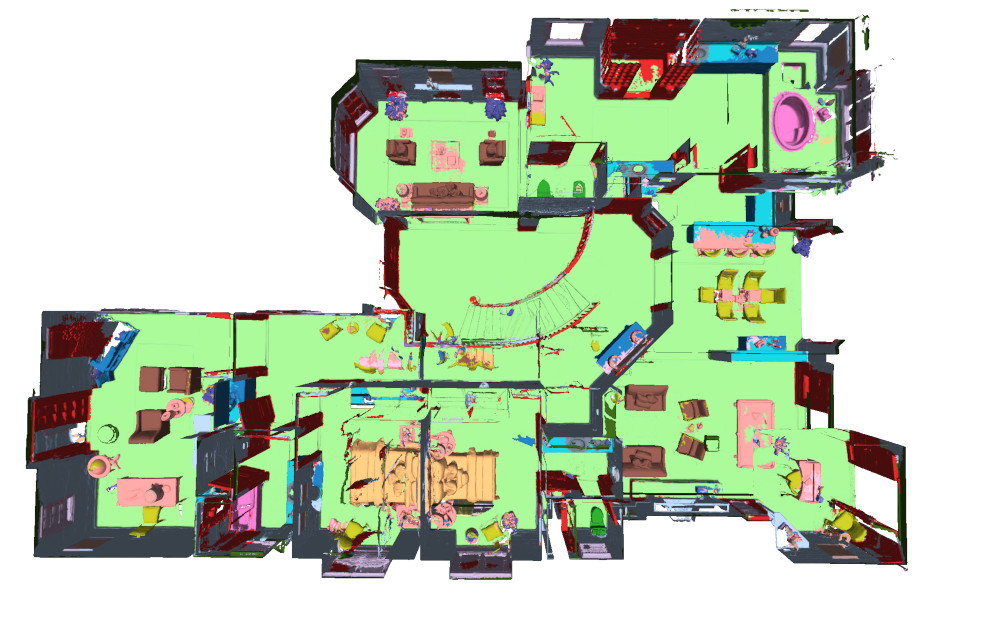} &
            \includegraphics[width=\ewidth]{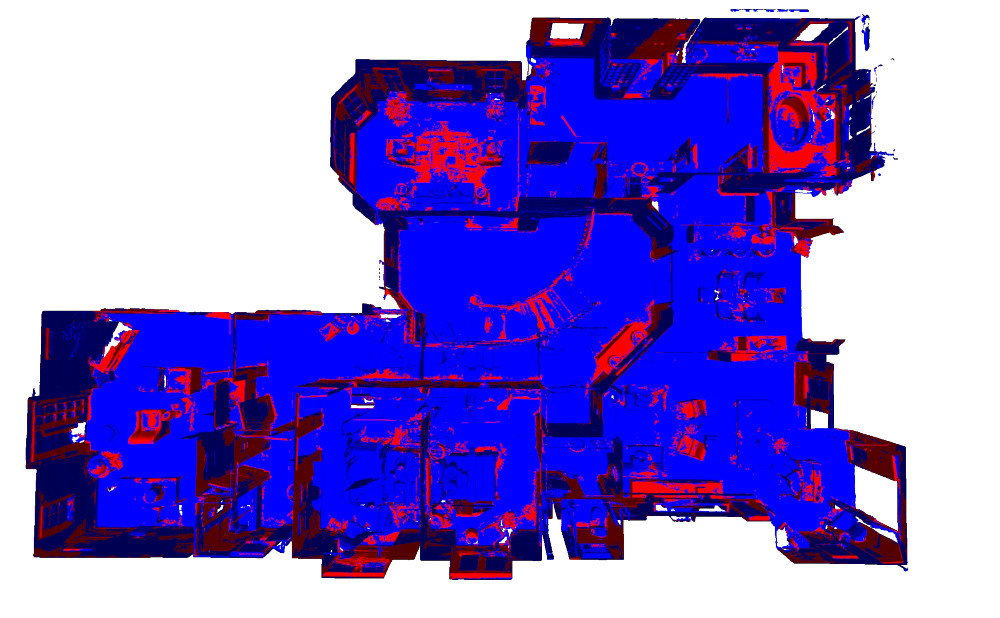}\\
            Our top 21-class semantic segmentation results & {\color{red}2D}-{\color{blue}3D} features selected for 21-class segmentation \vspace{1em}\\
            \includegraphics[width=\ewidth]{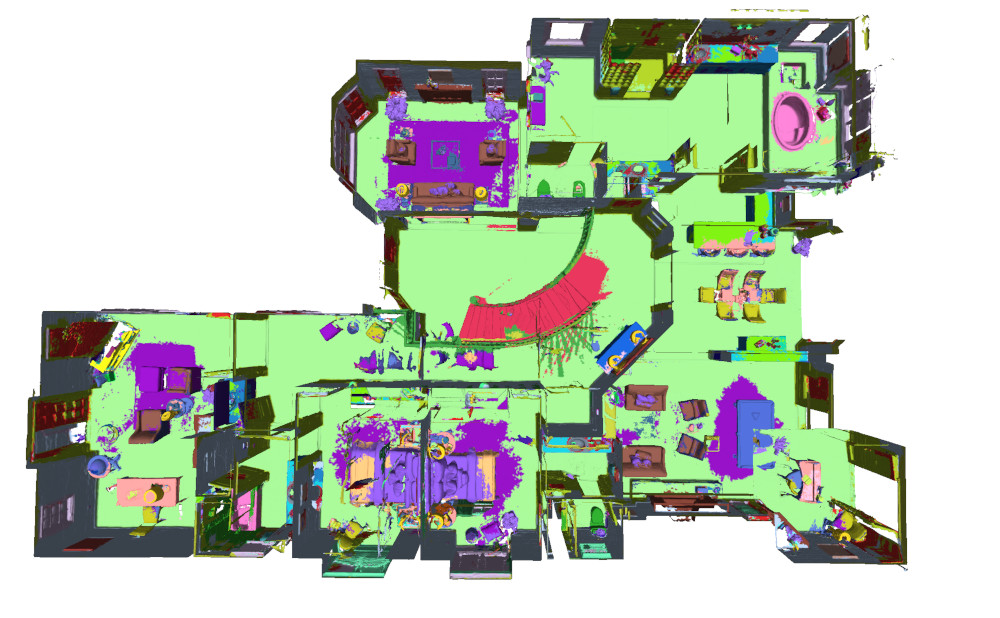} &
            \includegraphics[width=\ewidth]{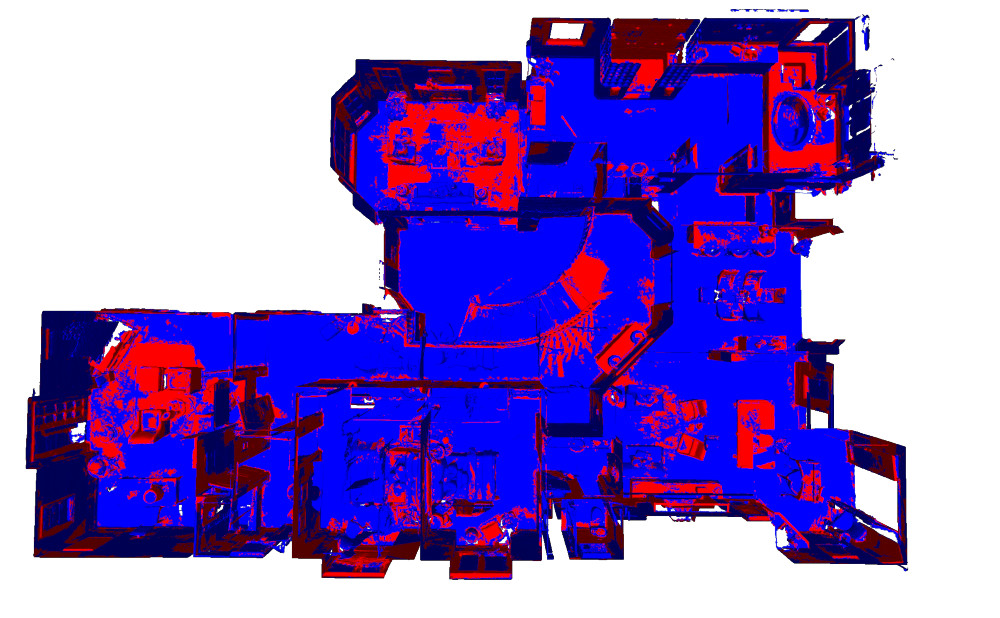}\\
            Our top 160-class semantic segmentation results & {\color{red}2D}-{\color{blue}3D} features selected for 160-class segmentation
            \\
        \end{tabular}
        \vspace{1em}
        \caption{
        \textbf{Study of Our 2D-3D Ensemble Model.}  We show semantic segmentation results and the feature selection of our ensemble model on a Matterport3D house. We show the comparison between the 21-class and 160-class prediction. As can be seen, when the number of classes increases, our ensemble model selects more 2D features for the segmentation. The reason can be that, when involving more fine-grained or long-tailed classes, 2D image features can better understand those fine-grained concept than purely from 3D point clouds.
        Points using 2D features for final segmentation are marked as {\color{red}red}, while points with 3D features are marked as {\color{blue}blue}.
        }
        \label{fig:ensemble_ratio}
\end{figure*}
}

\newcommand{\tableensembleratio}{
\begin{table}[!t]
    \centering
    \footnotesize
    \setlength{\tabcolsep}{0.25cm}
    \resizebox{0.43\textwidth}{!}{

    \begin{tabular}{l|c|c|c|c}
        \toprule
        {} & $K=21$ & $K=40$ & $K=80$ & $K=160$\\
        \midrule
        2D Features &28.56\% & 29.96\% & 31.58\% & 32.46\%\\
        3D Features &71.44\% & 70.04\% & 68.42\% & 67.54\%\\
        \bottomrule
    \end{tabular}}
    \caption{
        \textbf{Behavior of Ensemble Model.}   Each entry indicates the percentage of
        points for which the Ensemble Model selects 2D or 3D features for semantic segmentation on Matterport3D for different numbers of classes $K$ in the labelset.
        }
\label{tab:ensemble_ratio}
\end{table}
}

\newcommand{\tableablationfusion}{
\begin{table}[!t]
    \centering
    \footnotesize
    \setlength{\tabcolsep}{0.5cm}
    \resizebox{0.48\textwidth}{!}{

    \begin{tabular}{l|c|c|c}
        \toprule
        {} & Random & Median & Mean\\
        \midrule
        mIoU &38.2 & 40.1 & \textbf{41.4}\\
        mAcc &60.1 & 62.2 & \textbf{63.6}\\
        \bottomrule
    \end{tabular}}
    \caption{
        \textbf{Ablation on Multi-view Fusion Strategy.} We report mIoU and mAcc on ScanNet~\cite{Dai2017CVPR} with our OpenSeg feature fusion.
        }
\label{tab:ablation_fusion}
\end{table}
}

\newcommand{\tablenuscenesmapping}{
\begin{table}[!t]
    \centering
    \footnotesize
    \setlength{\tabcolsep}{0.25cm}
    \resizebox{0.48\textwidth}{!}{

    \begin{tabular}{l|l}
        \toprule
        nuScenes 16 labels & Our pre-defined labels\\
        \midrule
        barrier & barrier, barricade\\
        bicycle & bicycle\\
        bus & bus\\
        car  & car\\
        construction vehicle  &  \parbox[t]{4cm}{bulldozer, excavator, concrete mixer, crane, dump truck}\\
        motorcycle  & motorcycle\\
        pedestrian  & pedestrian, person\\
        traffic cone  & traffic cone\\
        trailer  & \parbox[t]{4cm}{trailer, semi trailer, cargo container, shipping container, freight container}\\
        truck  & truck\\
        driveable surface  & road\\
        other flat  & \parbox[t]{4cm}{curb, traffic island, traffic median}\\
        sidewalk  & sidewalk\\
        terrain  & \parbox[t]{4cm}{grass, grassland, lawn, meadow, turf, sod}\\
        manmade  & building, wall, pole, awning\\
        vegetation & \parbox[t]{4cm}{tree, trunk, tree trunk, bush, shrub, plant, flower, woods}\\
        \bottomrule
    \end{tabular}}
    \caption{
        \textbf{Label Mappings for nuScenes 16 Classes.} Here we list the total 43 pre-defined non-ambiguous class names corresponding to the 16 nuScenes classes. 
        }
\label{tab:nuscenes_mapping}
\end{table}
}

\newcommand{\tablethreedssg}{
\begin{table}[t]
\centering
\footnotesize
\setlength{\tabcolsep}{0.4cm}
    \resizebox{0.48\textwidth}{!}{
\begin{tabular}{l|c|c}
                      & mIoU & mAcc \\\toprule
MinkowskiNet (Fully supervised)         & \textbf{23.5}     & 30.6     \\\midrule
Ours - 2D Fusion      & 18.6   &31.9      \\
Ours - 3D Model distilled on ScanNet      & 15.3   &26.4      \\
Ours - 2D-3D Ensemble & 20.1   &\textbf{35.6}     \\\bottomrule
\end{tabular}}
\includegraphics[width=0.45\textwidth]{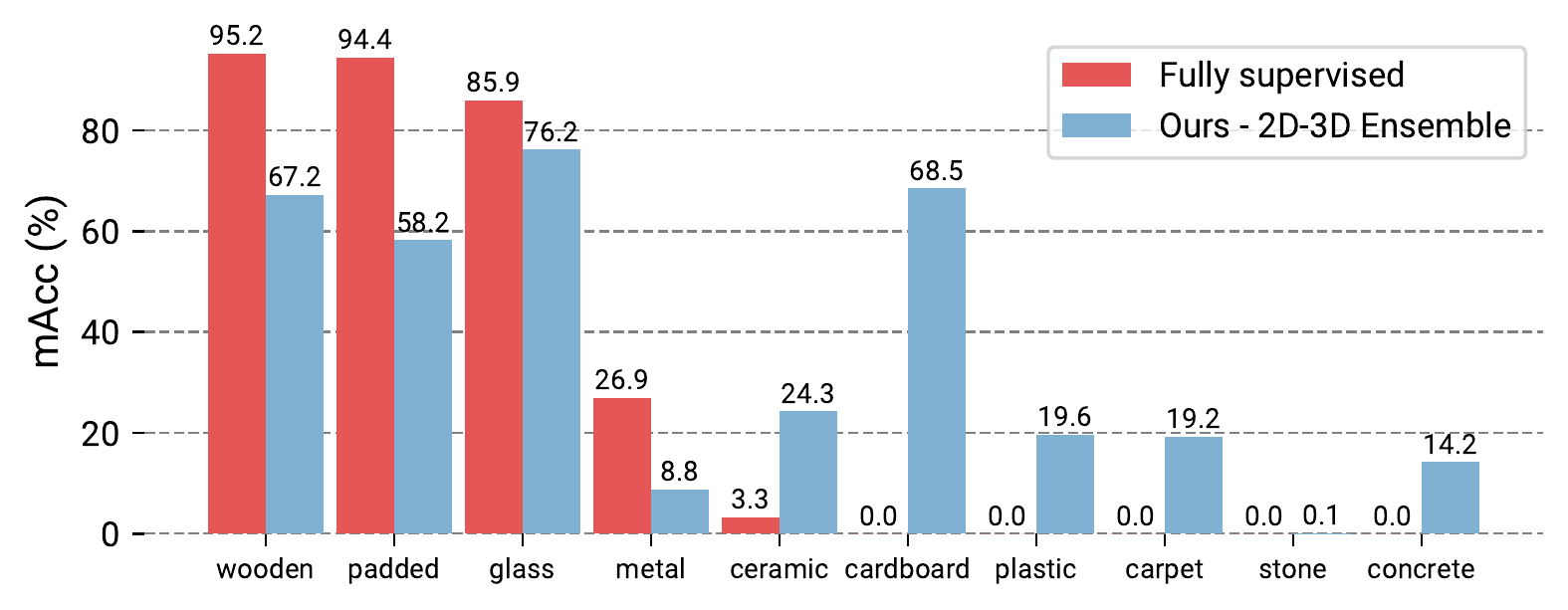}
\caption{
        \textbf{Comparison on 3DSSG~\cite{Wald2020CVPR} in Material Estimation.} We report the average of 10 material classes in  test set. Classes are sorted left-to-right by the number of training examples.
        }
\label{tab:3dssg}
\end{table}
}

\def\cvprPaperID{3937} %
\def\confName{CVPR}
\def\confYear{2023}

\definecolor{turquoise}{cmyk}{0.65,0,0.1,0.3}
\definecolor{purple}{rgb}{0.65,0,0.65}
\definecolor{dark_green}{rgb}{0, 0.5, 0}
\definecolor{orange}{rgb}{0.8, 0.6, 0.2}
\definecolor{red}{rgb}{0.8, 0.2, 0.2}
\definecolor{darkred}{rgb}{0.6, 0.1, 0.05}
\definecolor{blueish}{rgb}{0.0, 0.3, .6}
\definecolor{light_gray}{rgb}{0.7, 0.7, .7}
\definecolor{pink}{rgb}{1, 0, 1}
\definecolor{greyblue}{rgb}{0.25, 0.25, 1}

\definecolor{thirdbestcolor}{rgb}{1,1, 0.6}
\definecolor{secondbestcolor}{rgb}{1, 0.9, 0.6}
\definecolor{firstbestcolor}{rgb}{1, 0.6, 0.6}

\usepackage{enumitem}
\setlist[itemize]{noitemsep,leftmargin=.15in,topsep=.2em}
\setlist[enumerate]{noitemsep,leftmargin=*,topsep=.2em}

\begin{document}

\setlength{\floatsep}{0.50\floatsep}
\setlength{\textfloatsep}{0.50\textfloatsep}
\setlength{\dblfloatsep}{0.50\dblfloatsep}
\setlength{\dbltextfloatsep}{0.50\dbltextfloatsep}
\setlength{\abovedisplayskip}{3pt}
\setlength{\belowdisplayskip}{3pt}

\title{OpenScene: 3D Scene Understanding with Open Vocabularies}

\title{OpenScene: 3D Scene Understanding with Open Vocabularies}
\author{
Songyou Peng$^{1,2,3}$ \qquad Kyle Genova$^{1}$ \qquad Chiyu ``Max'' Jiang$^{4}$\qquad Andrea Tagliasacchi$^{1,5}$\\
Marc Pollefeys$^{2}$ \qquad Thomas Funkhouser$^{1}$ \vspace{0.05cm}\\
{\small $^{1}$ Google Research \quad $^{2}$ ETH Zurich \quad $^{3}$ MPI for Intelligent Systems, Tübingen \quad $^{4}$ Waymo LLC \quad $^{5}$ Simon Fraser University}
\\{\small\href{https://pengsongyou.github.io/openscene}{pengsongyou.github.io/openscene}}
}

\twocolumn[{%
\renewcommand\twocolumn[1][]{#1}%
\maketitle
\vspace{-3.4em}
\begin{center}
    \includegraphics[width=\textwidth]{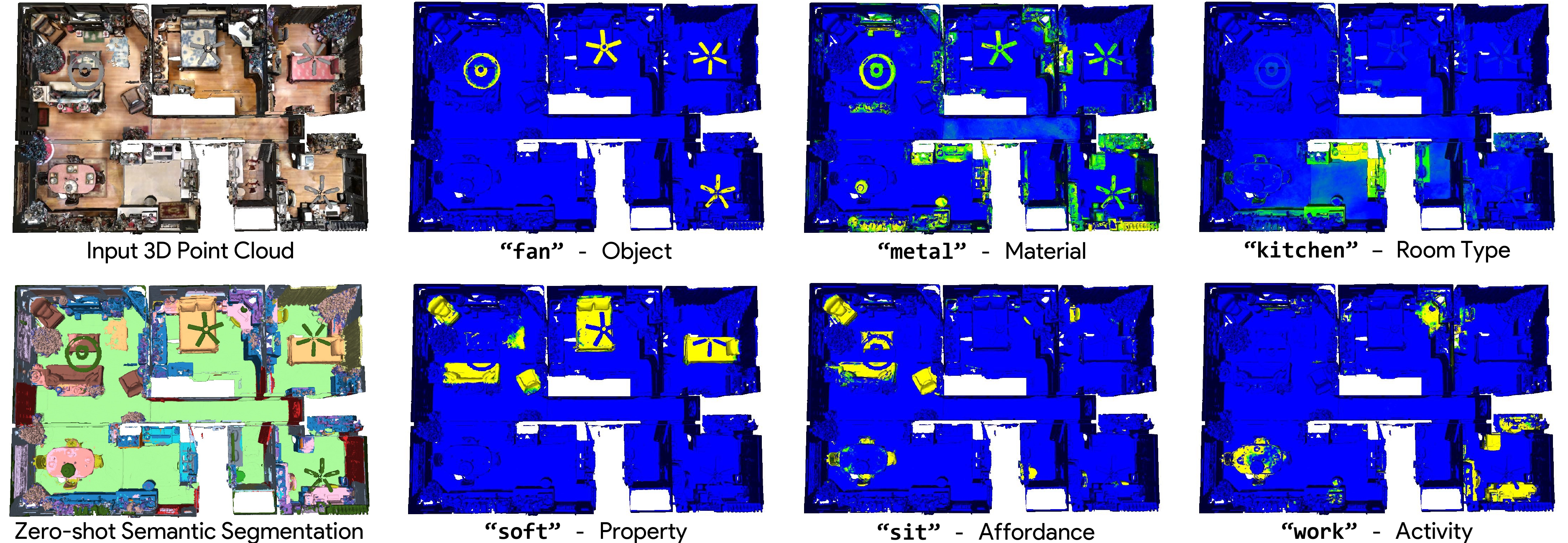}    
    \captionof{figure}{\textbf{Open-vocabulary 3D Scene Understanding.} We propose OpenScene, a zero-shot approach to 3D scene understanding that co-embeds dense 3D point features with image pixels and text.  The examples above show a 3D scene with surface points colored by how well they match a user-specified query string -- yellow is {\color{newyellow}highest}, green is {\color{newgreen}middle}, blue is {\color{blue}low}.  Harnessing the power of language-based features, OpenScene answers a wide variety of example queries, like ``soft'', ``kitchen'', or ``work'', without labeled 3D data.
    }
    \label{fig:teaser}
\end{center}
}]

\begin{abstract}
\vspace*{-0.2cm}

Traditional 3D scene understanding approaches rely on labeled 3D datasets to train a model for a single task with supervision. We propose OpenScene, an alternative approach where a model predicts dense features for 3D scene points that are co-embedded with text and image pixels in CLIP feature space. This zero-shot approach enables task-agnostic training and open-vocabulary queries. For example, to perform SOTA zero-shot 3D semantic segmentation it first infers CLIP features for every 3D point and later classifies them based on similarities to embeddings of arbitrary class labels. More interestingly, it enables a suite of open-vocabulary scene understanding applications that have never been done before. For example, it allows a user to enter an arbitrary text query and then see a heat map indicating which parts of a scene match. Our approach is effective at identifying objects, materials, affordances, activities, and room types in complex 3D scenes, all using a single model trained without any labeled 3D data.
\end{abstract}

\vspace*{-0.6cm}
\section{Introduction}~\label{sec:intro}
3D scene understanding is a fundamental task in
computer vision. 
Given a 3D mesh or point cloud with a set of posed RGB images, the goal is to infer the semantics, affordances,
functions, and physical properties of every 3D point.  
For example, given the house shown in Figure
\ref{fig:teaser}, we would like to predict which surfaces are
part of a fan (semantics), made of metal (materials), 
within a kitchen~(room types), where a person can sit~(affordances),
where a person can work~(functions), 
and which surfaces are soft~(physical properties).  
Answers to these queries can help a
robot interact intelligently with the scene  or help a person
understand it through interactive query and visualization.

Achieving this broad scene understanding goal is challenging due
to the diversity of possible queries.  Traditional 3D scene
understanding systems are
trained with supervision from benchmark datasets designed for
specific tasks (e.g., 3D semantic segmentation for a closed set of
20 classes~\cite{Dai2017CVPR,Chang2017THREEDV}).  They are each designed to answer one type of query
(is this point on a chair, table, or bed?), but
provide little assistance for related queries where training data
is scarce (e.g., segmenting rare objects) or other
queries with no 3D supervision (e.g., estimating material properties).

In this paper, we investigate how to use pre-trained text-image
embedding models (e.g., CLIP~\cite{Radford2021ICML}) to assist in 3D scene understanding.
These models have been trained from 
large datasets of captioned images to co-embed visual and language concepts in a shared feature space.
Recent work has shown that these models can be
used to increase the flexibility and generalizability of 2D image
semantic segmentation~\cite{Ghiasi2022ECCV, Li2022ICLR, Zhou2022ECCV, Rao2022CVPR, Zabari2021ARXIV, Xu2022CVPR}.   
However, nobody has investigated how to use them to improve
the diversity of queries possible for 3D scene understanding.

\figcoembedding

We present \textit{OpenScene}, a simple yet effective zero-shot 
approach for open-vocabulary 3D scene understanding.  Our key idea 
is to compute dense features for 3D points that are co-embedded 
with text strings and image pixels in the CLIP feature space (\figref{fig:coembedding}).
To achieve this, we establish
associations between 3D points and pixels from posed images in the 3D scene, and train a 3D network to embed points using CLIP pixel features as supervision.
This approach brings 3D points in alignment with pixels in the feature space,
which in turn are aligned with text features, and thus enables open vocabulary
queries on the 3D points.

Our 3D point embedding algorithm includes both 2D and 3D convolutions.
We first back-project the 3D position of the point into every image
and aggregate the features from the associated pixels using multi-view fusion.
Next, we train a sparse 3D convolutional network to perform feature extraction 
from only the 3D point cloud geometry with a loss that minimizes differences to the aggregated pixel features.
Finally, we ensemble the features produced by the 2D fusion and the 3D network into
a single feature for each 3D point.  This hybrid 2D-3D feature
strategy enables the algorithm to take advantage of salient patterns in both 
2D images and 3D geometry, and thus is 
more robust and descriptive than features from either domain alone.

Once we have computed features for every 3D point, 
we can perform a variety of 3D scene understanding queries.
Since the CLIP model is trained with natural language
captions, it captures concepts beyond object class labels,
including affordances, materials, attributes, and functions~(\figref{fig:teaser}).
For example, computing the similarity of 3D features with the embedding for ``soft'' produces the result shown in the bottom-left image of \figref{fig:teaser}, which highlights the couches, beds, and comfy chairs as the best matches.

Since our approach is zero-shot~(i.e. no use of labeled data for the target task), it does not perform as well as fully-supervised approaches on the limited set of tasks for which there is sufficient training data in traditional benchmarks (e.g., 3D semantic segmentation with 20 classes).
However, it does achieve significantly stronger performance on other tasks.  For example, it beats a fully-supervised approach on indoor 3D semantic segmentation with 40, 80, or 160 classes.
It also performs better than other zero-shot baselines, and can be used without any retraining on novel datasets even if they have different label sets. It works for indoor RGBD scans as well as outdoor driving captures.

\noindent
Overall, our contributions are summarized as follows:
\begin{itemize}
    \item We introduce open vocabulary 3D scene understanding tasks where arbitrary text
    queries are used for semantic segmentation, affordance estimation, room type classification, 
    3D object search, and 3D scene exploration.
    \item We propose OpenScene, a zero-shot method for extracting 3D dense features from an open vocabulary embedding space using multi-view fusion and 3D convolution. 
    \item We demonstrate that the extracted features can be used for 3D semantic segmentation with performance better than fully supervised methods for rare classes.
\end{itemize}

\section{Related Work}~\label{sec:related}
This paper draws on a large literature of previous work on 3D scene understanding, multi-modal embedding, and zero-shot learning. 

\boldparagraph{Closed-set 3D Scene Understanding}
There is a long history of work on 3D scene understanding
for vision and robotics applications.
Most prior work focuses on training models with ground-truth 3D labels~\cite{Hu2021CVPR,Wang2017SIGGRAPH,Choy2019CVPR,Qi2017NIPS,Robert2022CVPR,Schult2020CVPR,Huang2019CVPR,Hu2021ICCV,Li2022CVPR,Han2020CVPR,Nekrasov2021THREEDV}. These works have yielded network architectures and training protocols that have significantly pushed the boundary of several 3D scene understanding benchmarks, including 3D object classification~\cite{Wu2015CVPR}, 3D object detection and localization~\cite{Geiger2012CVPR,Caesar2020CVPR,Sun2020CVPR,Chen2020ECCV}, 3D semantic and instance segmentation~\cite{Dai2017CVPR,Chang2017THREEDV,Liao2022PAMI,Behley2019ICCV,Hua2016THREEDV}, 3D affordance prediction \cite{Deng2021CVPR,Li2019CVPR,Wang2021CVPR}, and so on.   The most closely related work to ours of this type is
Rozenberszki~\etal~\cite{Rozenberszki2022ARXIV}, since they use the CLIP embedding to pre-train a model for 3D semantic segmentation.  However, they only use the text embedding for point encoder pretraining, and then train the point decoder with 3D GT annotations afterwards.   Their focus is on using the CLIP embedding to achieve
better supervised 3D semantic segmentation, rather than open-vocabulary queries.

Another line of research performs 3D scene understanding experiments with only 2D ground truth supervisions~\cite{Genova2021THREEDV,Kundu2020ECCV,Mccormac2017ICRA,Sautier2022CVPR,Vineet2015ICRA,Wang2022CORL}.
For example, \cite{Genova2021THREEDV} generates pseudo 3D annotation by backprojecting and fusing the 2D predicted labels, from which they learn the 3D segmentation task.
However, their 2D network is trained with ground truth 2D labels.
A couple of works~\cite{Sautier2022CVPR,Liu2021ICCV} pretrain the 3D segmentation network using point-pixel pairs via contrastive learning between 2D and 3D features.
We also utilize 2D image features as our pseudo-supervision when training the 3D network and no 2D labels are needed.

All these approaches have mainly been applied with small predefined labelsets containing common object categories.
They do not work as well when the number of object categories increases, as tail classes have few training examples.
In contrast, we are able to segment with arbitrary labelsets without any re-training, and we show strong ability of understanding different contents, ranging from rare object types to even materials or physical properties, which is impossible for previous methods.

\boldparagraph{Open-Vocabulary 2D Scene Understanding}
The recent advances of large visual language models~\cite{Radford2021ICML,Alayrac2022ARXIV,Jia2021ICML} have enabled a remarkable level of robustness in zero-shot 2D scene understanding tasks, including recognizing long-tail objects in images.
However, the learned embeddings are often at the image level, thus not applicable for dense prediction tasks requiring pixel-level information.
Many recent efforts~\cite{Ghiasi2022ECCV,Li2022ICLR,Zhou2022ECCV,Rao2022CVPR,Zabari2021ARXIV,Xu2022CVPR,Gu2022ICLR,Kuo2022ARXIV,Ma2022BMVC,Liang2022ARXIV,Shafiullah2022ARXIV} attempt to correlate the dense image features with the embedding from large language models. 
In this way, given an image at test time, users can define arbitrary text labels to classify, detect, or segment the image. 

More recently, Ha and Song~\cite{Ha2022CORL} take a step forward and perform open-vocabulary partial scene understanding and completion given a single RGB-D frame as input.
This method is limited to small partial scenes and requires ground truth training data for supervision. 
In contrast, in this work, we solely rely on pretrained open-vocabulary 2D models and perform a series of 3D scene-level understanding tasks, without the need for any ground truth training data in 2D or 3D. 
Moreover,in the absence of 2D images, our method can perform 3D-only open-vocabulary scene understanding tasks based on a 3D point network distilled from an open-vocabulary 2D image model through 3D fusion.

\boldparagraph{Zero-shot Learning for 3D Point Clouds}
While there have been a number of studies on zero-shot learning for 2D images, their application to 3D is still recent and scarce.
A handful of works~\cite{Cheraghian2019MVA,Cheraghian2019BMVC,Cheraghian2020WACV,Cheraghian2022IJCV,Zhang2022CVPRb} attempt to address the 3D point classification and generation tasks.
More recently, \cite{Liu2021ARXIV,Michele2021THREEDV} investigated
zero-shot learning for semantic segmentation for 3D point clouds.
They train with supervision of 3D ground truth labels for a predefined set of seen classes and then evaluate on new unseen classes.  
However, these methods are still limited to the closed-set segmentation setting and still require GT training data for the majority of the 3D dataset.
Our method \textit{does not} require any labeled 3D data for training, and it handles a broad range of queries supported by a large language model.

\section{Method}~\label{sec:method}
\begin{figure}[!t]
  \centering
  \includegraphics[width=0.48\textwidth]{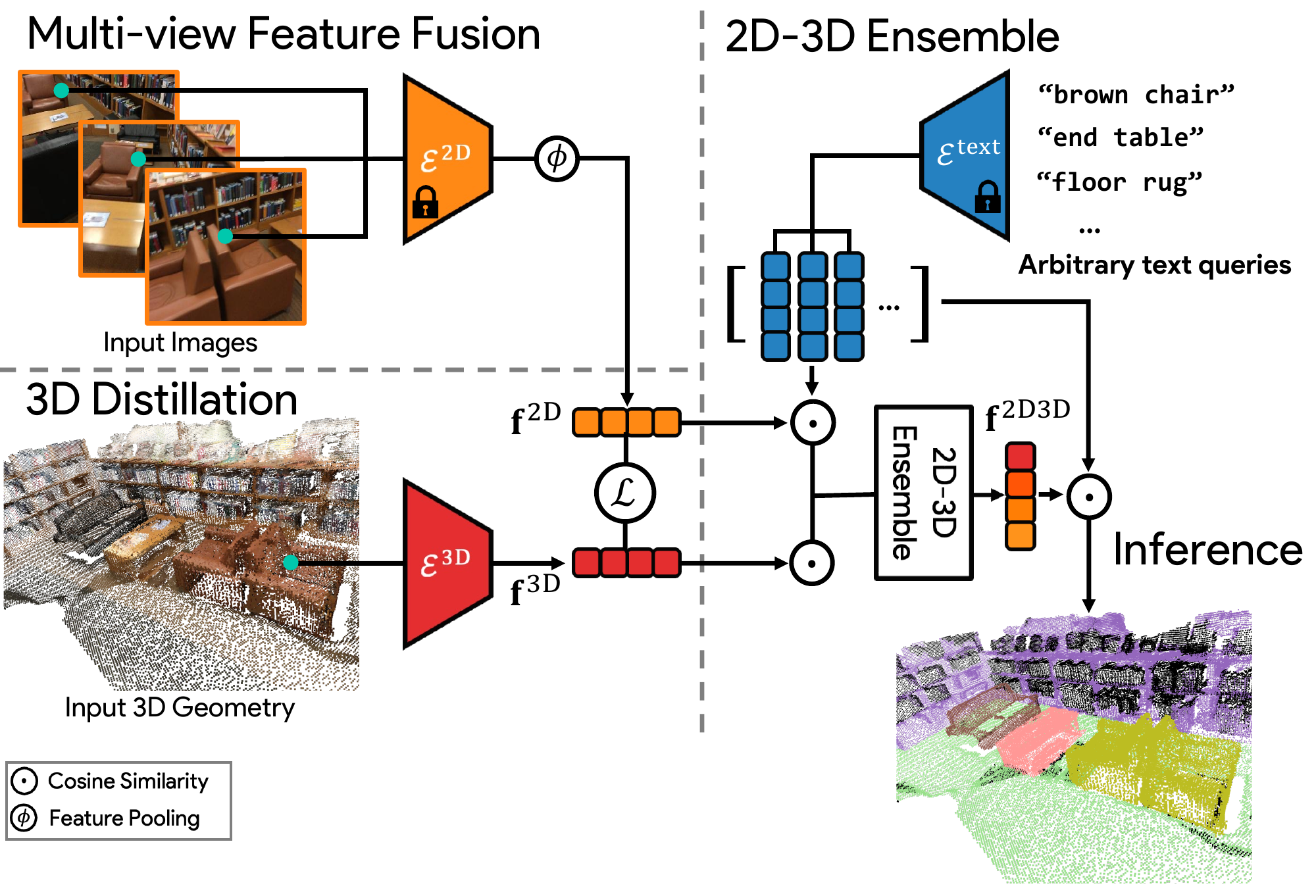}
  \caption{\textbf{Method  Overview.}  Given a 3D model (mesh or point cloud) and a set of posed images, we train a 3D network $\cE^\text{3D}$ to produce dense features for 3D points $\bff^\text{3D}$ with a distillation loss $\cL$ to multi-view fused features $\bff^\text{2D}$ for projected pixels. 
  We ensemble $\bff^\text{2D}$ and $\bff^\text{3D}$ based on cosine similarities to CLIP embeddings for an arbitrary set of queries to form $\bff^\text{2D3D}$.
  During inference, we can use the similarity scores between per-point features and given CLIP features to perform open-vocabulary 3D scene understanding tasks.
  }
  \label{fig:overview}
\end{figure}

An overview of our approach is illustrated in~\figref{fig:overview}.   We first compute per-pixel features for every image using a model pre-trained for open-vocabulary 2D semantic segmentation.   We then aggregate the pixel features from multiple views onto every 3D point to form a per-point fused feature vector~\secref{sec:fusion}.
We next distill a 3D network to reproduce the fused features using only the 3D point cloud as input~\secref{sec:distill}.  Next, we ensemble the fused 2D features and distilled 3D features into a single per-point feature~\secref{sec:ensemble} and use it to answer open-vocabulary queries~\secref{sec:inference}.

\subsection{Image Feature Fusion}
\label{sec:fusion}
The first step in our approach is to extract dense per-pixel embeddings for each RGB image from a 2D visual-language segmentation model, and then back-project them onto the 3D surface points of a scene.

\boldparagraph{Image Feature Extraction}
 Given RGB images with a resolution of $H\times W$, we can simply compute the per-pixel embeddings from the (frozen) segmentation model $\cE^\text{2D}$, denoted as $\bI_i\in\nR^{H\times W\times C}$, where $C$ is the feature dimension, and $i$ is the index spanning the total number of images.
 For~$\cE^\text{2D}$, we experiment with two pretrained image segmentation models OpenSeg~\cite{Ghiasi2022ECCV} and LSeg~\cite{Li2022ICLR}.

\boldparagraph{2D-3D Pairing}
Given a 3D surface point $\bp\in\nR^{3}$ in the point clouds $\bP \in\nR^{M\times 3}$ of a scene with $M$ points,
we calculate its corresponding pixel $\bu=(u, v)$ when the intrinsic matrix $I_i$ and world-to-camera extrinsic matrix $E_i$ of that frame $i$ are provided. 
We only consider the pinhole camera model in this paper so the projection can be represented as $\tilde{\bu} = I_i \;\cdot E_i \;\cdot \tilde{\bp}$, where $\tilde{\bu}$ and $\tilde{\bp}$ are the homogeneous coordinates of $\bu$ and $\bp$, respectively.
Note that for indoor datasets like ScanNet and Matterport3D where the depth images are provided, we also conduct occlusion tests to guarantee the pixel $\bu$ are only paired with a visible surface point $\bp$.

\boldparagraph{Fusing Per-Pixel Features}
With the 2D-3D pairing, the corresponding 2D features in frame $i$ for point $\bp$ can be written as $\bff_i = \bI_i(\bu)\in\nR^{C}$.
Now, assume a total number of $K$ views can be associated with point $\bp$, we can then fuse such 2D pixel embeddings to obtain a single feature vector for this point $\bff^{\text{2D}} = \phi(\bff_1, \cdots, \bff_K)$, where $\phi:\mathbb{R}^{K\times C}\mapsto\mathbb{R}^C$ is an average pooling operator for multi-view features.
An ablation study on different fusion strategies are discussed in supplemental.
After repeating the fusion process for each point, we can build a feature point cloud $\bF^{\text{2D}} = \{\bff_1^{\text{2D}}, \cdots, \bff_M^{\text{2D}}\}\in\nR^{M\times C}$.

\begin{table*}[!ht]
    \centering
    \footnotesize
    \setlength{\tabcolsep}{0.3cm}
    \resizebox{\textwidth}{!}{
        \begin{tabular}{l|ccccc|ccccc}
    \toprule
     &  \multicolumn{5}{c|}{mIoU} & \multicolumn{5}{c}{mAcc}\\ 
    {} & Bookshelf & Desk & Sofa & Toilet & Mean & Bookshelf & Desk & Sofa & Toilet & Mean\\
    \midrule
    3DGenZ~\cite{Michele2021THREEDV} &  6.3& 3.3&13.1&8.1 & 7.7 & 13.4 & 5.9 & 49.6 & 26.3 & 23.8\\ 
    MSeg Voting &  47.8 & 40.3 & 56.5 &68.8 & 53.4 & 50.1 & 67.7 & 69.8 & 81.0 & 67.2\\ 
    \textbf{Ours} - LSeg & \textbf{67.1} & \textbf{46.4} & \textbf{60.2} & \textbf{77.5} & \textbf{62.8} & \textbf{85.5} & 69.5 & 79.0 & 90.0 & 81.0\\
    \textbf{Ours} - OpenSeg &64.1 & 27.4 & 49.6 & 63.7 & 51.2 & 73.7 & \textbf{73.4} & \textbf{92.5} & \textbf{95.3} & \textbf{83.7}\\
    \bottomrule
\end{tabular}}
    \caption{
        \textbf{Comparison on Zero-shot 3D Semantic Segmentation.}
        We show quantitative comparison between our method and the most recent zero-shot 3D segmentation approach~\cite{Michele2021THREEDV} and a multi-view fusion baseline utilizing MSeg \cite{Lambert2020CVPR}. 
        Following~\cite{Michele2021THREEDV}, we take 4 classes (bookself, desk, sofa, toilet) out of 20 classes from ScanNet validation set for evaluation.
        Unlike \cite{Michele2021THREEDV}, which requires training on 16 seen classes, our approach does not train with any 2D or 3D ground labels on any classes. Still, both of our variants show significantly better performance in both mIoU and mAcc.        
        }
\label{tab:zero_shot}
\end{table*}

\subsection{3D Distillation}
\label{sec:distill}
The feature cloud $\bF^{\text{2D}}$ can be directly used for language-driven 3D scene understanding \textit{when} images are present.
Nevertheless, such fused features could lead to noisy segmentation due to potentially inconsistent 2D predictions.
Moreover, some tasks only provide 3D point clouds or meshes.
Therefore, we can distill such 2D visual-language knowledge into a 3D point network that only takes 3D point positions as input.

Specifically, given an input point cloud $\bP$, we seek to learn an encoder that outputs per-point embeddings:
\begin{equation}
    \bF^\text{3D} = \cE^\text{3D}(\bP), \quad \cE^\text{3D}:\nR^{M\times 3}\mapsto\nR^{M\times C}
\end{equation}
where $\bF^\text{3D} = \{\bff_1^{\text{3D}}, \cdots, \bff_M^{\text{3D}}\}$. To enforce the output of the network $\bF^\text{3D}$ to be consistent with the fused features $\bF^\text{2D}$, we use a cosine similarity loss:
\begin{equation}
    \cL = 1 - \text{cos}(\bF^\text{2D}, \bF^\text{3D})
    \label{eq:loss}
\end{equation}
We use MinkowskiNet18A~\cite{Choy2019CVPR} as our 3D backbone $\cE^\text{3D}$, and change the dimension of outputs  to $C$.

Since the open-vocabulary image embeddings from~\cite{Li2022ICLR,Ghiasi2022ECCV} are co-embedded with CLIP features, the output of our distilled 3D model naturally lives in the same embedding space as CLIP.
Therefore, even without any 2D observations, such text-3D co-embeddings $\bF^\text{3D}$ allow 3D scene-level understanding given arbitrary text prompts.
We show such results in the ablation study in~\secref{sec:ablations}.

\subsection{2D-3D Feature Ensemble}
\label{sec:ensemble}
Although one can already perform open-vocabulary queries with the 2D fused features~$\bF^\text{2D}$ or 3D distilled features~$\bF^\text{3D}$, here we introduce a 2D-3D ensemble method to obtain a hybrid feature to yield better performance.

The inspiration comes from the observation that 2D fused features specialize in predicting small objects (\eg a mug on the table) or ones with ambiguous geometry (e.g., a painting on the wall), while 3D features yield good predictions for objects with distinctive shapes (e.g. walls and floors).  We aim to combine the best of both.

Our ensemble method leverages a set of text prompts, either provided at inference or offline (e.g. predefined classes from public benchmarks like ScanNet, or arbitrary classes defined by users).   We first compute the embeddings for all the text prompts using the CLIP~\cite{Radford2021ICML} text encoder $\cE^\text{text}$, denoted as $\bT=\{\bt_1, \cdots, \bt_N\} \in\nR^{N \times C}$, where $N$ is the number of text prompts and $C$ the feature dimension.
Next, for each 3D point, we obtain its 2D fused and 3D distilled embeddings $\bff^\text{2D}$ and $\bff^\text{3D}$ (dropping the subscript for simplicity).  We can now correlate text features with these two sets of features via cosine similarity, respectively: 
\begin{equation}
    \bs^\text{2D}_n = \text{cos}(\bff^\text{2D}, \bt_n), \qquad
    \bs^\text{3D}_n = \text{cos}(\bff^\text{3D}, \bt_n)
    \label{eq:cosine_similarity}
\end{equation}
Once having the similarity scores \wrt every text prompt $\bt_n$, we can use the max value $\bs^\text{2D} = \max_{n}(\bs^\text{2D}_n)$ and~$\bs^\text{3D} = \max_{n}(\bs^\text{3D}_n)$ among all $N$ prompts as the \emph{ensemble scores} for both features. Our final 2D-3D ensemble feature $\bff^\text{2D3D}$ is simply the feature with the highest ensemble score.

\subsection{Inference}
\label{sec:inference}
With any per-point feature described in the previous subsections ($\bff^\text{2D}$, $\bff^\text{3D}$, or $\bff^\text{2D3D}$) and CLIP features from an arbitrary set of text prompts, we can estimate their similarities by simply calculating the cosine similarity score between them.  We use this similarity score for all of our scene understanding tasks. 
For example, for the zero-shot 3D semantic segmentation using 2D-3D ensemble features, the final segmentation for each 3D point is computed point-wise by $\argmax_{n}\{ \text{cos}(\bff^\text{2D3D}, \bt_n)\}$.

\section{Experiments}~\label{sec:experiments}
We ran a series of experiments to test how well the proposed methods work for a variety of 3D scene understanding tasks. We start by evaluating on traditional closed-set 3D semantic segmentation benchmarks (in order to be able to compare to previous work), and later demonstrate the more novel and exciting open-vocabulary applications in the next section.

\boldparagraph{Datasets}
To test our method in a variety of settings, we evaluate on three popular public benchmarks: ScanNet~\cite{Dai2017CVPR,Rozenberszki2022ARXIV}, Matterport3D~\cite{Chang2017THREEDV}, and nuScenes Lidarseg~\cite{Caesar2020CVPR}.  
These three datasets span a broad gamut of situations -- the first two provide RGBD images and 3D meshes of indoor scenes, and the last provides Lidar scans of outdoor scenes.  We use all three datasets to compare to alternative methods.  Moreover, Matterport3D is a complex dataset with highly detailed scenes, and thus provides the opportunity to stress open-vocabulary queries.

\subsection{Comparisons}
\boldparagraph{Comparison on zero-shot 3D semantic segmentation}
We first compare our approach to the most closely related work on zero-shot 3D semantic segmentation: MSeg~\cite{Lambert2020CVPR} Voting and 3DGenz~\cite{Michele2021THREEDV}.   MSeg Voting predicts a semantic segmentation for each posed image using MSeg~\cite{Lambert2020CVPR} with mapping to the corresponding label sets. For each 3D point, we perform majority voting of the \emph{logits} from multi-view images.  3DGenZ~\cite{Michele2021THREEDV} divides the 20 classes of the ScanNet dataset into 16 seen and 4 unseen classes, and trains a network utilizing the ground truth supervision on the seen classes to generate features for both sets.   

Following the experimental setup in ~\cite{Michele2021THREEDV}, we report the mIoU and mAcc values on their 4 unseen classes in~\tabref{tab:zero_shot}.  Our results on those classes is significantly better than~\cite{Michele2021THREEDV} (7.7\% vs 62.8\% mIoU), even though 3DGenz~\cite{Michele2021THREEDV} utilizes ground truth data for 16 seen classes and ours does not.  
We also outperform MSeg Voting.  In this case, the difference is mainly because our method (regress CLIP features and then classify) naturally models the similarities and differences between classes, where as the MSeg Voting approach (classify and then vote) treats every class as equally distinct from all other classes (a couch and a love seat are just as different as a couch and an airplane in their model).

\figbenchmark

\begin{table}[!t]
    \centering
    \footnotesize
    \setlength{\tabcolsep}{0.1cm}
    \resizebox{0.47\textwidth}{!}{
        
\begin{tabular}{l|cc|cc|cc}
    \toprule
    {} & \multicolumn{2}{c|}{nuScenes~\cite{Caesar2020CVPR}} & \multicolumn{2}{c|}{ScanNet~\cite{Dai2017CVPR}} & \multicolumn{2}{c}{Matterport~\cite{Chang2017THREEDV}}\\
    \midrule
    {} &mIoU & mAcc& mIoU & mAcc& mIoU & mAcc\\\hline
    \multicolumn{4}{l}{\textit{Fully-supervised methods}} \\
    \rowcolor{LightGrey}
    TangentConv~\cite{Tatarchenko2018CVPR} &- & - &40.9 & - & - & 46.8\\ 
    \rowcolor{LightGrey}
    TextureNet~\cite{Huang2019CVPR} &- & - &54.8 & - & -& 63.0\\
    \rowcolor{LightGrey}
    ScanComplete~\cite{Dai2018CVPR} &- & - & 56.6& - & -&44.9\\
    \rowcolor{LightGrey}
    DCM-Net~\cite{Schult2020CVPR} &- & - &65.8& - & - & 66.2\\
    \rowcolor{LightGrey}
    Mix3D~\cite{Nekrasov2021THREEDV} & - & - &\textbf{73.6} & -& -& -\\
    \rowcolor{LightGrey}
    VMNet~\cite{Hu2021ICCV} &- & - &73.2 & - & - &\textbf{67.2}\\
    \rowcolor{LightGrey}
    LidarMultiNet~\cite{Ye2022ARXIV} & \textbf{82.0} & - & - & - & - & -\\ 
    MinkowskiNet~\cite{Choy2019CVPR} & 78.0 & 83.7 & 69.0 & 77.5& 54.2 & 64.6\\ \hline
    \multicolumn{4}{l}{\textit{Zero-shot methods}} \\
    MSeg~\cite{Lambert2020CVPR} Voting & 31.0 & 36.9 & 45.6 & 54.4 & 33.4 & 39.0 \\
    \textbf{Ours} - LSeg  & 36.7 & 42.7 & \textbf{54.2} & 66.6 & \textbf{43.4} & 53.5\\ 
    \textbf{Ours} - OpenSeg  & \textbf{42.1} & \textbf{61.8} &47.5 & \textbf{70.7}& 42.6 & \textbf{59.2}\\
    \bottomrule
\end{tabular}}
    \caption{
        \textbf{Comparisons on Indoor and Outdoor Benchmarks.}
        We compare our method with both zero-shot and fully-supervised baselines for semantic segmentation of one outdoor dataset (nuScenes) and two indoor datasets (ScanNet and Matterport). 
        Note that our zero-shot method performs worse than SOTA approaches trained on this data, but comparable to supervised approaches from a few years ago, and better than the previous SOTA zero-shot approach.  Except for~\cite{Choy2019CVPR}, the numbers for fully-supervised methods (in gray) are taken from previous papers.  
        }
\label{tab:benchmark}
\end{table}

\begin{table}[!t]
    \vspace{0.2cm}
    \centering
    \footnotesize
    \setlength{\tabcolsep}{0.12cm}
        \begin{tabular}{l|c|c|c|c}
    \toprule
    {} & $K=21$ & $K=40$ & $K=80$ & $K=160$\\
    \midrule
    Fully-supervision~\cite{Choy2019CVPR} & \textbf{64.5} & 50.8&33.4&18.4\\ \hline
    \textbf{Ours} &59.2 & \textbf{50.9} & \textbf{34.6} & \textbf{23.1}\\
    \bottomrule
\end{tabular}\\
        (a) Results on different number of classes in mAcc
        \includegraphics[width=0.47\textwidth]{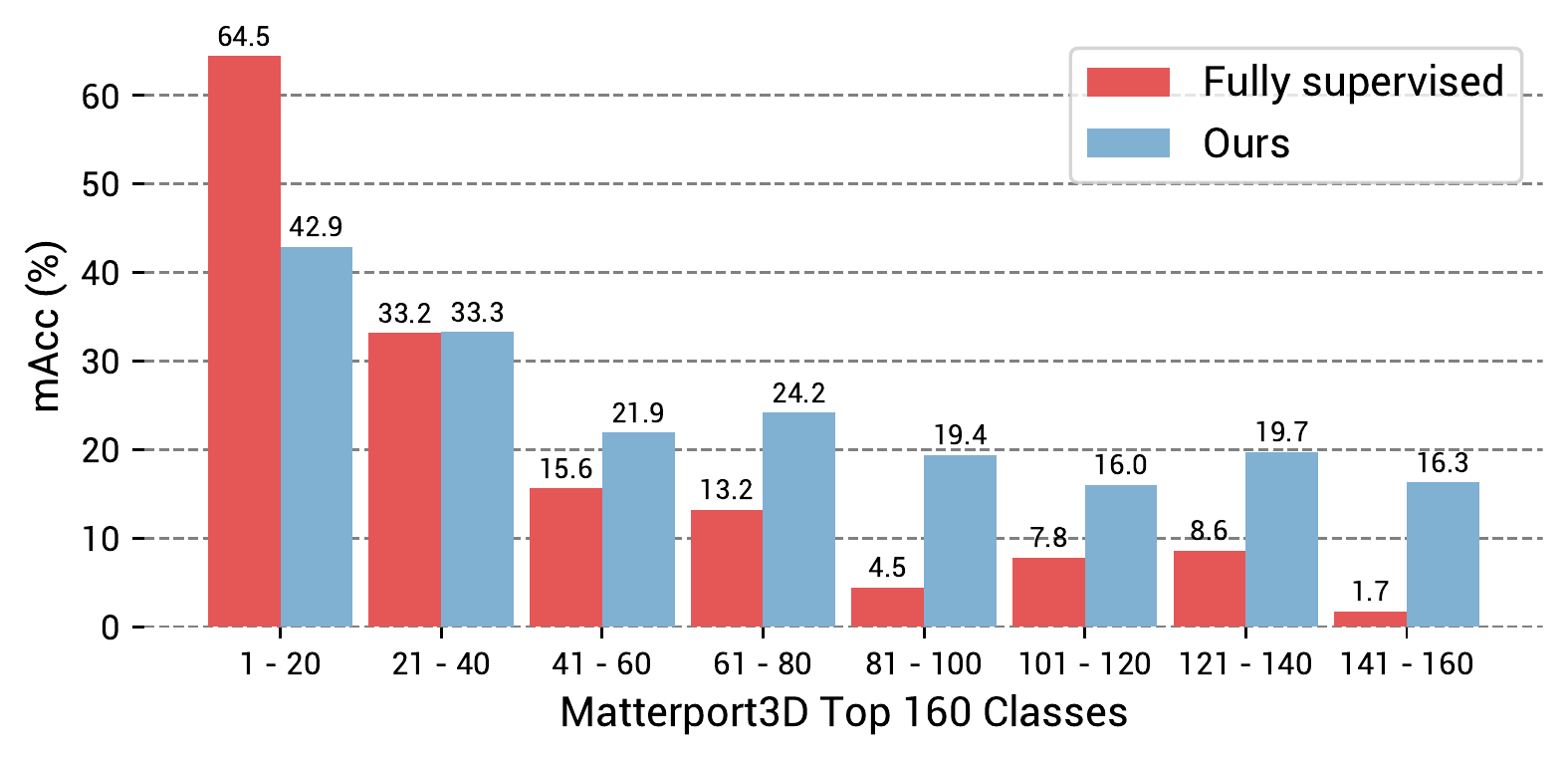}\\
        (b) Window evaluation on groups of 20 classes
    \caption{
        \textbf{Impact of Increasing the Number of Object Classes.}
        Here we show (a) mAcc on Matterport3D~\cite{Chang2017THREEDV} with different numbers of classes $K$, and (b) mAcc within a window of 20 classes ranked by decreasing numbers of examples in training set, \ie the right-most bars represent average of the 20 classes  with fewest examples (e.g., only 5 instances). 
        Even though the fully-supervised approach~\cite{Choy2019CVPR} is trained on each labelset separately, while our model is fixed for all label sets, we can handle the less-common / long-tail classes much better. 
        }
\label{tab:increase_classes}
\end{table}

\boldparagraph{Comparison on 3D semantic segmentation benchmarks}
In~\tabref{tab:benchmark} we compare our approach with both fully-supervised and zero-shot methods on all classes of the nuScenes~\cite{Caesar2020CVPR} validation set, ScanNet~\cite{Dai2017CVPR} validation set, and Matterport3D~\cite{Chang2017THREEDV} test set.
Again, we outperform the zero-shot baseline (MSeg Voting) on both mIoU and mAcc metrics all three datasets.
Although we have noticeable gap to the state-of-the-art fully-supervised approaches, our zero-shot method is surprisingly competitive with fully-supervised approaches from a few years ago \cite{Dai2018CVPR,Huang2019CVPR,Tatarchenko2018CVPR}.
Among all 3 datasets our approach has the smallest gap (only -11.6 mIoU and -8.0 mAcc) to the SOTA fully-supervised approach on Matterport3D.
We conjecture the reason is that Matterport3D is more diverse, which makes the fully-supervised training harder.

Visual comparisons of semantic segmentations are shown in ~\figref{fig:benchmark}.   They show that some of the predictions marked wrong in our results are actually either incorrect or ambiguous ground truth annotations.
For example, in the first row in~\figref{fig:benchmark}, we successfully segment the picture on the wall, while the GT misses it.
And in the nuScenes results, the truck composed of a trailer and the truck head is segmented correctly, but the annotation is not fine-grained enough to separate the parts.

\boldparagraphnoperiod{Impact of increasing the number of object classes.}
Besides the standard benchmarks with only a small set of classes, we also show comparisons as the number of object classes increases.
We evaluate on the most frequent $K$ classes\footnote{$K=21$ was from original Matterport3D benchmark. For $K=40, 80, 160$ we use most frequent $K$ classes of the NYU label set provided with the benchmark.} of Matterport3D, where $K=21, 40, 80, 160$.   For the baseline, we train a separate MinkowskiNet for each $K$.   However, for ours we use the \textit{same} model for all $K$, since it is class agnostic.

As shown in~\tabref{tab:increase_classes}~(a), when trained on only 21 classes, the fully-supervised method performs much better due to the rich 3D labels in the most common classes (wall, floor, chair, etc.).
However, with the increase of the number of classes, our zero-shot approach overtakes the fully-supervised approach, especially when $K$ gets large.
The reason is demonstrated in~\tabref{tab:increase_classes}~(b), where we show the mean accuracy for groups of 20 classes ranked by frequency.
Fully-supervised suffers severely in segmenting tail classes because there are only a few instances available in the training dataset. 
In contrast, we are more robust to such rare objects since we do
not rely upon any 3D labeled data.

\subsection{Ablation Studies \& Analysis}
\label{sec:ablations}
\boldparagraphnoperiod{Does it matter which 2D features are used?}
We tested our method with features extracted from both OpenSeg~\cite{Ghiasi2022ECCV} and LSeg~\cite{Li2022ICLR}.   In most experiments, we found the accuracy and generalizability of OpenSeg features to be better than LSeg (\tabref{tab:zero_shot}, \tabref{tab:benchmark}, and \tabref{tab:ablation_feature}), so we use OpenSeg for all experiments unless explicitly stated otherwise.

\begin{table}[!t]
    \centering
    \footnotesize
    \setlength{\tabcolsep}{0.3cm}
        \resizebox{0.47\textwidth}{!}{
        \begin{tabular}{cl|cc|cc}
    \toprule
    {} & {} & \multicolumn{2}{c|}{ScanNet~\cite{Dai2017CVPR}} & \multicolumn{2}{c}{Matterport3D~\cite{Chang2017THREEDV}}\\
    \cline{3-6}
    {}&{} &mIoU & mAcc & mIoU & mAcc\\\hline
    \multirow{3}{*}{\parbox{0.03\textwidth}{\textbf{Ours}\\LSeg}}   & 2D Fusion  & 50.0 & 62.7 & 32.3 & 40.0\\
    {} & 3D Distill & 52.9 & 63.2 & 41.9 & 51.2\\
    {}   & 2D-3D Ens. & \textbf{54.2} & 66.6 & \textbf{43.4} & 53.5\\\hline
    \multirow{3}{*}{\parbox{0.03\textwidth}{\textbf{Ours}\\OpenSeg}}   & 2D Fusion  & 41.4 & 63.6 & 32.4 & 45.0\\
    {} & 3D Distill & 46.0 & 66.3 & 41.3 & 55.1\\
    {}   & 2D-3D Ens. & 47.5 & \textbf{70.7} & 42.6 & \textbf{59.2}\\
    \bottomrule
\end{tabular}}
    \caption{
        \textbf{Ablation Study.}   Comparison of semantic segmentation performance of different 3D features computed by our method.
        }
\label{tab:ablation_feature}
\end{table}

\boldparagraphnoperiod{Is our 2D-3D ensemble method effective?}
In~\tabref{tab:ablation_feature}, we ablate the performance for predicting features on 3D points including only image feature fusion (\secref{sec:fusion}), only running the distilled MinkowskiNet (\secref{sec:distill}), and our full 2D-3D ensemble model (\secref{sec:ensemble}).
As can be seen, on all scenarios (different datasets, metrics, and 2D features), our proposed 2D-3D ensemble model performs the best.
This suggests that leveraging patterns in both 2D and 3D domains makes the ensemble features more robust and descriptive.

\tableensembleratio
\boldparagraphnoperiod{What features does our 2D-3D ensemble method use most?}
Here we study how our ensemble model selects among the 2D and 3D features, and investigate how it changes
with increasing numbers of classes in the label set.
As shown in~\tabref{tab:ensemble_ratio}, we find that the majority of predictions ($\sim70\%$) select the 3D features, 
corroborating the value of our 3D distillation model.
However, the percentage of predictions coming from 2D features increases with the number of classes, suggesting that
the 2D features are more important for long-tailed classes, which tend to be smaller in both size and 
number of training examples.

\section{Applications}
This section investigates new 3D scene understanding applications enabled by our approach.   Since the feature vectors estimated for every 3D point are co-embedded with text and images, it becomes possible to extract information about a scene using arbitrary text and image queries.  The following are just a few example applications.\footnote{Please note that all of these applications are zero-shot -- i.e., none of them leverage any labeled data from any 3D scene understanding dataset.}

\boldparagraph{Open-vocabulary 3D object search}
We first investigate whether it is possible to query a 3D scene database to find examples based on their names -- e.g., ``find a teddy bear in the Matterport3D test set.''
To do so, we ask a user to enter an arbitrary text string as a query, encode the CLIP embedding vector for the query, and then compute the cosine-similarity of that query vector with the features of every 3D point in the Matteport3D test set (containing 18 buildings with 406 indoor and outdoor regions) to discover the best matches.   In our implementation, we return at most one match per region (i.e., room, as defined in the dataset) to ensure diversity of the retrieval results.

\figretrieval

\begin{table}[!t]
    \centering
    \footnotesize
    \setlength{\tabcolsep}{0.15cm}
    \resizebox{0.42\textwidth}{!}{
        \begin{tabular}{l|c|c|c|c|c}
    \toprule
    Class             & \#    & \#    & \#    & \#      & \#  \\
    Name              & All   & Test  & Found & Missed  & New \\
    \midrule
    Fire Extinguisher & 25    & 3     & 3     & 0      & 0 \\
    Telephone         & 21    & 4     & 15    & 2      & 13 \\
    Exit Sign         & 15    & 5*    & 8     & 0      & 3 \\
    Piano             & 15    & 1     & 2     & 0      & 1 \\
    Ball              & 15    & 1*    & 4     & 0      & 3 \\
    Hat               & 15    & 1*    & 1     & 0      & 0 \\
    Bulletin Board    & 6     & 0     & 1     & 0      & 1 \\
    Globe             & 5     & 2     & 5     & 0      & 3 \\
    Teddy Bear        & 2     & 0     & 1     & 0      & 1 \\
    Toy Giraffe       & 1     & 1     & 1     & 0      & 0 \\
    Yellow Egg-Shaped Vase   & 1 & 0     & 1     & 0      & 1 \\
    \bottomrule
\end{tabular}

}
    \caption{
        \textbf{Open-vocabulary 3D Search Results.}
        Each row depicts a search of the Matterport3D test set for a class given as a text query.   The columns list the \# of instances in the ground truth for the whole dataset (\# All), the \# in the test set (\# Test, counting clusters of nearby objects as one when marked with a '*'), the \# of top matches found with 100\% precision (\# Found), the \# of GT instances missed amongst those top matches (\# Missed), and the \# newly discovered that were not in the GT (\# New).
        }
\label{tab:retrieval}
\end{table}

\figproperties
\figimageretrieval

\figref{fig:retrieval} shows a few example top-1 results. 
Most other specific text queries yield nearly perfect results.  To evaluate that observation quantitatively, we chose a sampling of 10 raw categories from the ground truth set of Matterport3D, retrieved the best matching 3D points from the test set, and then visually verified the correctness of the top matches.  For each query, \tabref{tab:retrieval} reports the numbers of instances in  the test set (\# Test) along with the number of instances found with 100\% precision before the first mistake in the ranked list.   The results are very encouraging.  In all of these queries, only two ground truth instances were missed (two telephones).   On the other hand, 26 instances were found among these top matches that were not correctly labeled in the ground truth, including 13 telephones.   Overall, these results suggest that our open-vocabulary retrieval application identifies these relatively rare classes at least as well as the manually labeling process did.   See the supplemental material for the full set of results.

\boldparagraph{Image-based 3D object detection}
We next investigate whether it is possible to query a 3D scene database to retrieve examples based on similarities to a given input image~-- e.g., ``find points in a Matterport3D building that match this image.''  Given a set of query images, we encode them with CLIP image encoder, compute cosine-similarities to 2D-3D ensemble features for 3D points, and then threshold to produce a 3D object detection and mask, see~\figref{fig:img_retrieval}.  Note that the pool table and dining table are identified correctly, even though both are types of ``tables.''

\boldparagraph{Open-vocabulary 3D scene understanding and exploration}
Finally, we investigate whether it is possible to query a 3D scene to understand properties that extend beyond category labels.   Since the CLIP embedding space is trained with a massive corpus of text, it can represent far more than category labels -- it can encode physical properties, surface materials, human affordances, potential functions, room types, and so on.   We hypothesize that we can use the co-embedding our 3D points with the CLIP features to \textit{discover} these types of information about a scene.

\figref{fig:query_features} shows some example results for querying about physical properties, surface materials, and potential sites of activities.   
From these examples, we find that the OpenScene is indeed able to relate words  to relevant areas of the scene -- e.g., the beds, couches, and stuffed chairs match ``Comfy,'' the oven and fireplace match ``Hot,'' and the piano keyboard matches ``Play.''  This diversity of 3D scene understanding would be difficult to achieve with fully supervised methods without massive 3D labeling efforts.   In the authors' opinion, this is the most interesting result of the paper.

\section{Limitations and Future Work}

This paper introduces a task-agnostic method to co-embed 3D points in a feature space with text and image pixels and demonstrates its utility for zero-shot, open-vocabulary 3D scene understanding.  It achieves state-of-the-art for zero-shot 3D semantic segmentation on standard benchmarks, outperforms supervised approaches in 3D semantic segmentation with many class labels, and enables new open-vocabulary applications where arbitrary text and image queries can be used to query 3D scenes, all without using any labeled 3D data. These results suggest a new direction for 3D scene understanding, where foundation models trained from massive multi-modal datasets guide 3D scene understanding systems rather than training them only with small labeled 3D datasets.  

There are several limitations of our work and still much to do to realize the full potential of the proposed approach.   First, the inference algorithm could probably take better advantage of pixel features when images are present at test time using earlier fusion (we tried this with limited success).   Second, the experiments could be expanded to investigate the limits of open-vocabulary 3D scene understanding on a wider variety of tasks.   We evaluated extensively on closed-set 3D semantic segmentation, but provide only qualitative results for other tasks since 3D benchmarks with ground truth are scarce.  In future work, it will be interesting to design experiments to quantify the success of open vocabulary queries for tasks where ground truth is not available.

\boldparagraph{Acknowledgement}
We sincerely thank Golnaz Ghiasi for providing guidance on using OpenSeg model. Our appreciation extends to Huizhong Chen, Yin Cui, Tom Duerig, Dan Gnanapragasam, Xiuye Gu, Leonidas Guibas,
Nilesh Kulkarni, Abhijit Kundu, Hao-Ning Wu, Louis Yang, Guandao Yang, Xiaoshuai Zhang, Howard Zhou, and Zihan Zhu for helpful discussion. We are also grateful to Charles R. Qi and Paul-Edouard Sarlin for their proofreading.

{\small
\bibliographystyle{ieee_fullname}
\bibliography{egbib}

\begin{thebibliography}{10}\itemsep=-1pt

\bibitem{Alayrac2022ARXIV}
Jean-Baptiste Alayrac, Jeff Donahue, Pauline Luc, Antoine Miech, Iain Barr,
  Yana Hasson, Karel Lenc, Arthur Mensch, Katie Millican, Malcolm Reynolds,
  et~al.
\newblock Flamingo: a visual language model for few-shot learning.
\newblock In {\em NeurIPS}, 2022.

\bibitem{Behley2019ICCV}
Jens Behley, Martin Garbade, Andres Milioto, Jan Quenzel, Sven Behnke, Cyrill
  Stachniss, and Jurgen Gall.
\newblock Semantickitti: A dataset for semantic scene understanding of lidar
  sequences.
\newblock In {\em ICCV}, 2019.

\bibitem{Caesar2020CVPR}
Holger Caesar, Varun Bankiti, Alex~H Lang, Sourabh Vora, Venice~Erin Liong,
  Qiang Xu, Anush Krishnan, Yu Pan, Giancarlo Baldan, and Oscar Beijbom.
\newblock nuscenes: A multimodal dataset for autonomous driving.
\newblock In {\em CVPR}, 2020.

\bibitem{Chang2017THREEDV}
Angel Chang, Angela Dai, Thomas Funkhouser, Maciej Halber, Matthias Niessner,
  Manolis Savva, Shuran Song, Andy Zeng, and Yinda Zhang.
\newblock Matterport3d: Learning from rgb-d data in indoor environments.
\newblock In {\em 3DV}, 2017.

\bibitem{Chen2020ECCV}
Dave~Zhenyu Chen, Angel~X Chang, and Matthias Nie{\ss}ner.
\newblock Scanrefer: 3d object localization in rgb-d scans using natural
  language.
\newblock In {\em ECCV}, 2020.

\bibitem{Cheraghian2019BMVC}
Ali Cheraghian, Shafin Rahman, Dylan Campbell, and Lars Petersson.
\newblock Mitigating the hubness problem for zero-shot learning of 3d objects.
\newblock In {\em BMVC}, 2019.

\bibitem{Cheraghian2020WACV}
Ali Cheraghian, Shafin Rahman, Dylan Campbell, and Lars Petersson.
\newblock Transductive zero-shot learning for 3d point cloud classification.
\newblock In {\em WACV}, 2020.

\bibitem{Cheraghian2022IJCV}
Ali Cheraghian, Shafin Rahman, Townim~F Chowdhury, Dylan Campbell, and Lars
  Petersson.
\newblock Zero-shot learning on 3d point cloud objects and beyond.
\newblock {\em IJCV}, 2022.

\bibitem{Cheraghian2019MVA}
Ali Cheraghian, Shafin Rahman, and Lars Petersson.
\newblock Zero-shot learning of 3d point cloud objects.
\newblock In {\em MVA}, 2019.

\bibitem{Choy2019CVPR}
Christopher Choy, JunYoung Gwak, and Silvio Savarese.
\newblock 4d spatio-temporal convnets: Minkowski convolutional neural networks.
\newblock In {\em CVPR}, 2019.

\bibitem{Dai2017CVPR}
Angela Dai, Angel~X Chang, Manolis Savva, Maciej Halber, Thomas Funkhouser, and
  Matthias Nie{\ss}ner.
\newblock Scannet: Richly-annotated 3d reconstructions of indoor scenes.
\newblock In {\em CVPR}, 2017.

\bibitem{Dai2018CVPR}
Angela Dai, Daniel Ritchie, Martin Bokeloh, Scott Reed, J{\"u}rgen Sturm, and
  Matthias Nie{\ss}ner.
\newblock Scancomplete: Large-scale scene completion and semantic segmentation
  for 3d scans.
\newblock In {\em CVPR}, 2018.

\bibitem{Deng2021CVPR}
Shengheng Deng, Xun Xu, Chaozheng Wu, Ke Chen, and Kui Jia.
\newblock 3d affordancenet: A benchmark for visual object affordance
  understanding.
\newblock In {\em CVPR}, 2021.

\bibitem{Fan2022CVPR}
Hehe Fan, Xiaojun Chang, Wanyue Zhang, Yi Cheng, Ying Sun, and Mohan
  Kankanhalli.
\newblock Self-supervised global-local structure modeling for point cloud
  domain adaptation with reliable voted pseudo labels.
\newblock In {\em CVPR}, 2022.

\bibitem{Geiger2012CVPR}
Andreas Geiger, Philip Lenz, and Raquel Urtasun.
\newblock Are we ready for autonomous driving? the kitti vision benchmark
  suite.
\newblock In {\em CVPR}, 2012.

\bibitem{Genova2021THREEDV}
Kyle Genova, Xiaoqi Yin, Abhijit Kundu, Caroline Pantofaru, Forrester Cole,
  Avneesh Sud, Brian Brewington, Brian Shucker, and Thomas Funkhouser.
\newblock Learning 3d semantic segmentation with only 2d image supervision.
\newblock In {\em 3DV}, 2021.

\bibitem{Ghiasi2022ECCV}
Golnaz Ghiasi, Xiuye Gu, Yin Cui, and Tsung-Yi Lin.
\newblock Open-vocabulary image segmentation.
\newblock In {\em ECCV}, 2022.

\bibitem{Gu2022ICLR}
Xiuye Gu, Tsung-Yi Lin, Weicheng Kuo, and Yin Cui.
\newblock Open-vocabulary object detection via vision and language knowledge
  distillation.
\newblock In {\em ICLR}, 2022.

\bibitem{Ha2022CORL}
Huy Ha and Shuran Song.
\newblock Semantic abstraction: Open-world 3d scene understanding from 2d
  vision-language models.
\newblock In {\em CoRL}, 2022.

\bibitem{Han2020CVPR}
Lei Han, Tian Zheng, Lan Xu, and Lu Fang.
\newblock Occuseg: Occupancy-aware 3d instance segmentation.
\newblock In {\em CVPR}, 2020.

\bibitem{Hu2021CVPR}
Wenbo Hu, Hengshuang Zhao, Li Jiang, Jiaya Jia, and Tien-Tsin Wong.
\newblock Bidirectional projection network for cross dimension scene
  understanding.
\newblock In {\em CVPR}, 2021.

\bibitem{Hu2021ICCV}
Zeyu Hu, Xuyang Bai, Jiaxiang Shang, Runze Zhang, Jiayu Dong, Xin Wang,
  Guangyuan Sun, Hongbo Fu, and Chiew-Lan Tai.
\newblock Vmnet: Voxel-mesh network for geodesic-aware 3d semantic
  segmentation.
\newblock In {\em ICCV}, 2021.

\bibitem{Hua2016THREEDV}
Binh-Son Hua, Quang-Hieu Pham, Duc~Thanh Nguyen, Minh-Khoi Tran, Lap-Fai Yu,
  and Sai-Kit Yeung.
\newblock Scenenn: A scene meshes dataset with annotations.
\newblock In {\em 3DV}, 2016.

\bibitem{Huang2019CVPR}
Jingwei Huang, Haotian Zhang, Li Yi, Thomas Funkhouser, Matthias Nie{\ss}ner,
  and Leonidas~J Guibas.
\newblock Texturenet: Consistent local parametrizations for learning from
  high-resolution signals on meshes.
\newblock In {\em CVPR}, 2019.

\bibitem{Jia2021ICML}
Chao Jia, Yinfei Yang, Ye Xia, Yi-Ting Chen, Zarana Parekh, Hieu Pham, Quoc Le,
  Yun-Hsuan Sung, Zhen Li, and Tom Duerig.
\newblock Scaling up visual and vision-language representation learning with
  noisy text supervision.
\newblock In {\em ICML}, 2021.

\bibitem{Kingma2015ICLR}
Diederik~P Kingma and Jimmy Ba.
\newblock Adam: A method for stochastic optimization.
\newblock In {\em ICLR}, 2015.

\bibitem{Kundu2020ECCV}
Abhijit Kundu, Xiaoqi Yin, Alireza Fathi, David Ross, Brian Brewington, Thomas
  Funkhouser, and Caroline Pantofaru.
\newblock Virtual multi-view fusion for 3d semantic segmentation.
\newblock In {\em ECCV}, 2020.

\bibitem{Kuo2022ARXIV}
Weicheng Kuo, Yin Cui, Xiuye Gu, AJ Piergiovanni, and Anelia Angelova.
\newblock F-vlm: Open-vocabulary object detection upon frozen vision and
  language models.
\newblock In {\em ICLR}, 2023.

\bibitem{Lambert2020CVPR}
John Lambert, Zhuang Liu, Ozan Sener, James Hays, and Vladlen Koltun.
\newblock Mseg: A composite dataset for multi-domain semantic segmentation.
\newblock In {\em CVPR}, 2020.

\bibitem{Li2022ICLR}
Boyi Li, Kilian~Q Weinberger, Serge Belongie, Vladlen Koltun, and Ren{\'e}
  Ranftl.
\newblock Language-driven semantic segmentation.
\newblock In {\em ICLR}, 2022.

\bibitem{Li2022CVPR}
Jinke Li, Xiao He, Yang Wen, Yuan Gao, Xiaoqiang Cheng, and Dan Zhang.
\newblock Panoptic-phnet: Towards real-time and high-precision lidar panoptic
  segmentation via clustering pseudo heatmap.
\newblock In {\em CVPR}, 2022.

\bibitem{Li2019CVPR}
Xueting Li, Sifei Liu, Kihwan Kim, Xiaolong Wang, Ming-Hsuan Yang, and Jan
  Kautz.
\newblock Putting humans in a scene: Learning affordance in 3d indoor
  environments.
\newblock In {\em CVPR}, 2019.

\bibitem{Liang2022ARXIV}
Feng Liang, Bichen Wu, Xiaoliang Dai, Kunpeng Li, Yinan Zhao, Hang Zhang,
  Peizhao Zhang, Peter Vajda, and Diana Marculescu.
\newblock Open-vocabulary semantic segmentation with mask-adapted clip.
\newblock In {\em CVPR}, 2023.

\bibitem{Liao2022PAMI}
Yiyi Liao, Jun Xie, and Andreas Geiger.
\newblock Kitti-360: A novel dataset and benchmarks for urban scene
  understanding in 2d and 3d.
\newblock {\em IEEE TPAMI}, 2022.

\bibitem{Liu2021ARXIV}
Bo Liu, Shuang Deng, Qiulei Dong, and Zhanyi Hu.
\newblock Language-level semantics conditioned 3d point cloud segmentation.
\newblock {\em arXiv preprint arXiv:2107.00430}, 2021.

\bibitem{Liu2021ICCV}
Yunze Liu, Qingnan Fan, Shanghang Zhang, Hao Dong, Thomas Funkhouser, and Li
  Yi.
\newblock Contrastive multimodal fusion with tupleinfonce.
\newblock In {\em ICCV}, 2021.

\bibitem{Ma2022BMVC}
Chaofan Ma, Yuhuan Yang, Yanfeng Wang, Ya Zhang, and Weidi Xie.
\newblock Open-vocabulary semantic segmentation with frozen vision-language
  models.
\newblock In {\em BMVC}, 2022.

\bibitem{Mccormac2017ICRA}
John McCormac, Ankur Handa, Andrew Davison, and Stefan Leutenegger.
\newblock Semanticfusion: Dense 3d semantic mapping with convolutional neural
  networks.
\newblock In {\em ICRA}, 2017.

\bibitem{Michele2021THREEDV}
Bj{\"o}rn Michele, Alexandre Boulch, Gilles Puy, Maxime Bucher, and Renaud
  Marlet.
\newblock Generative zero-shot learning for semantic segmentation of 3d point
  clouds.
\newblock In {\em 3DV}, 2021.

\bibitem{Nekrasov2021THREEDV}
Alexey Nekrasov, Jonas Schult, Or Litany, Bastian Leibe, and Francis Engelmann.
\newblock Mix3d: Out-of-context data augmentation for 3d scenes.
\newblock In {\em 3DV}, 2021.

\bibitem{Paszke2019NEURIPS}
Adam Paszke et~al.
\newblock Pytorch: An imperative style, high-performance deep learning library.
\newblock In {\em NeurIPS}, 2019.

\bibitem{Qi2017NIPS}
Charles~Ruizhongtai Qi, Li Yi, Hao Su, and Leonidas~J Guibas.
\newblock Pointnet++: Deep hierarchical feature learning on point sets in a
  metric space.
\newblock In {\em NIPS}, 2017.

\bibitem{Radford2021ICML}
Alec Radford, Jong~Wook Kim, Chris Hallacy, Aditya Ramesh, Gabriel Goh,
  Sandhini Agarwal, Girish Sastry, Amanda Askell, Pamela Mishkin, Jack Clark,
  et~al.
\newblock Learning transferable visual models from natural language
  supervision.
\newblock In {\em ICML}, 2021.

\bibitem{Rao2022CVPR}
Yongming Rao, Wenliang Zhao, Guangyi Chen, Yansong Tang, Zheng Zhu, Guan Huang,
  Jie Zhou, and Jiwen Lu.
\newblock Denseclip: Language-guided dense prediction with context-aware
  prompting.
\newblock In {\em CVPR}, 2022.

\bibitem{Robert2022CVPR}
Damien Robert, Bruno Vallet, and Loic Landrieu.
\newblock Learning multi-view aggregation in the wild for large-scale 3d
  semantic segmentation.
\newblock In {\em CVPR}, 2022.

\bibitem{Rozenberszki2022ARXIV}
David Rozenberszki, Or Litany, and Angela Dai.
\newblock Language-grounded indoor 3d semantic segmentation in the wild.
\newblock In {\em ECCV}, 2022.

\bibitem{Sautier2022CVPR}
Corentin Sautier, Gilles Puy, Spyros Gidaris, Alexandre Boulch, Andrei Bursuc,
  and Renaud Marlet.
\newblock Image-to-lidar self-supervised distillation for autonomous driving
  data.
\newblock In {\em CVPR}, 2022.

\bibitem{Schult2020CVPR}
Jonas Schult, Francis Engelmann, Theodora Kontogianni, and Bastian Leibe.
\newblock Dualconvmesh-net: Joint geodesic and euclidean convolutions on 3d
  meshes.
\newblock In {\em CVPR}, 2020.

\bibitem{Shafiullah2022ARXIV}
Nur Muhammad~Mahi Shafiullah, Chris Paxton, Lerrel Pinto, Soumith Chintala, and
  Arthur Szlam.
\newblock Clip-fields: Weakly supervised semantic fields for robotic memory.
\newblock {\em arXiv preprint arXiv:2210.05663}, 2022.

\bibitem{Sun2020CVPR}
Pei Sun, Henrik Kretzschmar, Xerxes Dotiwalla, Aurelien Chouard, Vijaysai
  Patnaik, Paul Tsui, James Guo, Yin Zhou, Yuning Chai, Benjamin Caine, et~al.
\newblock Scalability in perception for autonomous driving: Waymo open dataset.
\newblock In {\em CVPR}, 2020.

\bibitem{Tatarchenko2018CVPR}
Maxim Tatarchenko, Jaesik Park, Vladlen Koltun, and Qian-Yi Zhou.
\newblock Tangent convolutions for dense prediction in 3d.
\newblock In {\em CVPR}, 2018.

\bibitem{Van2008JMLR}
Laurens Van~der Maaten and Geoffrey Hinton.
\newblock Visualizing data using t-sne.
\newblock {\em JMLR}, 2008.

\bibitem{Vineet2015ICRA}
Vibhav Vineet, Ondrej Miksik, Morten Lidegaard, Matthias Nie{\ss}ner, Stuart
  Golodetz, Victor~A Prisacariu, Olaf K{\"a}hler, David~W Murray, Shahram
  Izadi, Patrick P{\'e}rez, et~al.
\newblock Incremental dense semantic stereo fusion for large-scale semantic
  scene reconstruction.
\newblock In {\em ICRA}, 2015.

\bibitem{Wald2020CVPR}
Johanna Wald, Helisa Dhamo, Nassir Navab, and Federico Tombari.
\newblock Learning 3d semantic scene graphs from 3d indoor reconstructions.
\newblock In {\em CVPR}, 2020.

\bibitem{Wang2021CVPR}
Jiashun Wang, Huazhe Xu, Jingwei Xu, Sifei Liu, and Xiaolong Wang.
\newblock Synthesizing long-term 3d human motion and interaction in 3d scenes.
\newblock In {\em CVPR}, 2021.

\bibitem{Wang2017SIGGRAPH}
Peng-Shuai Wang, Yang Liu, Yu-Xiao Guo, Chun-Yu Sun, and Xin Tong.
\newblock O-cnn: Octree-based convolutional neural networks for 3d shape
  analysis.
\newblock {\em ACM TOG}, 2017.

\bibitem{Wang2022CORL}
Yue Wang, Vitor~Campagnolo Guizilini, Tianyuan Zhang, Yilun Wang, Hang Zhao,
  and Justin Solomon.
\newblock Detr3d: 3d object detection from multi-view images via 3d-to-2d
  queries.
\newblock In {\em CoRL}, 2022.

\bibitem{Wu2015CVPR}
Zhirong Wu, Shuran Song, Aditya Khosla, Fisher Yu, Linguang Zhang, Xiaoou Tang,
  and Jianxiong Xiao.
\newblock 3d shapenets: A deep representation for volumetric shapes.
\newblock In {\em CVPR}, 2015.

\bibitem{Xu2022CVPR}
Jiarui Xu, Shalini De~Mello, Sifei Liu, Wonmin Byeon, Thomas Breuel, Jan Kautz,
  and Xiaolong Wang.
\newblock Groupvit: Semantic segmentation emerges from text supervision.
\newblock In {\em CVPR}, 2022.

\bibitem{Ye2022ARXIV}
Dongqiangzi Ye, Zixiang Zhou, Weijia Chen, Yufei Xie, Yu Wang, Panqu Wang, and
  Hassan Foroosh.
\newblock Lidarmultinet: Towards a unified multi-task network for lidar
  perception.
\newblock {\em arXiv preprint arXiv:2209.09385}, 2022.

\bibitem{Zabari2021ARXIV}
Nir Zabari and Yedid Hoshen.
\newblock Open-vocabulary semantic segmentation using test-time distillation.
\newblock In {\em ECCVW}, 2022.

\bibitem{Zhang2022CVPRb}
Renrui Zhang, Ziyu Guo, Wei Zhang, Kunchang Li, Xupeng Miao, Bin Cui, Yu Qiao,
  Peng Gao, and Hongsheng Li.
\newblock Pointclip: Point cloud understanding by clip.
\newblock In {\em CVPR}, 2022.

\bibitem{Zhou2022ECCV}
Chong Zhou, Chen~Change Loy, and Bo Dai.
\newblock Extract free dense labels from clip.
\newblock In {\em ECCV}, 2022.

\end{thebibliography}
}

\clearpage
\setcounter{section}{0}
\setcounter{figure}{0}
\setcounter{table}{0}
\renewcommand\thesection{\Alph{section}}
\renewcommand\thetable{\Alph{table}}
\renewcommand\thefigure{\Alph{figure}}

\title{OpenScene: 3D Scene Understanding with Open Vocabularies\\---Supplementary Material---}
\author{
Songyou Peng$^{1,2,3}$ \qquad Kyle Genova$^{1}$ \qquad Chiyu ``Max'' Jiang$^{4}$\qquad Andrea Tagliasacchi$^{1,5}$\\
Marc Pollefeys$^{2}$ \qquad Thomas Funkhouser$^{1}$ \vspace{0.05cm}\\
{\small $^{1}$ Google Research \quad $^{2}$ ETH Zurich \quad $^{3}$ MPI for Intelligent Systems, Tübingen \quad $^{4}$ Waymo LLC \quad $^{5}$ Simon Fraser University}
\\{\small\href{https://pengsongyou.github.io/openscene}{pengsongyou.github.io/openscene}}
}

\maketitle

In this \textbf{supplementary document}, we first provide implementation details in~\secref{sec:implement}. 
Next, we supply additional investigation of our methods in~\secref{sec:investigation}. 
Full results of open-vocabulary object retrieval experiments are shown in~\secref{sec:retrieve}.
More results of open-vocabulary scene exploration results can be found in~\secref{sec:explore}.

\section{Implementation Details}\label{sec:implement}

\boldparagraph{3D Distillation}
We implement our pipeline in PyTorch~\cite{Paszke2019NEURIPS}.
To distill $\cE^\text{3D}$, we use Adam~\cite{Kingma2015ICLR} as the optimizer with an initial learning rate of $1e{-}4$ and train for $100$ epochs.
For MinkowskiNet we use a voxel size of 2cm for ScanNet and Matterport3D experiments, and 5cm for nuScenes.
For indoor datasets, we input all points of a scene to the 3D backbone to have the full contexts, but for the distillation loss (Eq. 2) in the paper we only supervise with 20K uniformly sampled point features at every iteration due to the memory constraints.
For nuScenes, we input all Lidar points within the half-second segments, and only train with point features at the last time stamp.
We use a batch size of 8 for ScanNet and Matterport3D with a single NVIDIA A100 (40G). For nuScenes, we use a batch size of 16 with 4 A100 GPUs.
It takes around 24 hours to train, and 0.1 seconds for inference.
Moreover, for all dataset we only take in the 3D point position as input to the MinkowskiNet during distillation. 

\boldparagraph{More Details of Feature Fusion}
For Matterport3D and nuScenes, we use all images of each scene for fusion, while for ScanNet, we sample 1 out of every 20 video frames.

As for the occlusion test, for dataset like ScanNet and Matterport3D where the depth map is provided for each RGB image, we do occlusion test to guarantee that a pixel is only paired with a \emph{visible} surface point.
For every surface point, we first find its corresponding pixel in an image, and we can obtain the distance between that pixel and 3D point. The 3D points and pixel are only paired when the difference between the distance and the depth value of that pixel is smaller than a threshold $\sigma$.
The threshold $\sigma$ is proportional to the depth value $D$. We use $\sigma = 0.2D$ for ScanNet due to the highly noisy depths and $\sigma=0.02D$ for Matterport.
For pixels with ``invalid'' regions of the depth map, we do not project their features to 3D points.

For nuScenes Lidar points, since no depth images are provided, no occlusion test is conducted, and we only use the synchronized images and the corresponding Lidar points on the last timestamp of a 0.5 second segment.

\boldparagraph{Our nuScenes Evaluation}
Unlike ScanNet and Matterport where each 3D surface point usually having multiple corresponding images, in nuScenes most Lidar points only have one corresponding view, maximum two views at the same time stamp.
Therefore, we directly project the single pixel feature to most Lidar points, and use average pooling when there are 2 views.
There are 16 classes in the nuScenes lidarseg benchmark, and some of these class names are ambiguous. 
Since our method can take in arbitrary text prompts, we can pre-define some non-ambiguous classes names for each class, and then map the predictions from these non-ambiguous classes back to the 16 classes. The pre-defined classes names are listed in~\tabref{tab:nuscenes_mapping}.

\tablenuscenesmapping

\boldparagraph{MSeg~\cite{Lambert2020CVPR} Voting}
MSeg supports a unified taxonomy of 194 classes. We use their official image semantic segmentation code\footnote{\url{https://github.com/mseg-dataset/mseg-semantic}} and their pretrained MSeg-3m-1080p model.
MSeg already provided the mapping from some of 194 classes to 20 ScanNet classes, so we directly use the mapping. For Matterport3D, we simply add the mapping from ``ceiling'' in the MSeg labelset. As for nuScenes, we manually define the mapping from MSeg to nuScenes 16 labelsets. However, for the ``construction vehicles'', ``traffic cone'', and 'other flat', there is no mapping at all, so we set them to unknown. 

As for the majority voting for MSeg multi-view predictions, what we do is the following. Given a surface point and its corresponding multi-view MSeg semantic segmentation, we take the class in the majority of views as the voting results for this point. If there are two classes and only two views, we directly flip the coin to decide point labels.

\boldparagraph{Simple Prompt Engineering}
Given a set of text prompts, we use a simple prompt engineering trick before extract CLIP text features. For each object class ``XX'' (except for ``other'') we modify the text prompts to ``a XX in a scene'', for instance  ``a chair in a scene''. With such a simple modification, we observe +2.3 mIoU performance boost with our LSeg ensemble model for ScanNet evaluation.
We apply the trick for all our benchmark comparison experiments.

\section{Additional Analysis}\label{sec:investigation}

\boldparagraphnoperiod{Can we transfer to another dataset with different labelsets?} 

\begin{table}[!t]
    \centering
    \small
    \setlength{\tabcolsep}{0.05cm}
    \resizebox{0.47\textwidth}{!}{
        \begin{tabular}{l|c|c||c|c}
    \toprule
    {} & \multicolumn{2}{c||}{\textbf{3D-only model}} & \multicolumn{2}{c}{\textbf{2D-3D ensemble model}}\\\midrule
    {Evaluation Datasets} & ScanNet20 & Matterport40 & ScanNet20 & Matterport40 \\\midrule
    {} & mIoU (\%) & mAcc (\%) & mIoU (\%) & mAcc (\%)\\\hline
    Distill w/ ScanNet images &46.1 & 37.6 &47.7 & 46.4\\
    Distill w/ Matterport images & 38.0 & 47.1 & 47.1 & 50.9\\
    \bottomrule
\end{tabular}
}
    \caption{
        \textbf{Domain Transfer with Open Vocabularies.}. These results show that it is possible to apply our models trained on ScanNet~\cite{Dai2017CVPR} to a novel 3D semantic segmentation task with a different labelset in Matterport3D~\cite{Chang2017THREEDV}, and vice versa.   Since our trained models are task-agnostic (they predict only CLIP features), they can be applied to arbitrary label sets without retraining.
       }
\label{tab:transfer}
\end{table}

Here we investigate the ability of our trained models to handle domain transfer between 3D segmentation benchmarks with different labelsets.  We train on one dataset (e.g., ScanNet20) and then test on another (e.g., Matterport40) without any retraining (\tabref{tab:transfer}).   Since our trained model is task agnostic (it predicts only CLIP features), it does not over-fit to the classes of the training set, and thus can transfer to other datasets with different classes directly.   Doing the same using a fully-supervised approach would require a sophisticated domain-transfer algorithm (e.g., \cite{Fan2022CVPR}).

\boldparagraph{Ablation on multi-view fusion strategy}
\tableablationfusion
We ablate different multi-view feature fusion strategies in~\tabref{tab:ablation_fusion}. Random means that having multiple features corresponding to one surface point, we randomly assign one feature to the 3D point. For Median, we take the feature that has the smallest Euclidean distance in the feature space to all other features.
As can be seen, the simple average pooling yields the best results, and we use it for all our experiments. 

\boldparagraph{Visualization of our 2D-3D ensemble model}
 In~\figref{fig:ensemble_ratio} we study our ensemble model on how to select 2D and 3D features for prediction on a Matterport3D house based on different labelsets. 
 First, we can notice that our ensemble model uses 3D features for those large areas like floors and walls, while 2D features are preferrable for smaller objects.
 Second, when comparing the feature selections using 21 and 160 classes, we can see that when the number of classes increases, our ensemble model selects more 2D features for the segmentation. The possible reason is that 2D image features can better understand those fine-grained concept than purely from 3D point clouds.
 For example, on the bottom-right there is a pool table there. When using 21 class labels, it is segmented as a table, so 3D features are preferrable. When using 160 class labels for 2D-3D ensemble, it is much easier to understand the concept of ``pool table'' using 2D images than 3D point clouds.

\boldparagraph{Definition of Zero-Shot Learning (ZSL)}
The terminology for ZSL is ambiguous in the literature. In a theoretical ZSL system, there should be no training data of any kind from seen classes.
However, almost all real-world ZSL systems utilize general-purpose feature extractors pretrained for proxy tasks on large datasets.
For example, 3DGenZ~\cite{Michele2021THREEDV} proposes a ZSL variant utilizing image features pretrained on ImageNet (see section 4.5 in their paper), while OpenSeg~\cite{Ghiasi2022ECCV}, LSeg~\cite{Li2022ICLR}, CLIP~\cite{Radford2021ICML}, and ALIGN~\cite{Jia2021ICML} propose ZSL methods trained on alt-text, as we also do. 
The authors of those papers all describe their methods as ZSL.
We followed the same terminology as the latter one.

\section{Full Results of Open-vocabulary Object Retrieval}\label{sec:retrieve}

This section details the experiments in the main paper on open-vocabulary search in a 3D scenes database (Figure 4, Table 6). We employed our 2D-3D ensemble method to produce features for the Matterport3D test set's mesh vertices, using NYU160 as the labelset for ensembling. For each given search query, the vertices were sorted by cosine similarity to the CLIP text query embedding, resulting in a ranked retrieval list with one match per region.



When selecting the queries to use in the experiment, we limited our selections to raw category names provided with the ground truth of the dataset.   This allows us to reason about how many examples exist in the test set so that we can know how many matches to expect.   Among those candidate queries, we used two strategies to select ones to test: 1) we chose several of the most specific categories (e.g., ``yellow egg-shaped vase'') in order to test the method on the most difficult cases, and 2) we chose all raw categories with 15 ground truth examples in the entire Matterport3D dataset that had clear definitions (which excepted ``lounge chair,'' ``side table,'' and ``office table'') in order to avoid bias in the selection of queries (i.e., no cherry-picking).

\begin{figure*}[p]
        \centering
        \hfill{}
        \vspace{-1em}
         \includegraphics[width=0.95\textwidth]{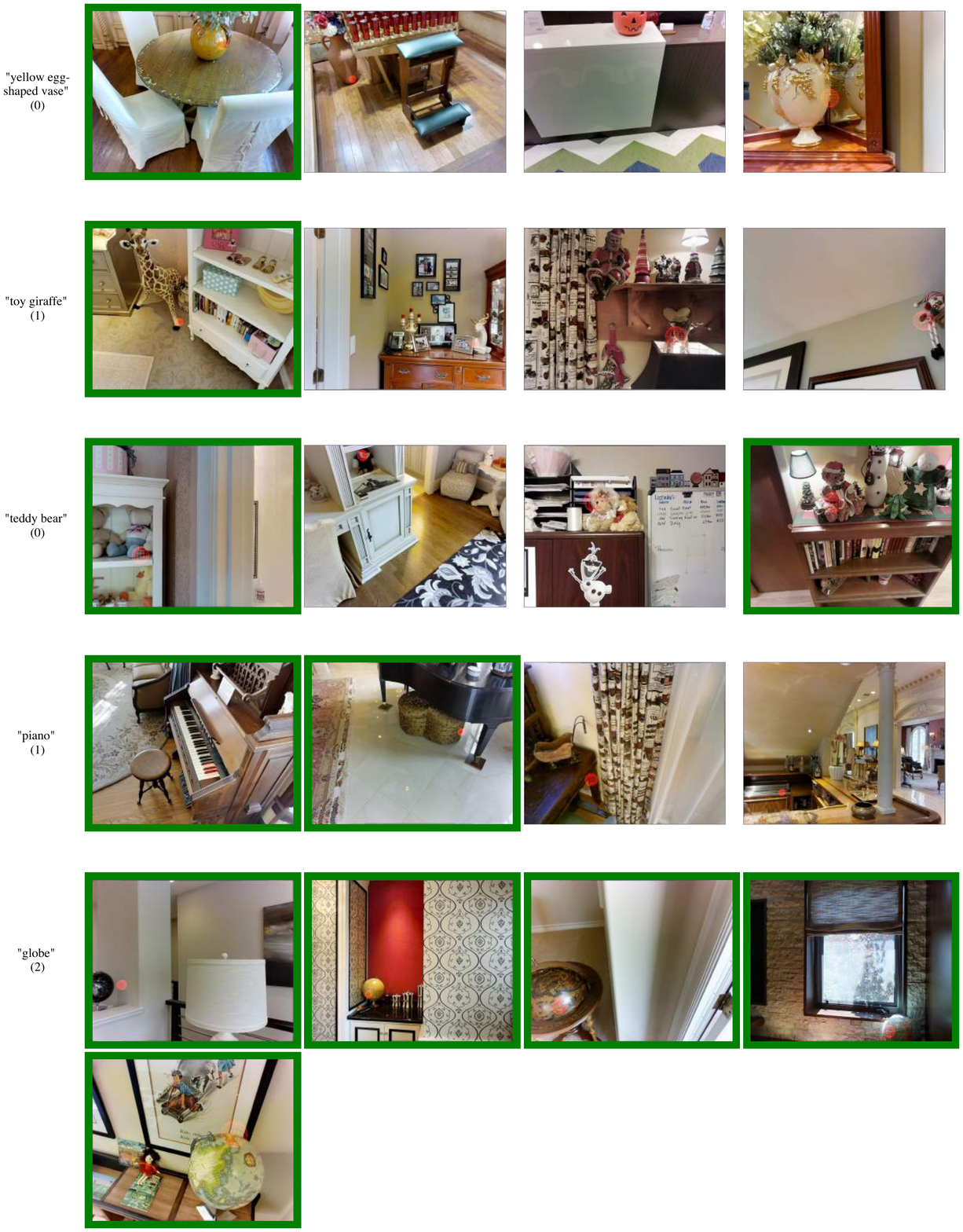}
         \caption{
        \textbf{Example object retrieval results (page 1 of 4)}.   The query text is in the left column, with the number of ground truth instances in the Matterport test set listed in parentheses below.   The images show top matching 3D points in the Matterport test set ranked from left to right (note the red wireframe sphere around the matching point in each image). Correct matches are marked with green borders.   The one incorrect match is marked with a red border (in page 3 of 4).   Others marked with gray borders are not wrong (since there are no further objects matching the query according to the ground truth), but are shown as examples of near matches.
        }
        \label{fig:object_queries1}
\end{figure*}

\begin{figure*}[p]
        \centering
        \setlength{\tabcolsep}{0.1em}
        \renewcommand{\arraystretch}{0.5}
        \hfill{}
         \includegraphics[width=0.95\textwidth]{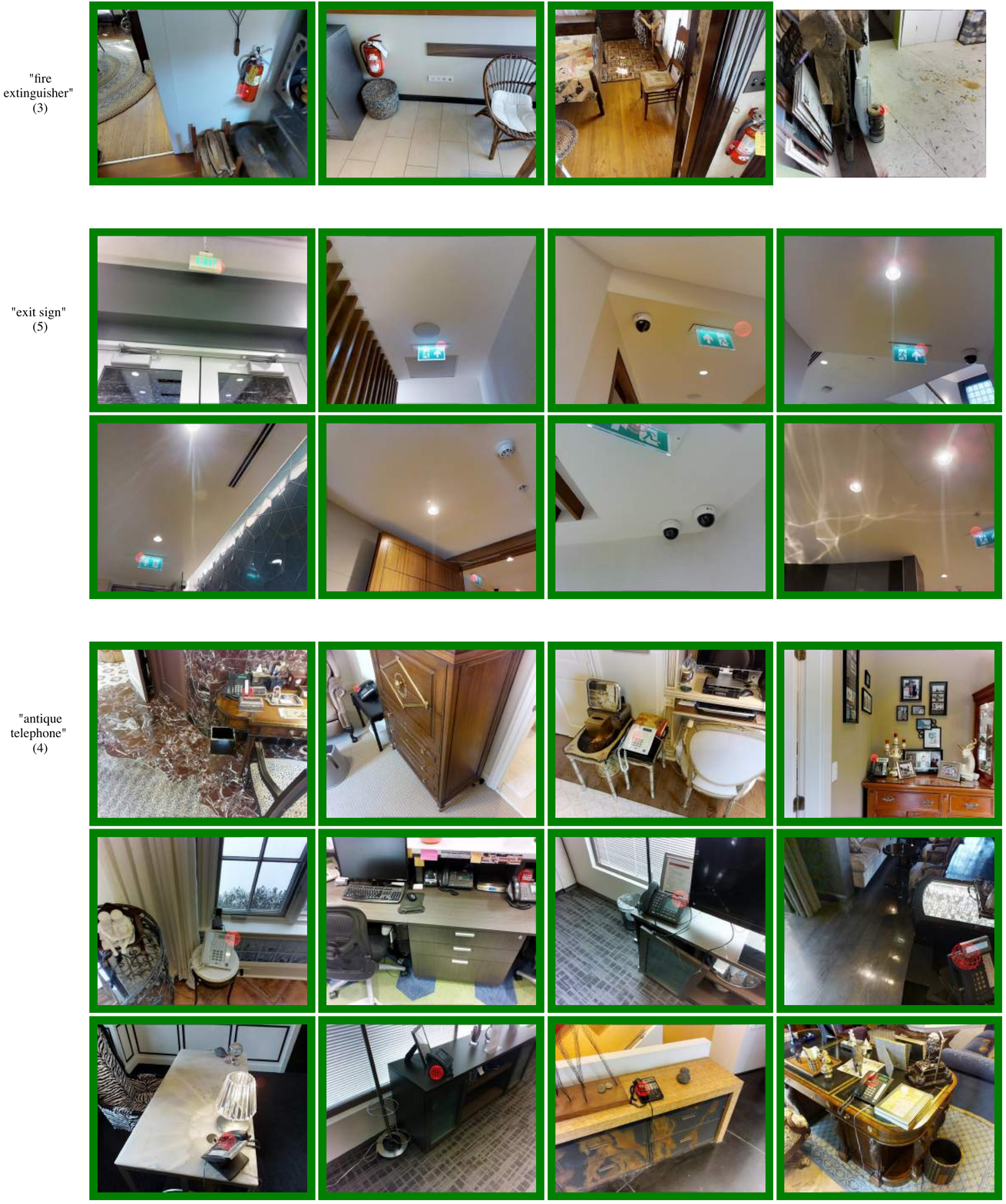}
         \caption{
        \textbf{Example object retrieval results (page 2 of 4)}.   See caption of Figure \ref{fig:object_queries1} for details.
        }
        \label{fig:object_queries2}
\end{figure*}

\begin{figure*}[p]
        \centering
        \setlength{\tabcolsep}{0.1em}
        \renewcommand{\arraystretch}{0.5}
        \hfill{}
         \includegraphics[width=0.95\textwidth]{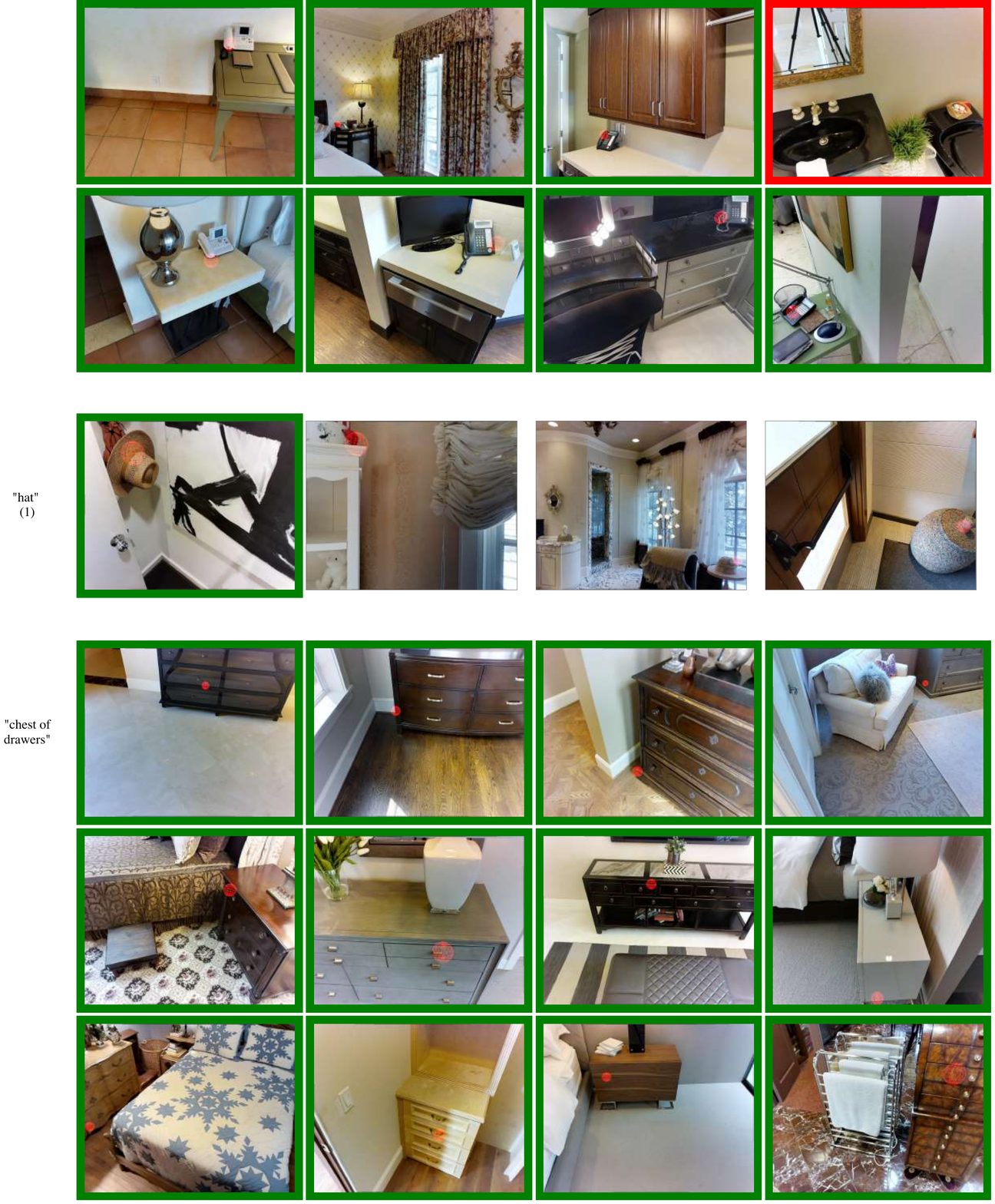}
         \caption{
        \textbf{Example object retrieval results (page 3 of 4)}.   See caption of Figure \ref{fig:object_queries1} for details.
        }
        \label{fig:object_queries3}
\end{figure*}

\begin{figure*}[p]
        \centering
        \setlength{\tabcolsep}{0.1em}
        \renewcommand{\arraystretch}{0.5}
        \hfill{}
         \includegraphics[width=0.95\textwidth]{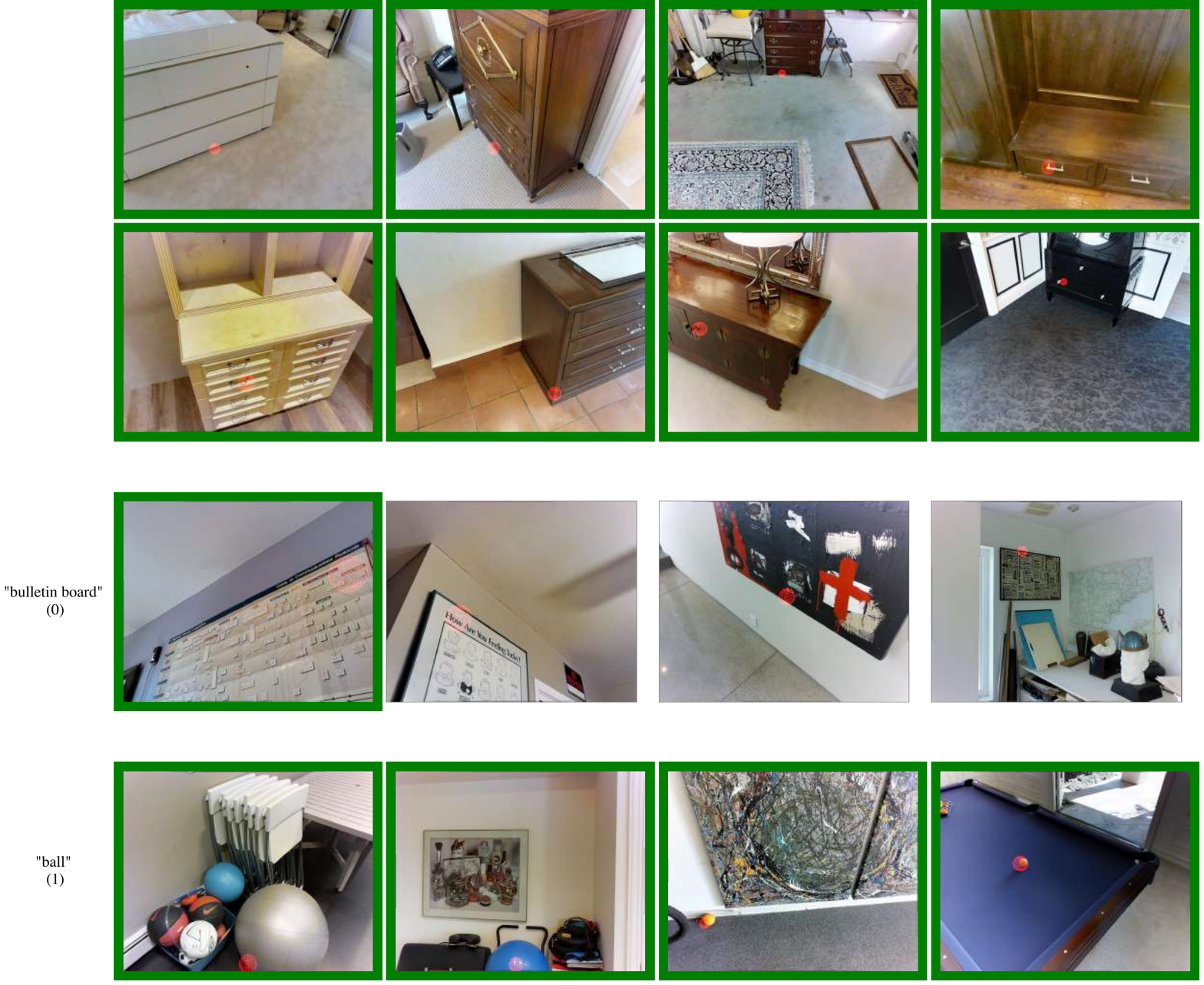}
         \caption{
        \textbf{Example object retrieval results (page 4 of 4)}.   See caption of Figure \ref{fig:object_queries1} for details.
        }
        \label{fig:object_queries4}
\end{figure*}

Figures \ref{fig:object_queries1}-\ref{fig:object_queries4} display ranked retrieval lists, with the best match first and expected matches in parentheses. A red wireframe sphere highlights the top match. Green/red-bordered images show correct/incorrect matches, with incorrect ones ranked lower than the last ground truth instance. Gray-bordered images demonstrate near misses, not expected as matches due to their rank exceeding labeled ground truth examples.

We can see that the algorithm is able to retrieve very specific objects from the database with great precision.   For example, when queried with ``yellow egg-shaped vase,'' its top match is indeed a match (which was not labeled in the ground truth), and the following retrieval results are tan vase, a pumpkin, and a white egg-shaped vase with gold decorations.   Similarly, when queried with ``teddy bear,'' it retrieves two teddy bears (neither labeled in the ground truth), a stuffed monkey, and a stuffed lion among the top four matches.   Among all the queries in all of the experiments, The only false positive occurred with ``telephone'' where a bowl of stones ranked 15th, while two ground truth instances ranked 25th and 29th. In this case, 29 of the top 30 matches were correct (20 are shown in Figure \ref{fig:object_queries3}).

These results suggest that the open-vocabulary features computed with our 2D-3D ensemble algorithm are very effective at retrieving object types with specific names.  Further experiments are required to understand the limitations.

\section{More Results of Open-vocabulary 3D Scene Exploration}\label{sec:explore}
\tablethreedssg

The main paper demonstrates that open-vocabulary queries can be used to explore the content of 3D scenes via text queries (Figures 1 and 6 of the main paper).   This section provides more examples demonstrating the power of open-vocabulary exploration.  Please see the supplemental video for a live demo.

For each query, the user types a text string (e.g., ``glass''), it gets encoded with the pre-trained CLIP text encoder, we compute the cosine similarity with features we've computed for every 3D vertex, and then color the vertices by similarity (yellow is high, green is middle, blue is low).

Figures \ref{fig:object_queries}-\ref{fig:abstract_queries} show results for a broad range of queries, including ones that describe object categories in~\figref{fig:object_queries}, room types in~\figref{fig:room_type_queries}, activities in~\figref{fig:activity_queries},  colors in \figref{fig:color_queries}, materials in~\figref{fig:material_queries}, and abstract concepts in~\figref{fig:abstract_queries}.

Please note the power of using language models learned via CLIP to reason about scene attributes and abstract concepts that would be difficult to label in a supervised setting.   For example, searching for ``store'' highlights 3D points mainly on closets and cabinets (middle-right of \figref{fig:activity_queries}), and searching for ``cluttered'' yields points in a particularly busy closet (top-right of \figref{fig:abstract_queries}).  These examples demonstrate the power of the proposed approach for scene \emph{understanding}, which goes far beyond semantic segmentation.

\boldparagraph{Quantitative results on 3DSSG dataset~\cite{Wald2020CVPR}}
To evaluate 3D scene exploration performance, we conduct an experiment on the 3DSSG dataset that has annotations in object-level material estimation.
In~\tabref{tab:3dssg}, we compare material class predictions for the 3DSSG test set using variants of our approach trained on ScanNet and a fully-supervised MinkowskiNet.
Findings align with the paper: 1) 2D-3D ensembling is our best variant, 2) it underperforms fully-supervised methods for classes with abundant examples, and 3) it excels for classes with fewer examples.
\figensembleratio

\newcommand{\eewidth}{0.48\textwidth}
\begin{figure*}[t]
        \centering
        \setlength{\tabcolsep}{0.1em}
        \renewcommand{\arraystretch}{0.5}
        \hfill{}
        \begin{tabular}{cc}
            \includegraphics[width=\eewidth]{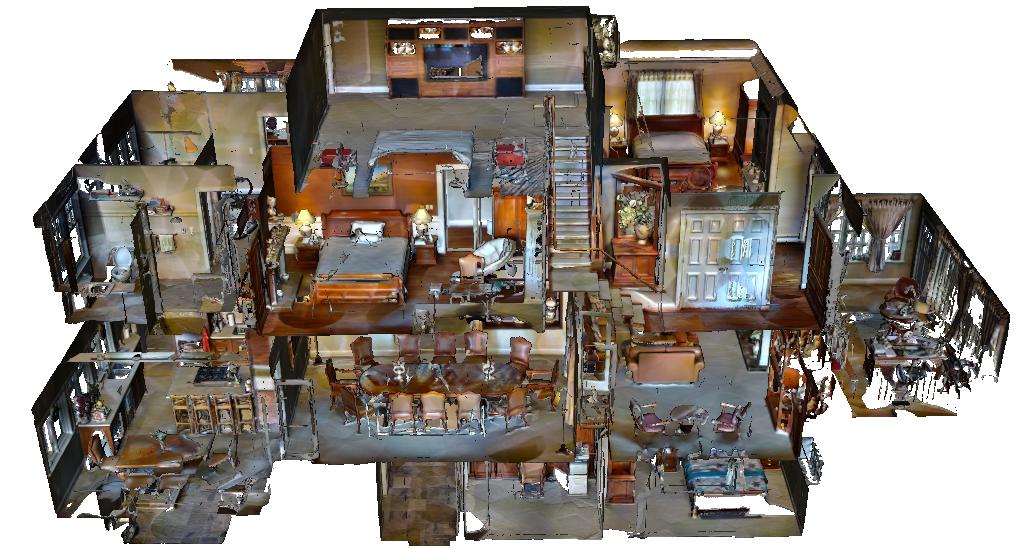} &
            \includegraphics[width=\eewidth]{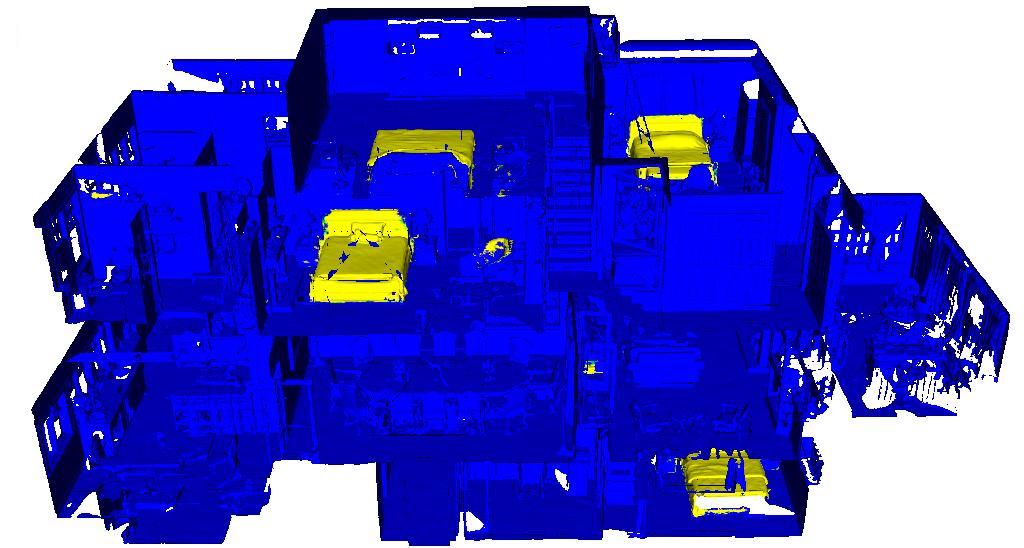} \\
            Input 3D Geometry & ``bed''\vspace{1em}\\
              \includegraphics[width=\eewidth]{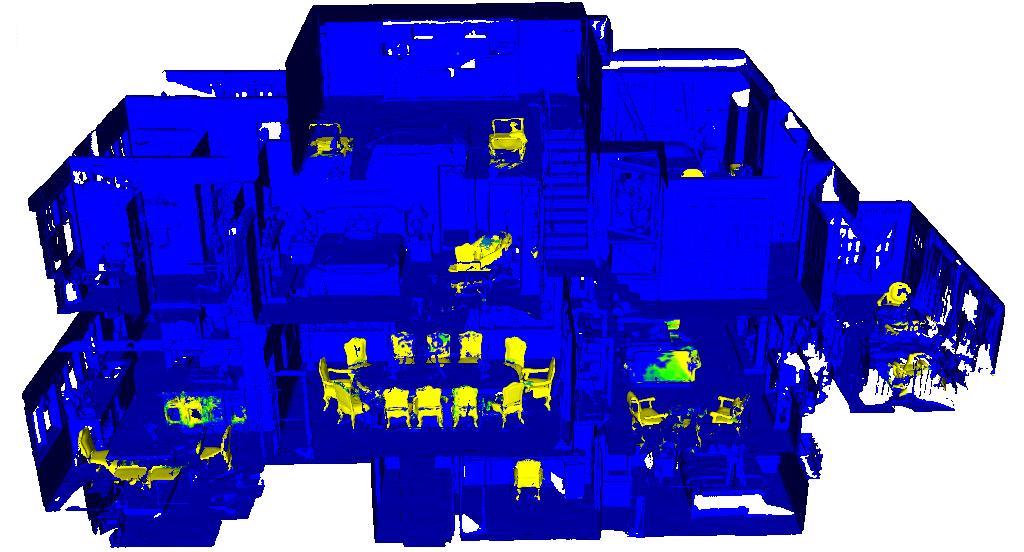} &
             \includegraphics[width=\eewidth]{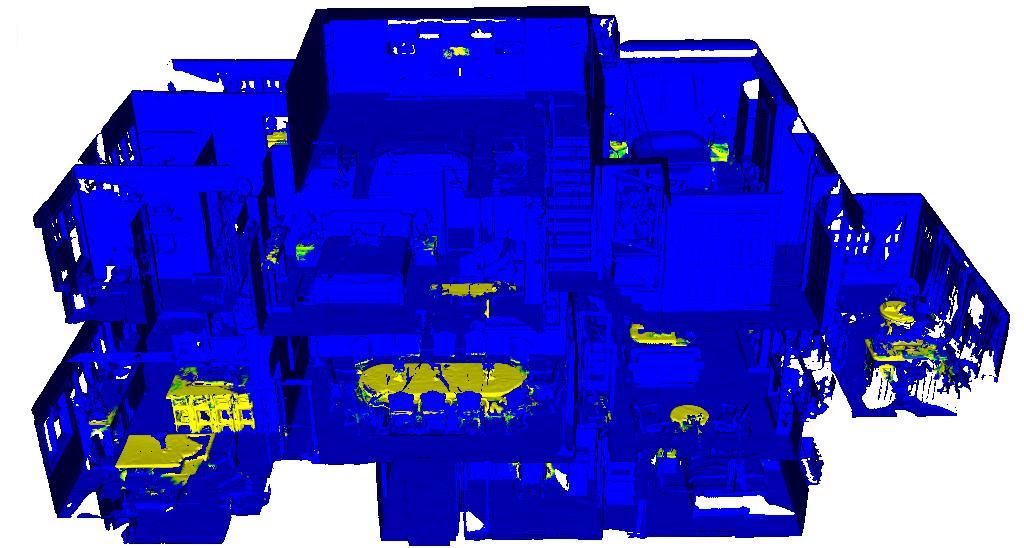} \\
             ``chair'' & ``table''\vspace{1em}\\
             \includegraphics[width=\eewidth]{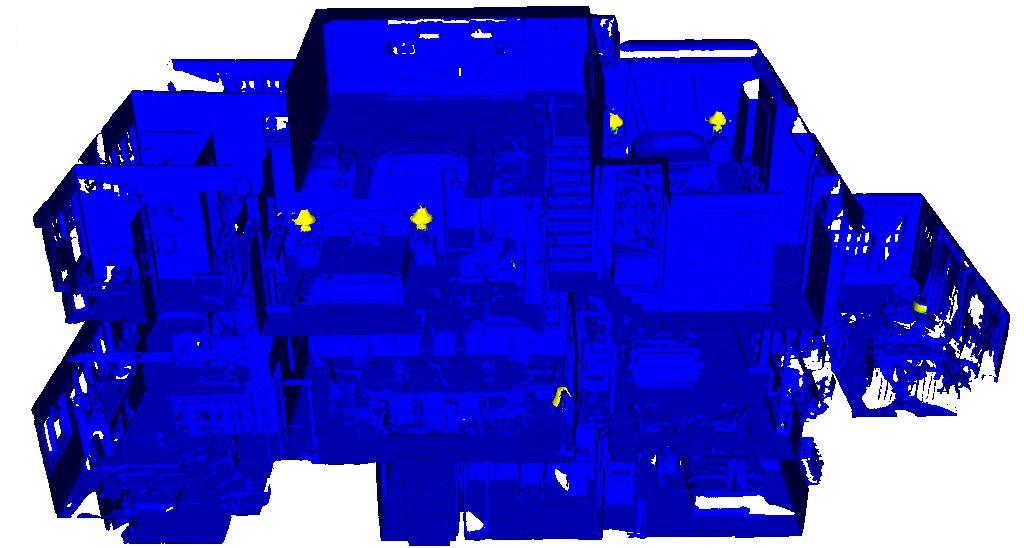} &
               \includegraphics[width=\eewidth]{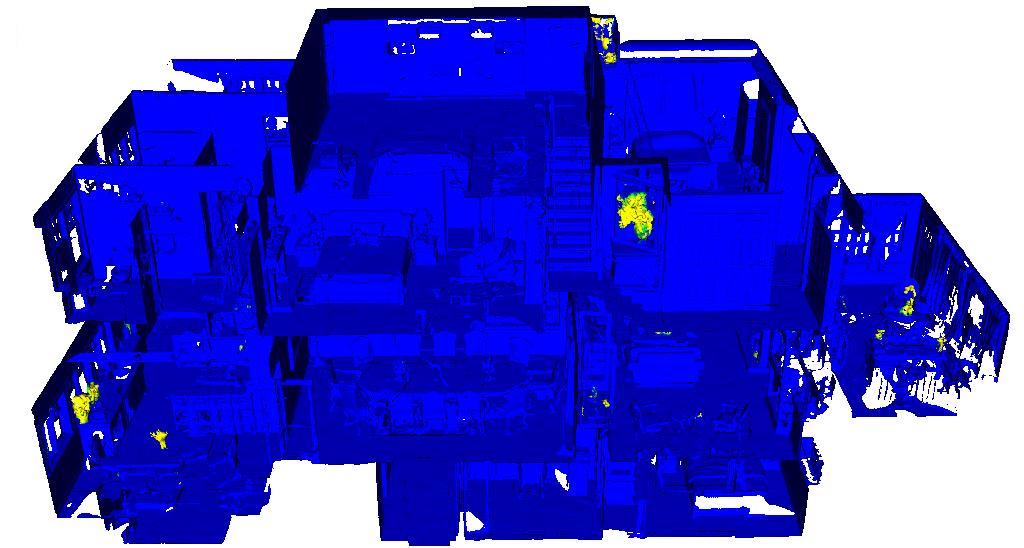} \\
               ``lamp'' & ``plant''\vspace{1em}\\
         \end{tabular}
        \caption{
        \textbf{Open-Vocabulary Queries for Common Object Types.}
        }
        \label{fig:object_queries}
\end{figure*}

\begin{figure*}[t]
        \centering
        \setlength{\tabcolsep}{0.1em}
        \renewcommand{\arraystretch}{0.5}
        \hfill{}
        \begin{tabular}{cc}
            \includegraphics[width=\eewidth]{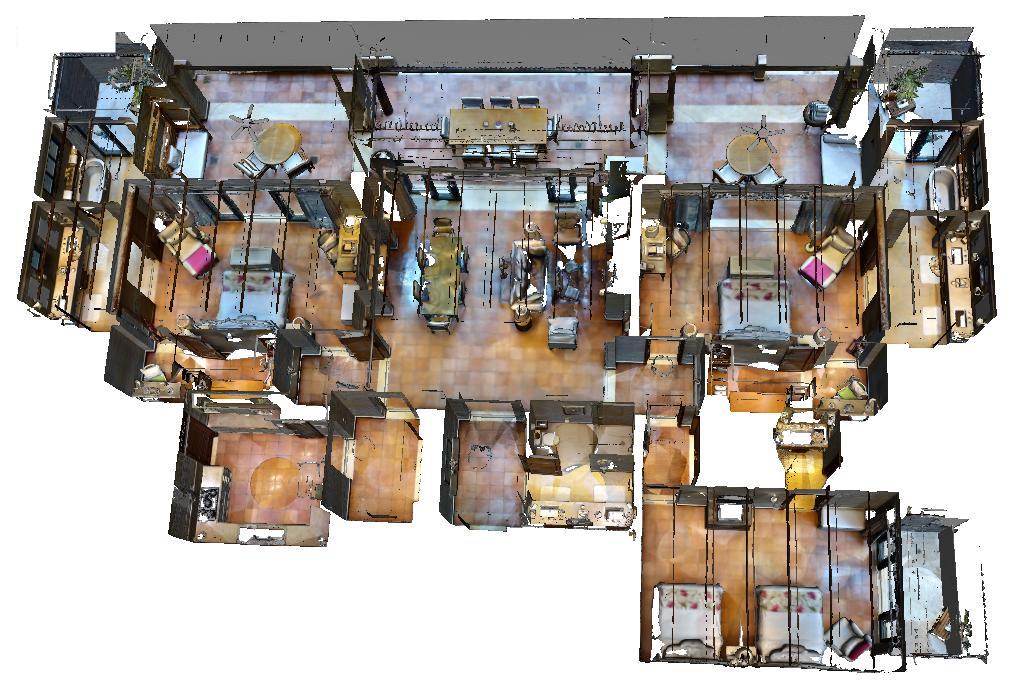} &
            \includegraphics[width=\eewidth]{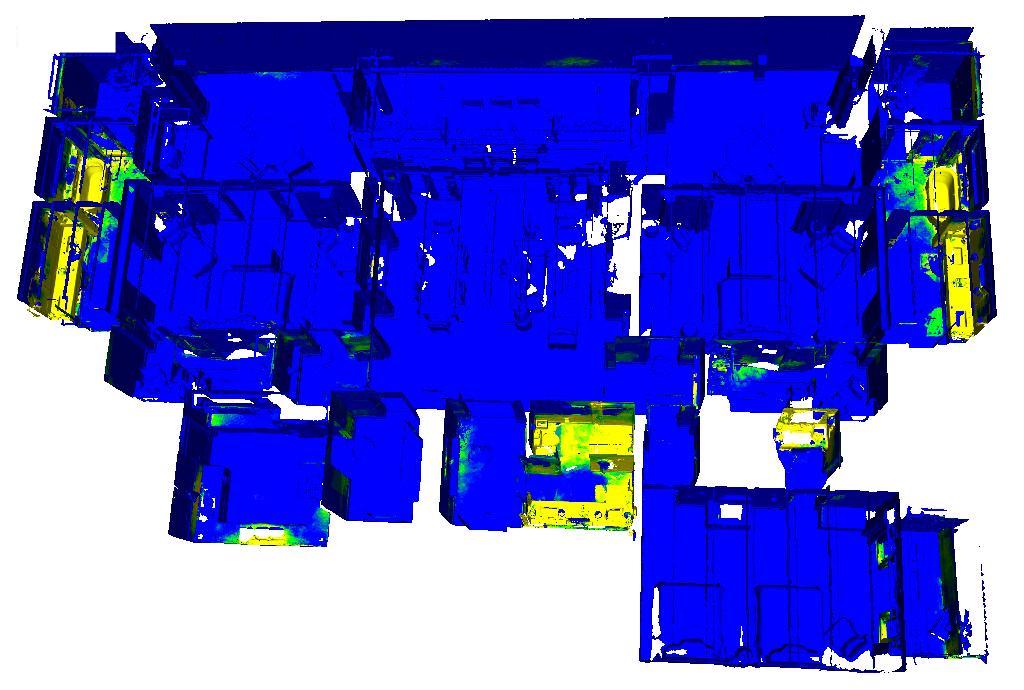} \\
            Input 3D Geometry & ``bathroom''\vspace{1em}\\
              \includegraphics[width=\eewidth]{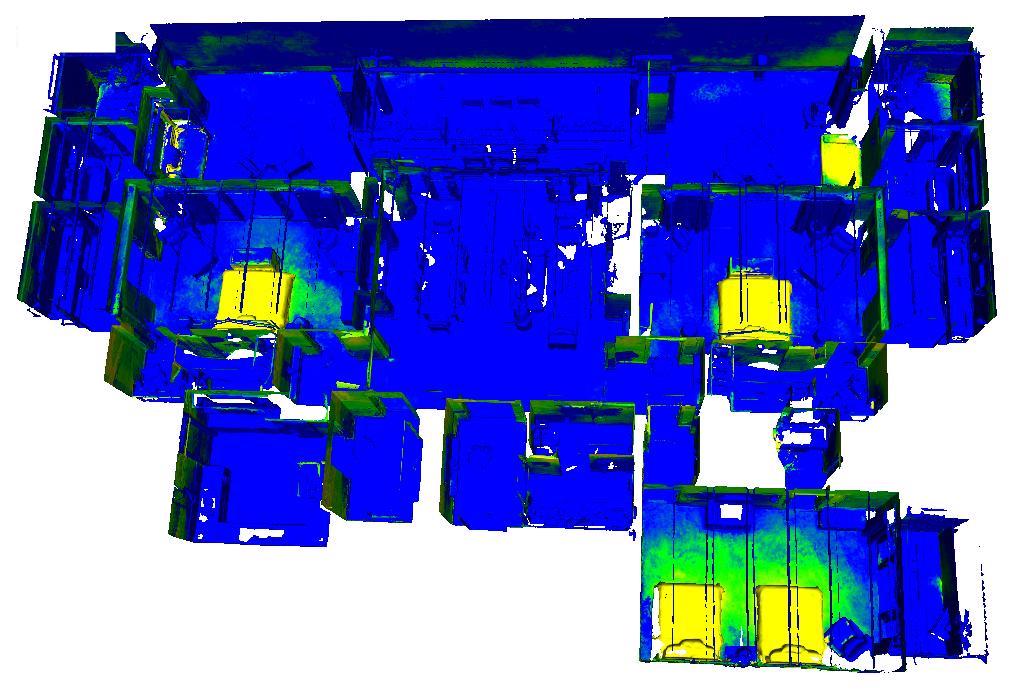} &
             \includegraphics[width=\eewidth]{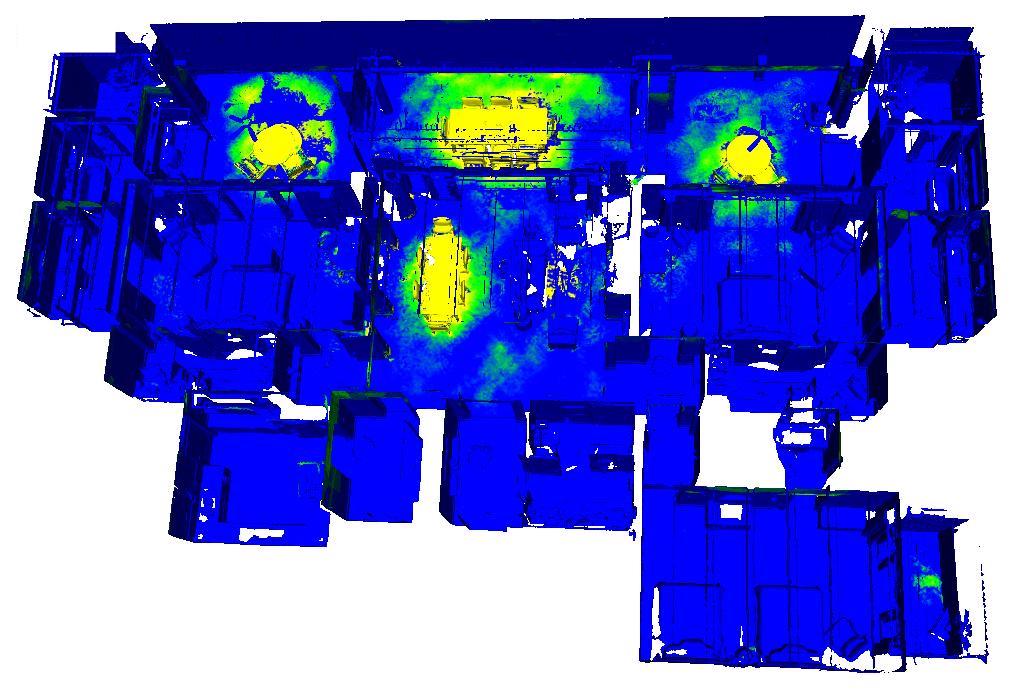} \\
             ``bedroom'' & ``dining room''\vspace{1em}\\
             \includegraphics[width=\eewidth]{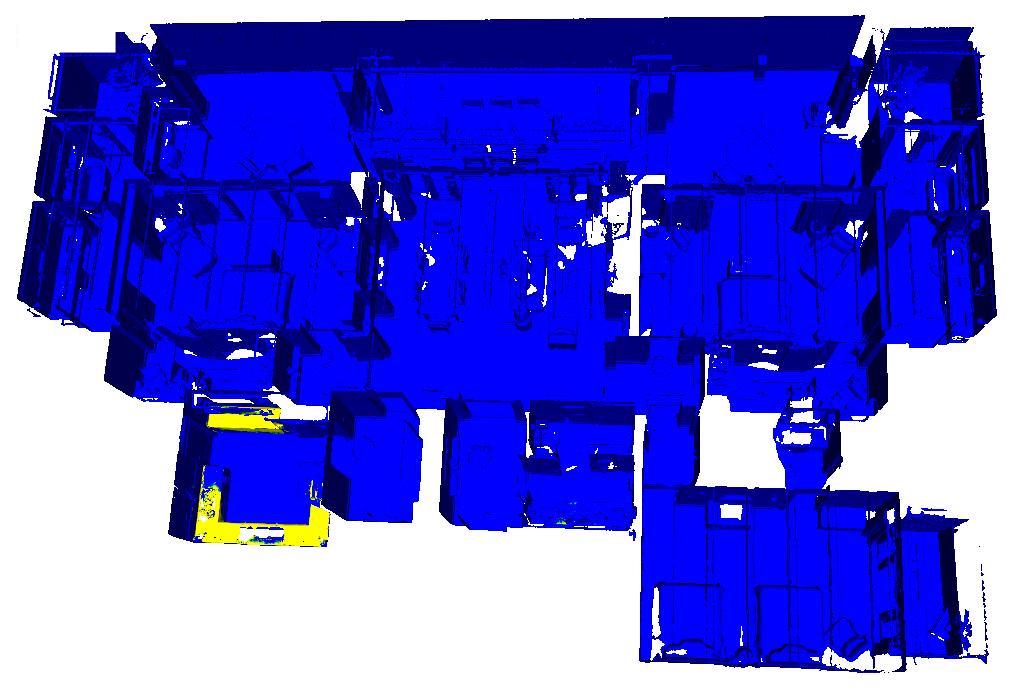} &
               \includegraphics[width=\eewidth]{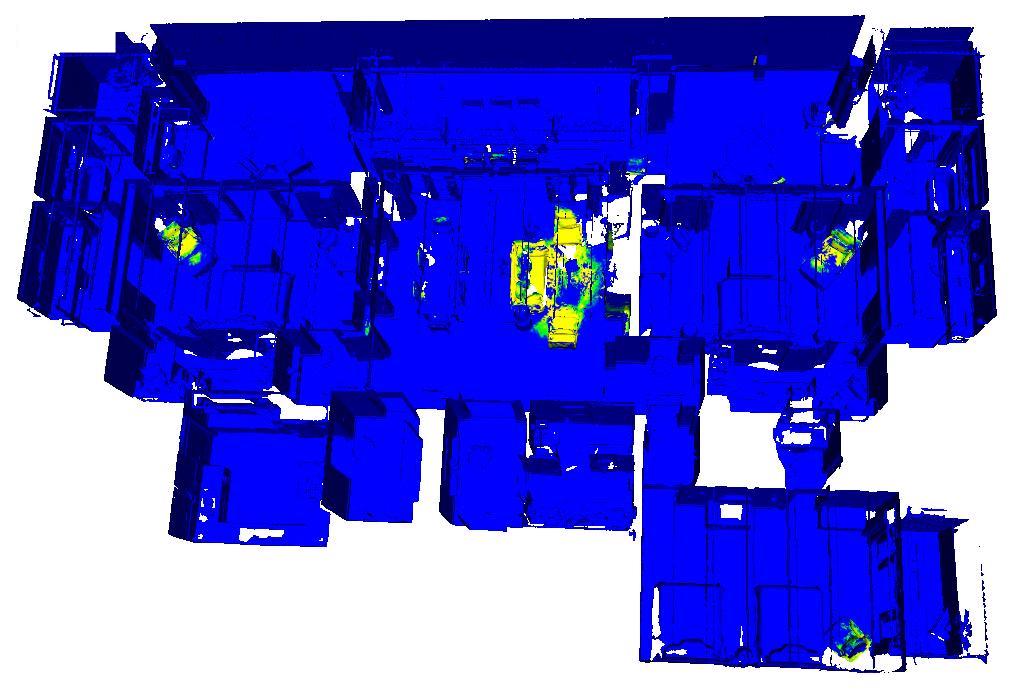} \\
               ``kitchen'' & ``living room''\vspace{1em}\\
         \end{tabular}
        \caption{
        \textbf{Open-Vocabulary Queries for Room Types.}
        }
        \label{fig:room_type_queries}
\end{figure*}

\begin{figure*}[t]
        \centering
        \setlength{\tabcolsep}{0.1em}
        \renewcommand{\arraystretch}{0.5}
        \hfill{}
        \begin{tabular}{cc}
            \includegraphics[width=\eewidth]{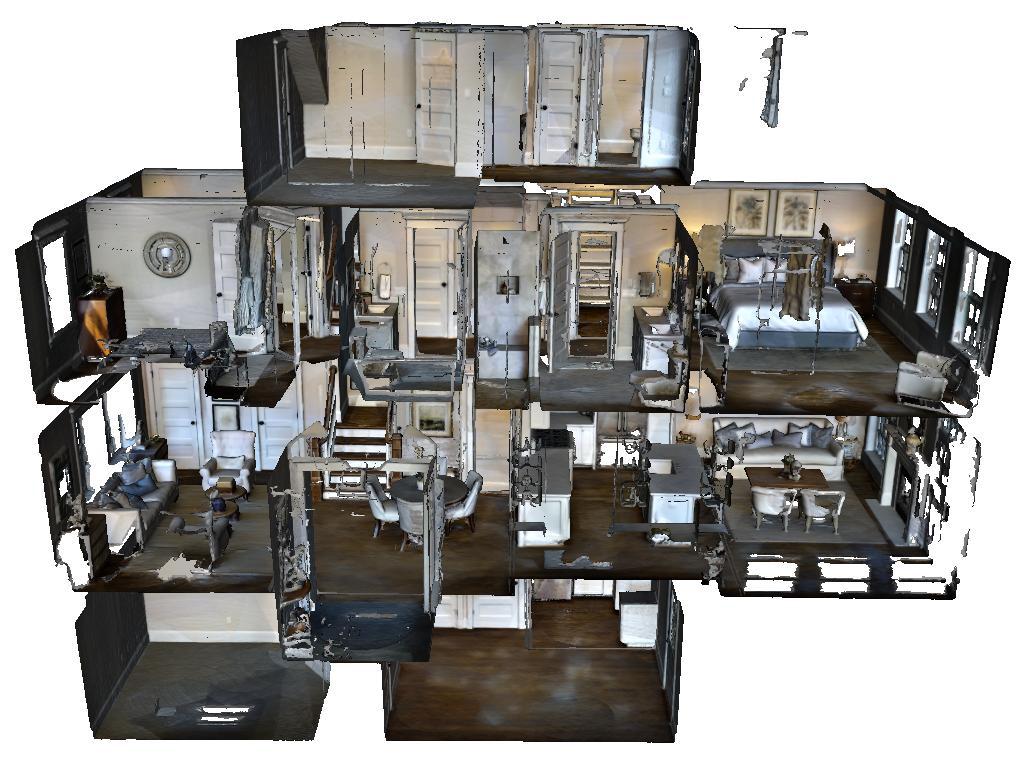} &
             \includegraphics[width=\eewidth]{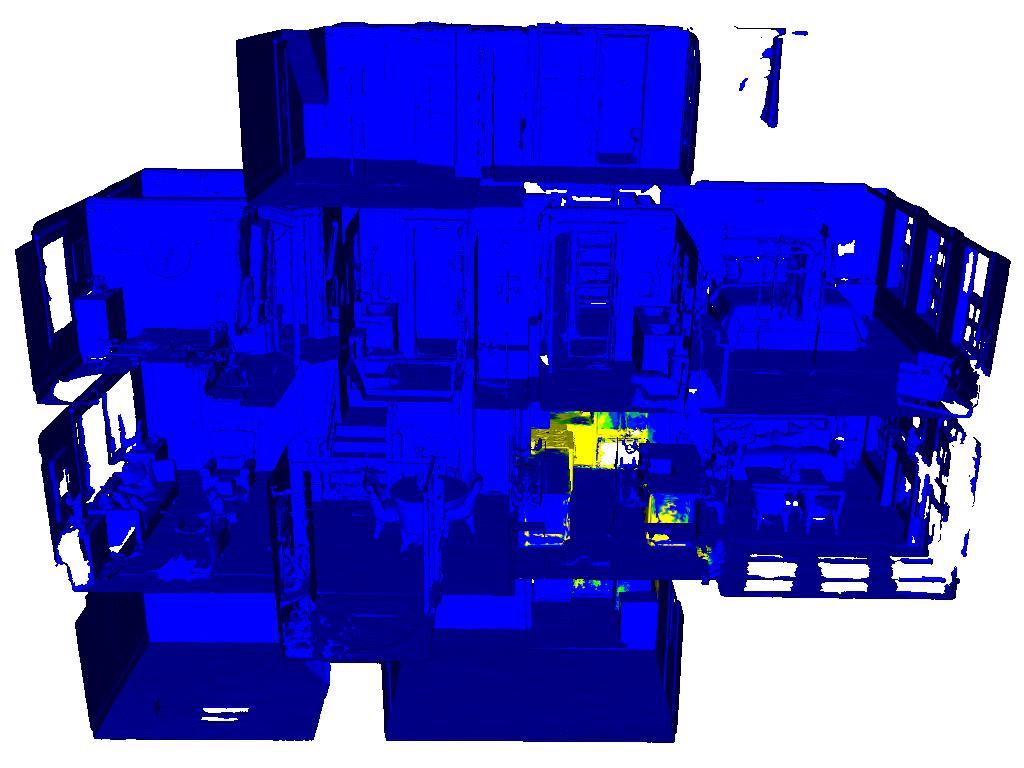} \\
             Input 3D Geometry & ``cook''\vspace{1em}\\
              \includegraphics[width=\eewidth]{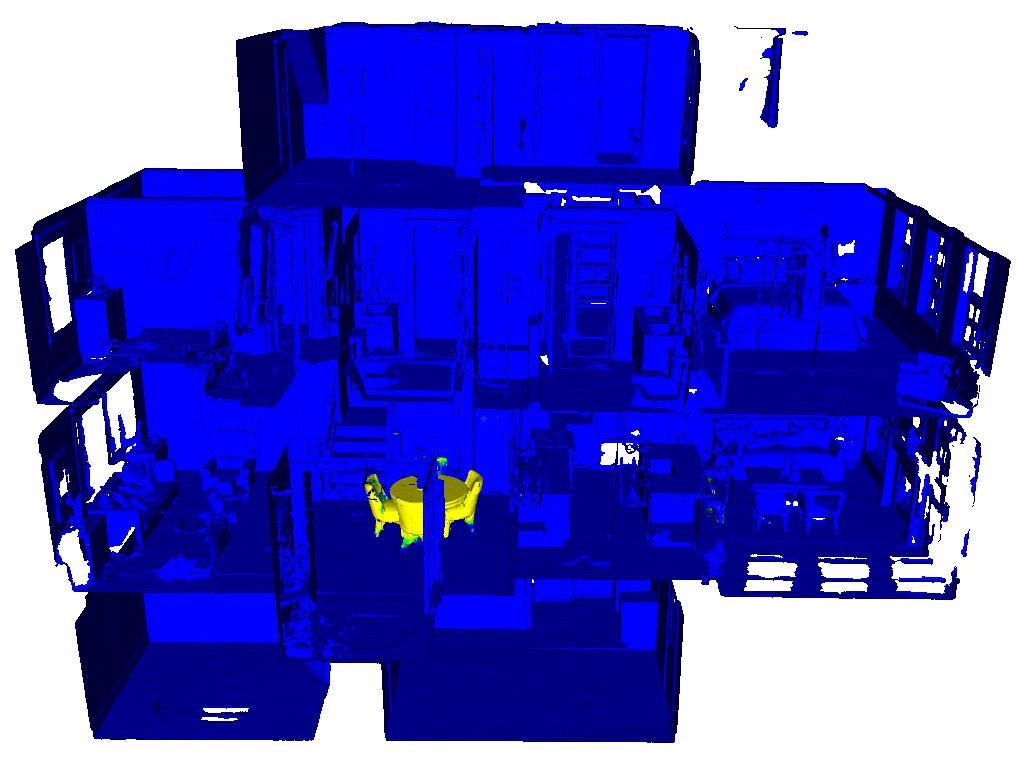} &
             \includegraphics[width=\eewidth]{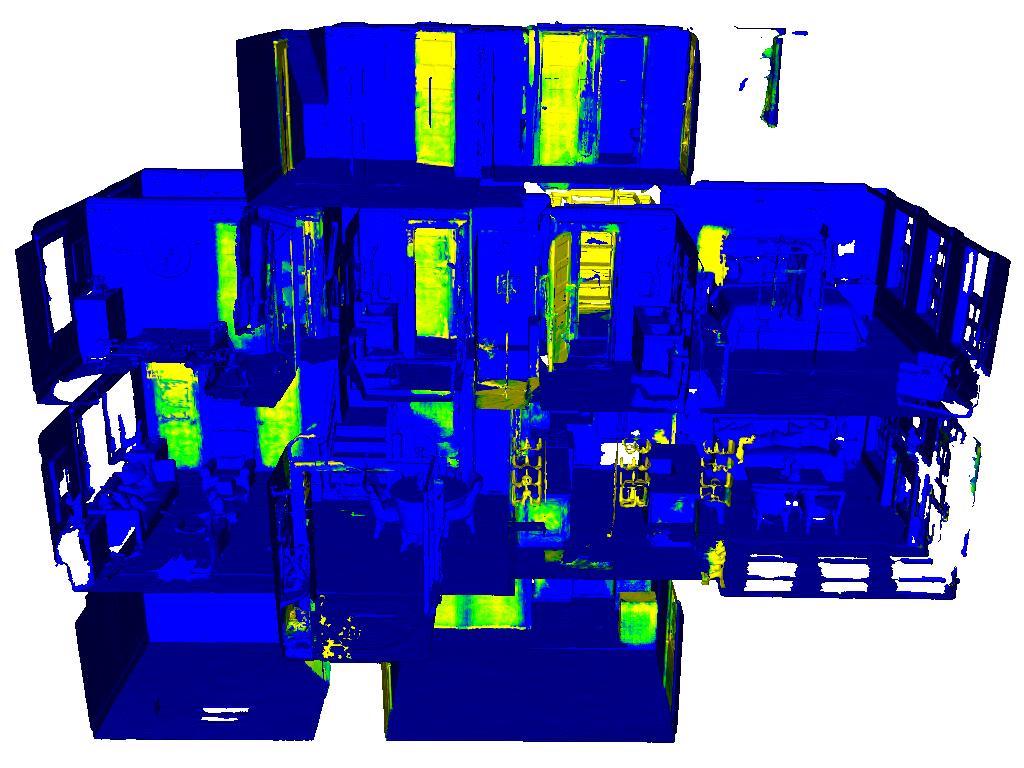} \\
             ``dine'' & ``store''\vspace{1em}\\
             \includegraphics[width=\eewidth]{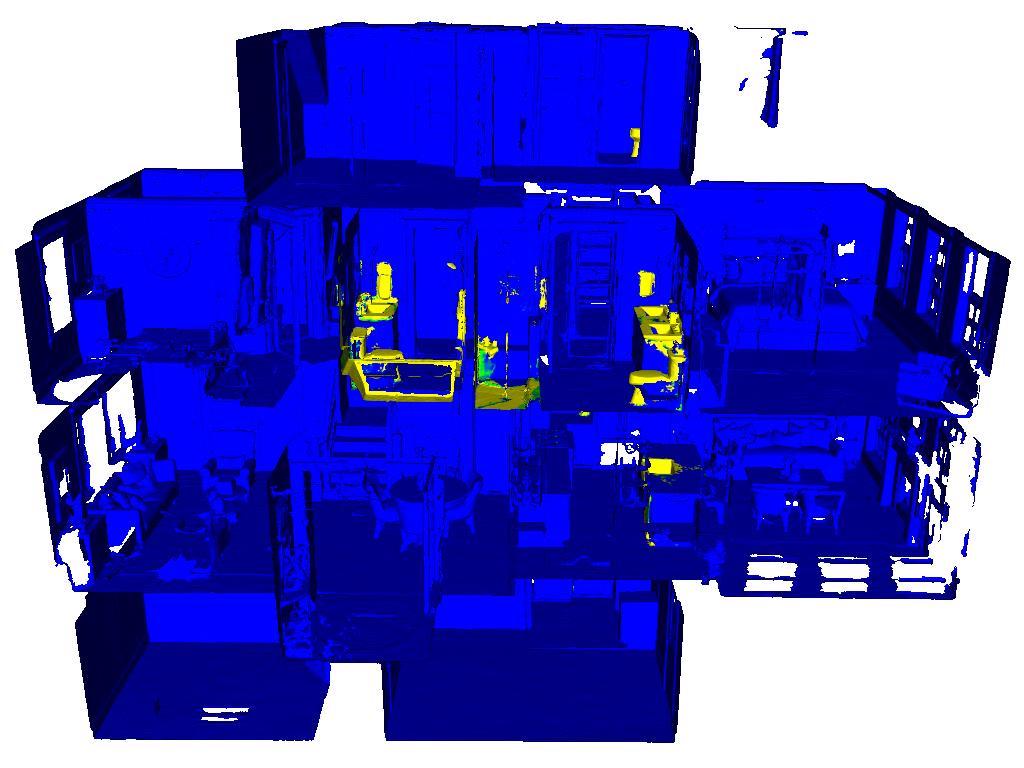} &
             \includegraphics[width=\eewidth]{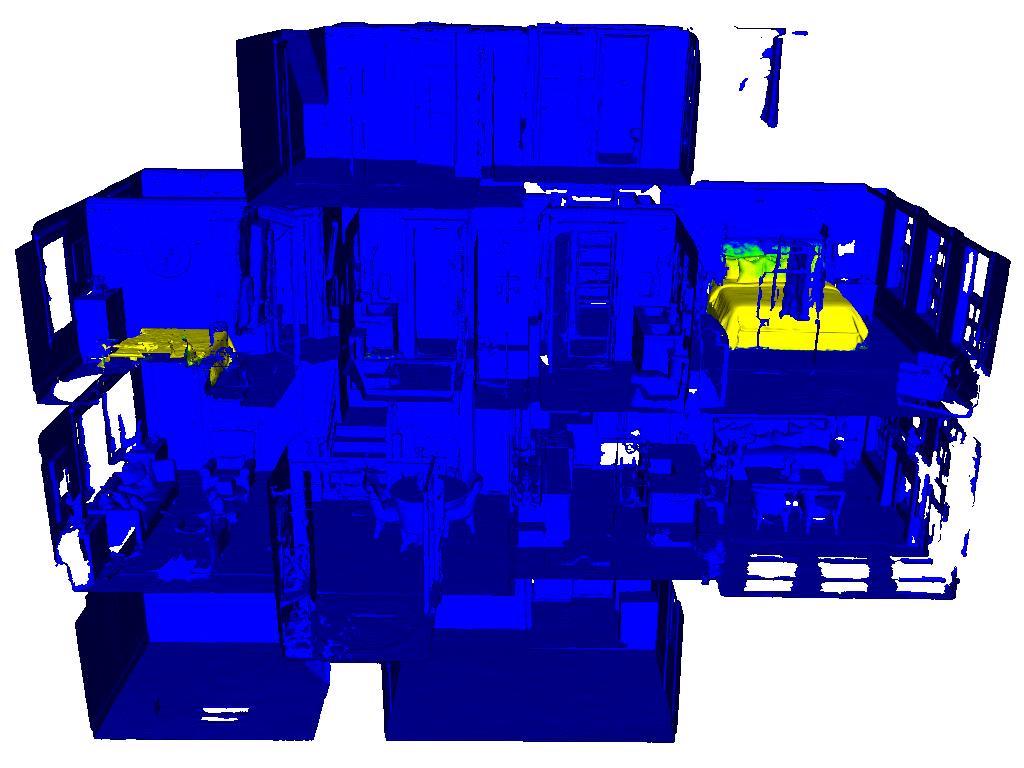} \\
             ``wash'' & ``sleep''\vspace{1em}\\
         \end{tabular}
        \caption{
        \textbf{Open-Vocabulary Queries for Activity Sites.}
        }
        \label{fig:activity_queries}
\end{figure*}

\begin{figure*}[t]
        \centering
        \setlength{\tabcolsep}{0.1em}
        \renewcommand{\arraystretch}{0.5}
        \hfill{}
        \begin{tabular}{cc}
            \includegraphics[width=\eewidth]{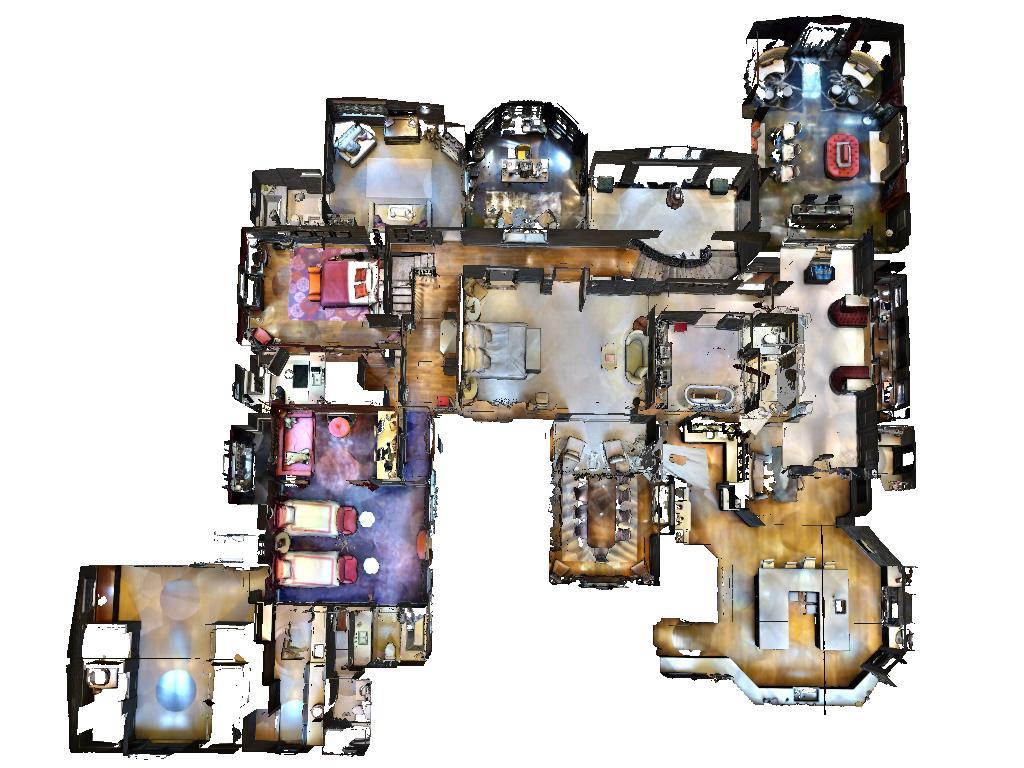} &
            \includegraphics[width=\eewidth]{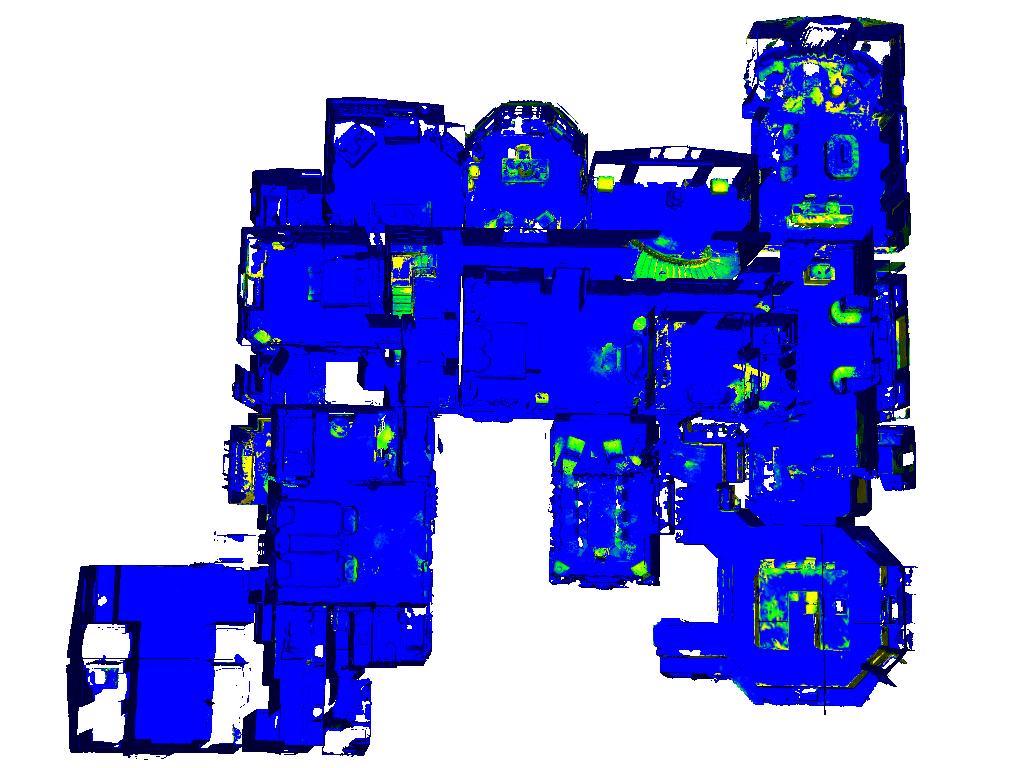} \\
            Input 3D Geometry & ``black''\vspace{1em}\\
            \includegraphics[width=\eewidth]{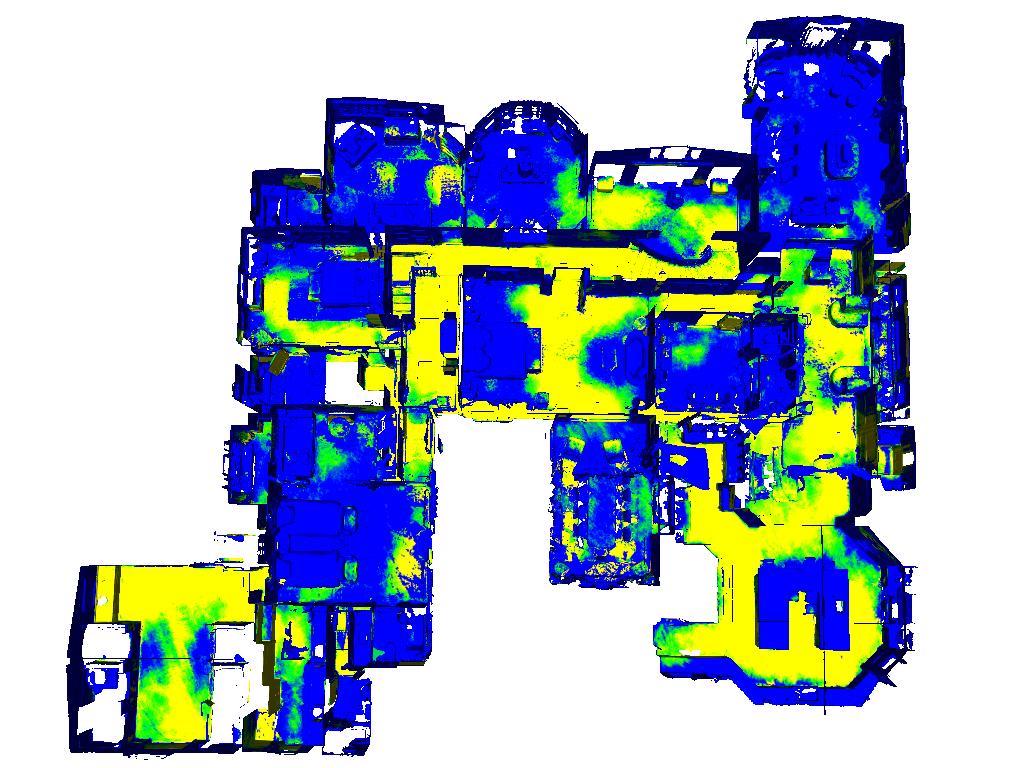} &
            \includegraphics[width=\eewidth]{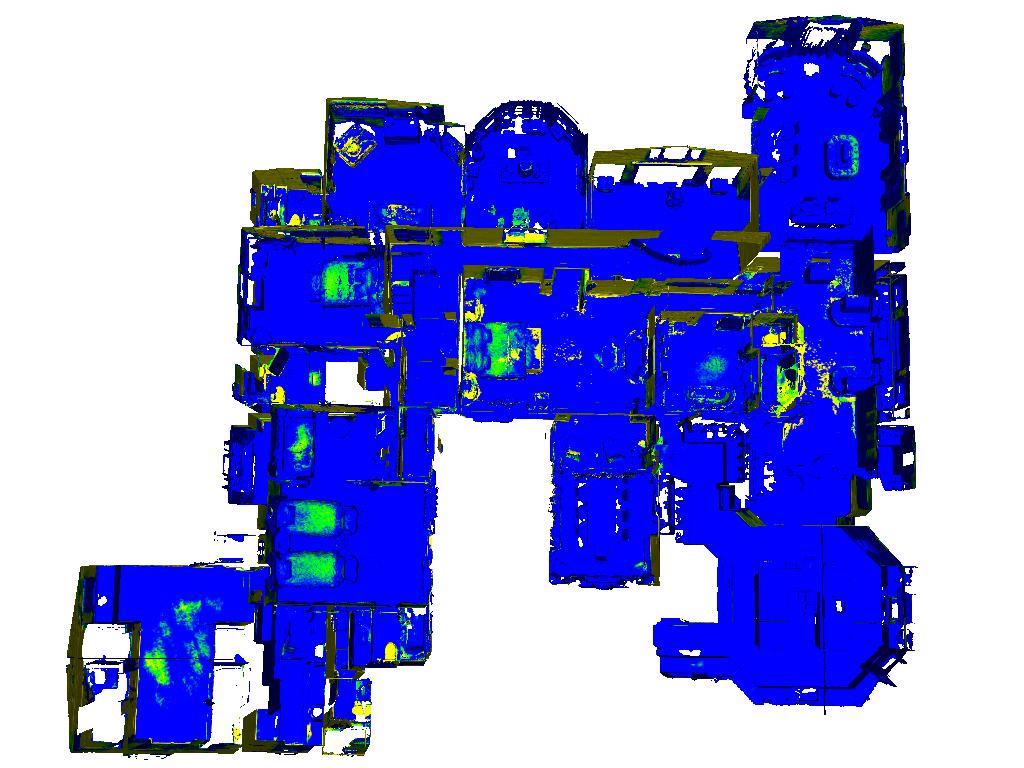} \\
            ``brown'' & ``white''\vspace{1em}\\
            \includegraphics[width=\eewidth]{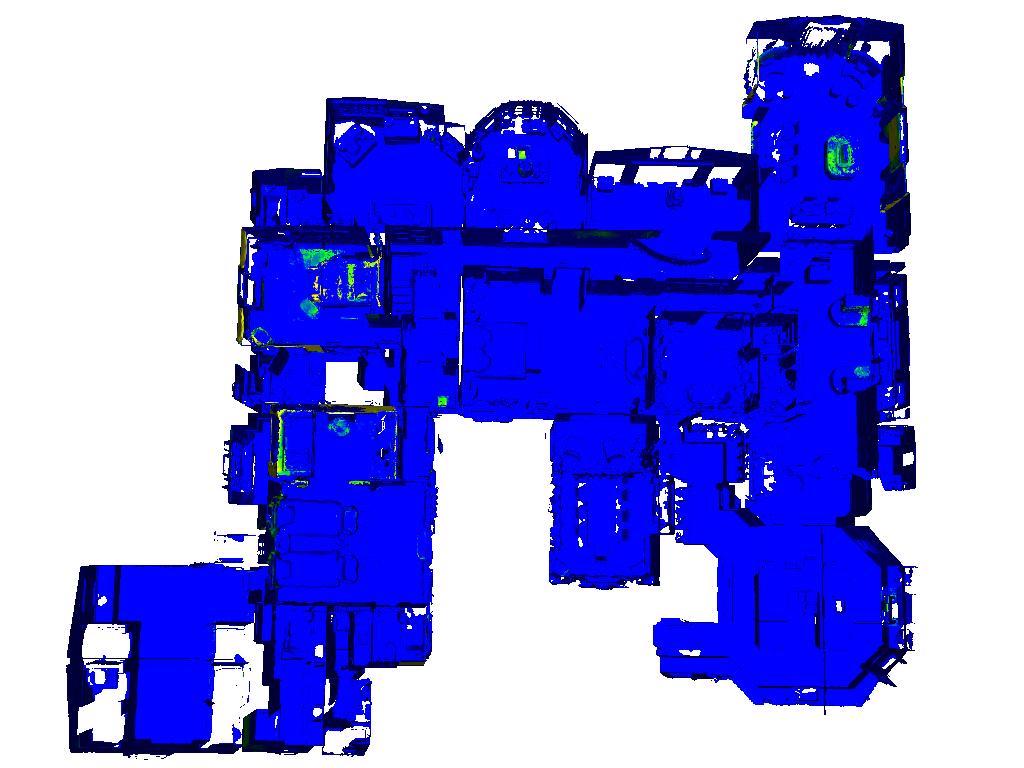} &
            \includegraphics[width=\eewidth]{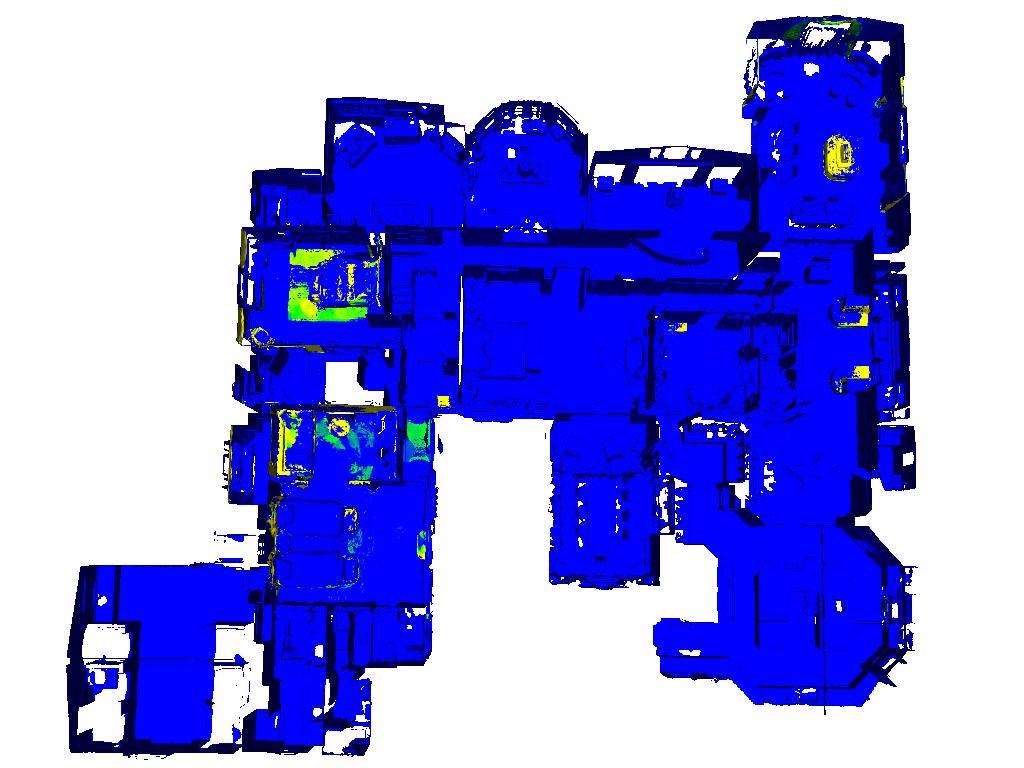} \\
            ``orange'' & ``red''\vspace{1em}\\
        \end{tabular}
        \caption{
        \textbf{Open-Vocabulary Queries for Colors.}
        }
        \label{fig:color_queries}
\end{figure*}

\begin{figure*}[t]
        \centering
        \setlength{\tabcolsep}{0.1em}
        \renewcommand{\arraystretch}{0.5}
        \hfill{}
        \begin{tabular}{cc}
            \includegraphics[width=\eewidth]{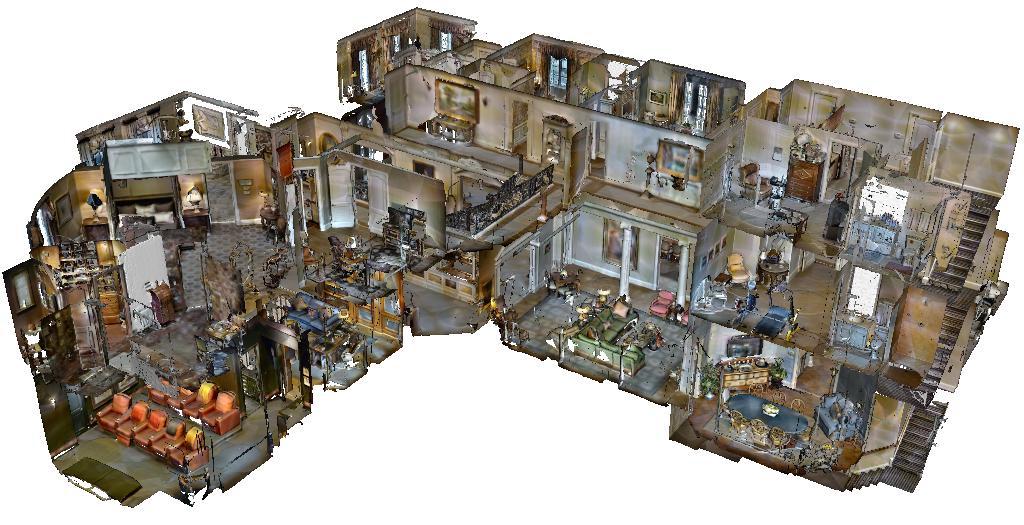} &
            \includegraphics[width=\eewidth]{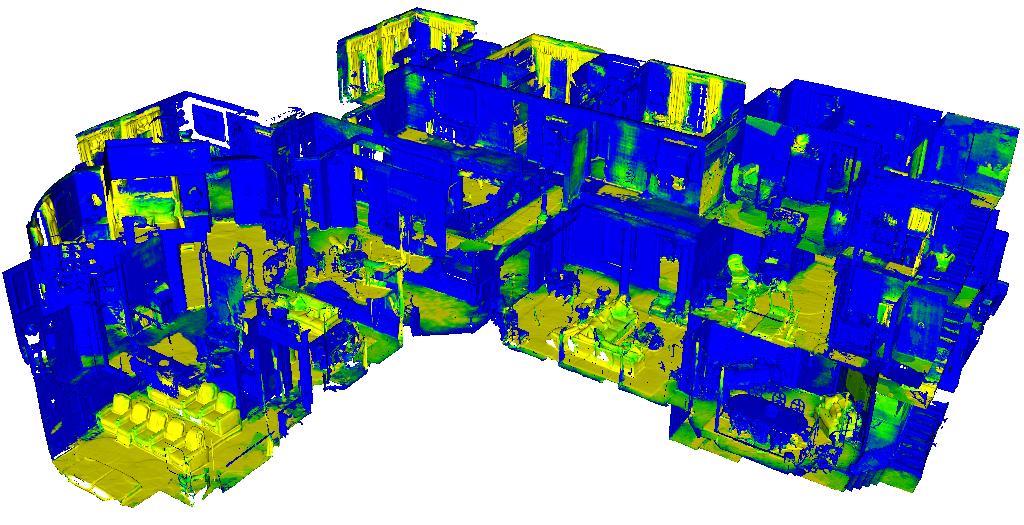} \\
            Input 3D Geometry & ``fabric''\vspace{1em}\\
              \includegraphics[width=\eewidth]{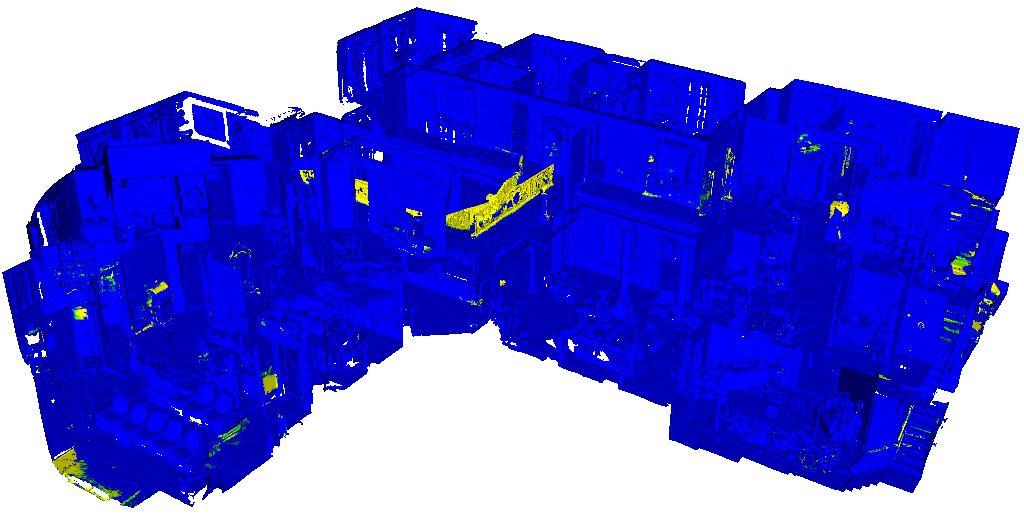} &
             \includegraphics[width=\eewidth]{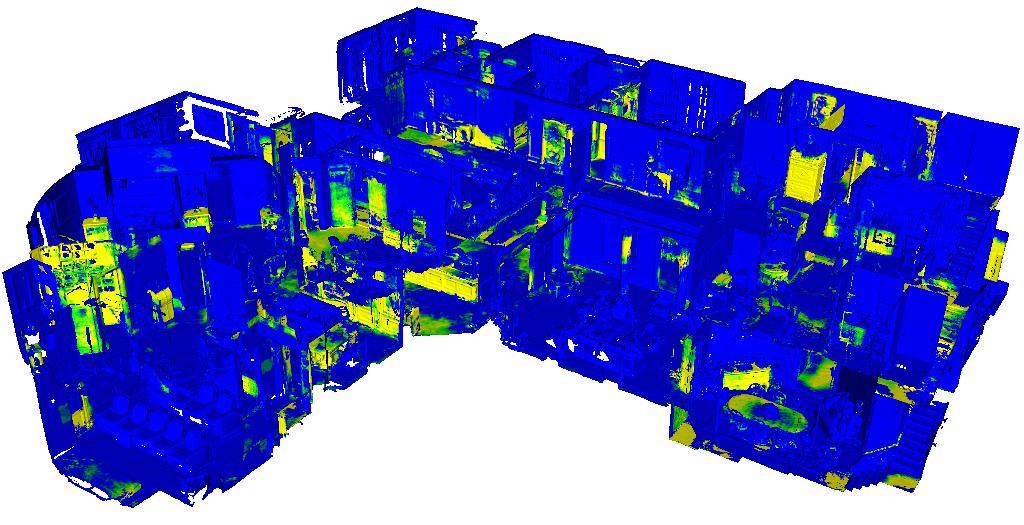} \\
             ``metal'' & ``wood''\vspace{1em}\\
         \end{tabular}
        \caption{
        \textbf{Open-Vocabulary Queries for Materials.}
        }
        \label{fig:material_queries}
\end{figure*}

\begin{figure*}[t]
        \centering
        \setlength{\tabcolsep}{0.1em}
        \renewcommand{\arraystretch}{0.5}
        \begin{tabular}{cc}
            \includegraphics[width=\eewidth]{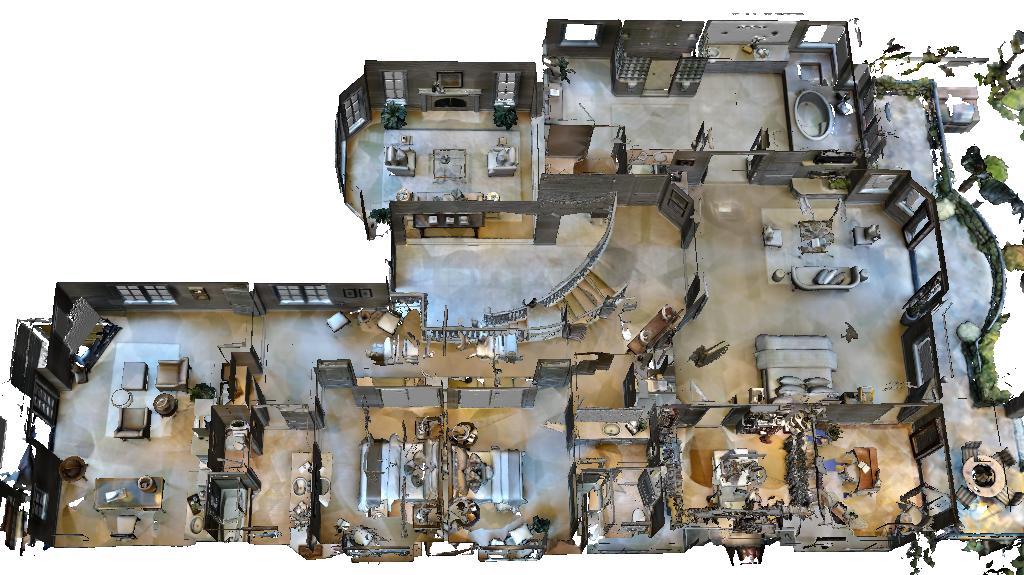} &
            \includegraphics[width=\eewidth]{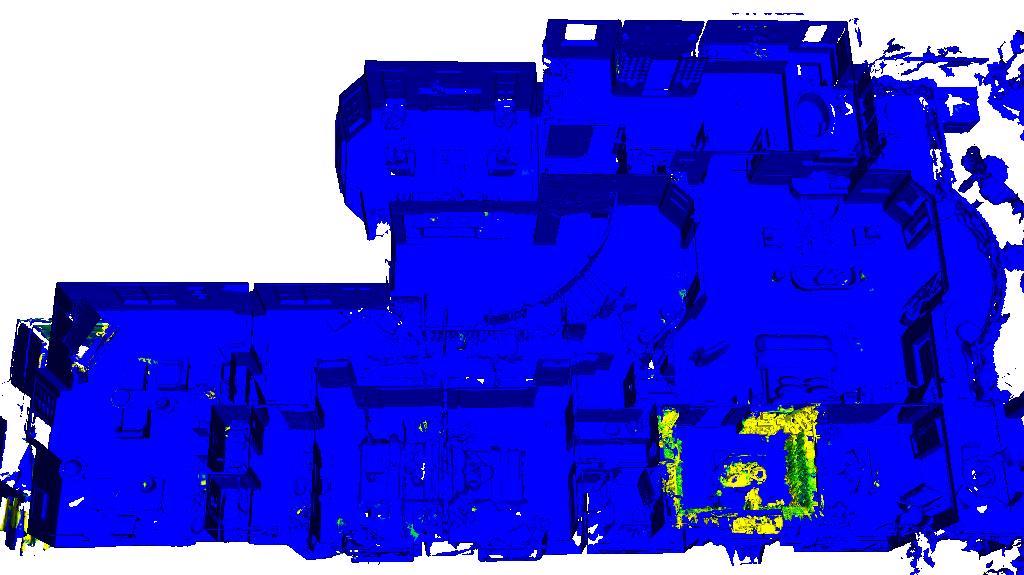} \\
            Input 3D Geometry & ``cluttered''\vspace{1em}\\
            \includegraphics[width=\eewidth]{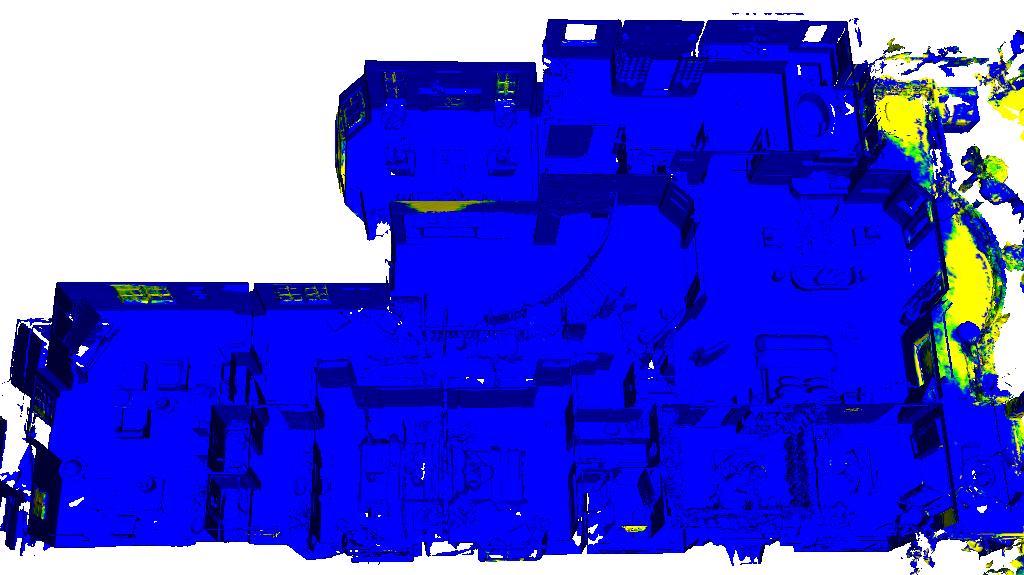} &
            \includegraphics[width=\eewidth]{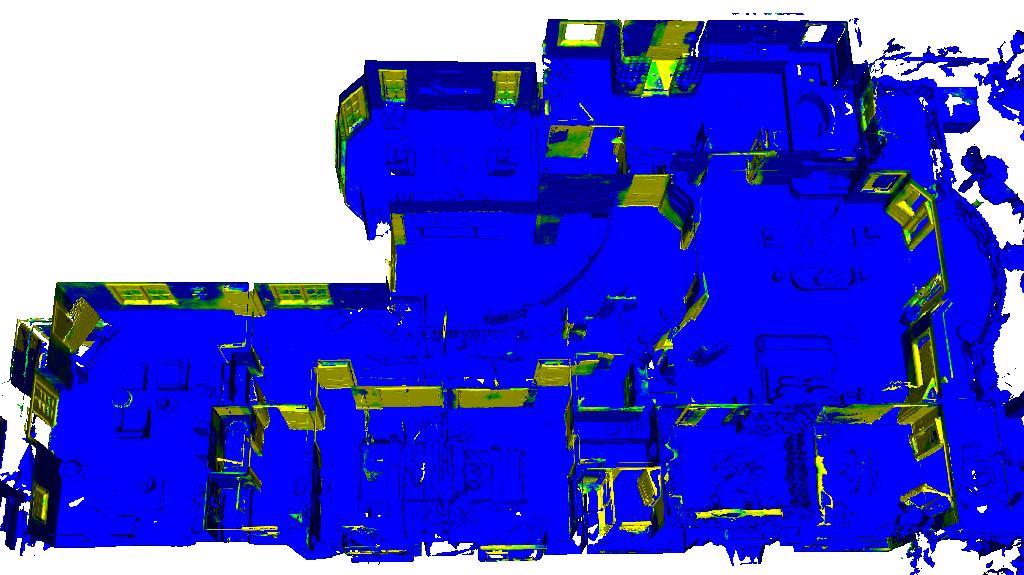} \\
            ``outdoor'' & ``open''\vspace{1em}\\
            \includegraphics[width=\eewidth]{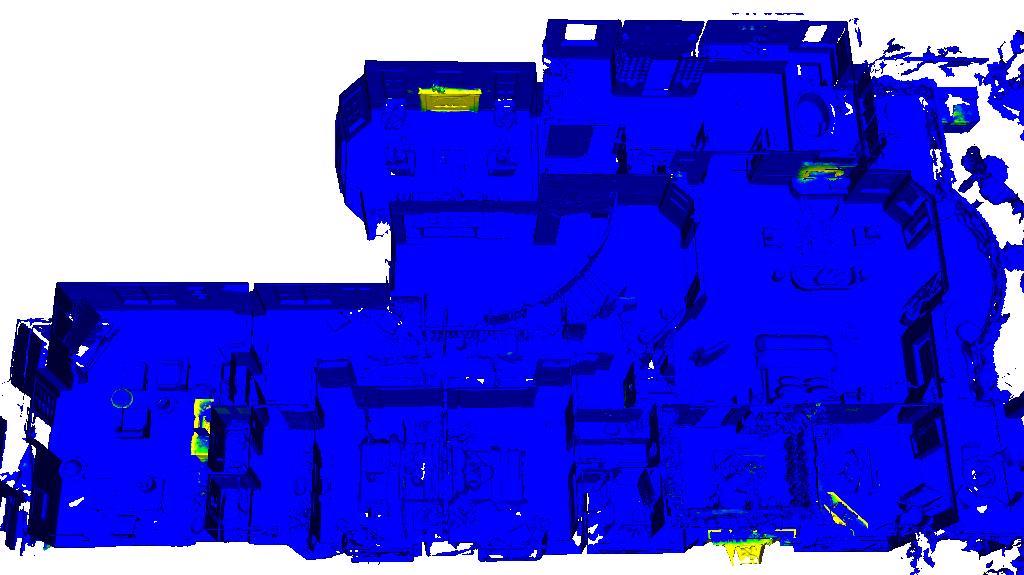} &
            \includegraphics[width=\eewidth]{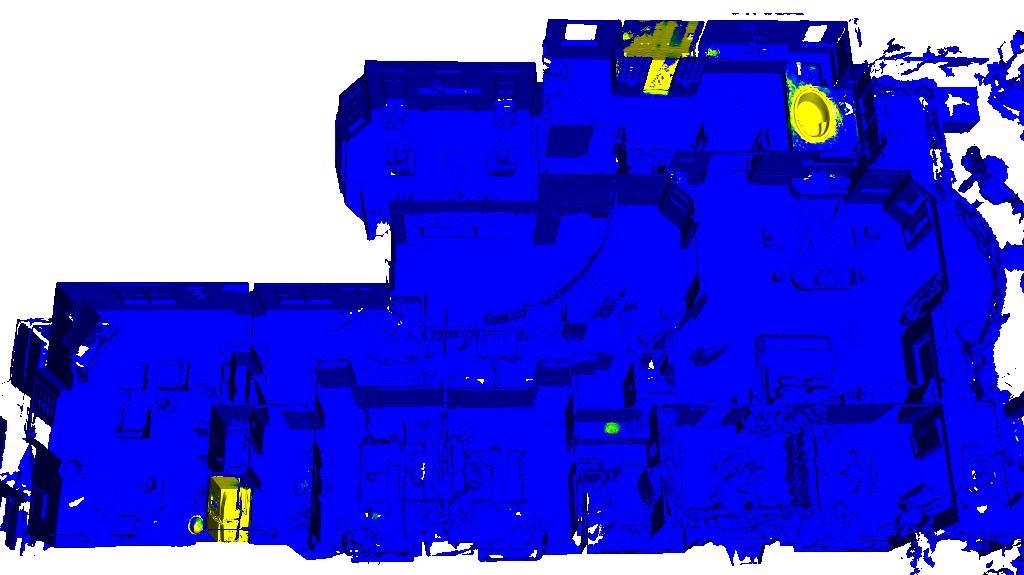} \\
            ``fire'' & ``water''\vspace{1em}\\
         \end{tabular}
        \caption{
        \textbf{Open-Vocabulary Queries for Abstract Concepts.}
        }
        \label{fig:abstract_queries}
\end{figure*}

\end{document}